\documentclass[iicol,sn-mathphys-num]{sn-jnl}

\usepackage{graphicx}%
\usepackage{multirow}%
\usepackage{amsmath,amssymb,amsfonts}%
\usepackage{amsthm}%
\usepackage{mathrsfs}%
\usepackage[title]{appendix}%
\usepackage{xcolor}%
\usepackage{textcomp}%
\usepackage{manyfoot}%
\usepackage{booktabs}%
\usepackage{algorithm}%
\usepackage{algorithmicx}%
\usepackage{algpseudocode}%
\usepackage{listings}%
\usepackage{threeparttable}
\usepackage{colortbl} 
\usepackage{subcaption}
\usepackage{multirow}
\usepackage{graphicx}
\usepackage{float}
\usepackage{booktabs}
\usepackage{pifont}
\usepackage[table,xcdraw]{xcolor}
\usepackage{makecell}
\usepackage{siunitx} 
\usepackage{newfloat}
\usepackage{listings}
\usepackage{tcolorbox}
\usepackage{pdfrender}
\usepackage{xcolor}

\newcommand{\rev}[1]{{\color{black}#1}}

\theoremstyle{thmstyleone}%
\theoremstyle{thmstyletwo}%

\theoremstyle{thmstylethree}%

\tcbuselibrary{breakable}

\newcommand{\mybold}[1]{%
  \textpdfrender{
    TextRenderingMode=2, 
    LineWidth=0.2pt,     
  }{#1}%
}

\definecolor{supervisorframeColor}{rgb}{18, 153, 254}  
\definecolor{supervisorBackColor}{rgb}{251,238,254} 
\newsavebox\CBox

\begin{document}

\title[Article Title]{StoryVideoQA: Scaling Deep Video Understanding with a Large-Scale, Multi-Genre and Auto-Generated Dataset}


\author[1,2,3]{\fnm{Zhengqian} \sur{Wu}}\email{2023202110068@whu.edu.cn}
\author[1]{\fnm{Zhixian} \sur{Liu}}
\author[1]{\fnm{Aodong} \sur{Chen}}
\author[1]{\fnm{Jingyang} \sur{Zhang}}
\author[1,2,3]{\fnm{Ruizhe} \sur{Li}}\email{2020300004016@whu.edu.cn}
\author[1]{\fnm{Hanlin} \sur{Ge}}\email{gehanlin@whu.edu.cn}
\author[1,2,3]{\fnm{Zhongyuan} \sur{Wang}}\email{wzy\_hope@163.com}
\author[1,2,3]{\fnm{Chunxia} \sur{Xiao}}\email{cxxiao@whu.edu.cn}

\author*[1,2,3]{\fnm{Chao} \sur{Liang}}\email{cliang@whu.edu.cn}
\affil[1]{\orgdiv{School of Computer Science}, \orgname{Wuhan University}, \orgaddress{\city{Wuhan}, \postcode{430072}, \state{Hubei}, \country{China}}}

\affil[2]{\orgdiv{National Engineering Research Center for Multimedia Software}, \orgaddress{\country{China}}}

\affil[3]{\orgdiv{Hubei Key Laboratory of Multimedia and Network Communication Engineering}, \orgaddress{\country{China}}}



\abstract{
    Video question answering (VideoQA) aims to answer questions about given videos. While existing approaches excel on factoid VideoQA, they struggle with deep video understanding (DVU), which requires the comprehension of complex storylines.
    This challenge arises from the inherent long-range video content, multi-faceted question types, and instance-level story elements, all of which constrain the scale and diversity of manually constructed DVU datasets.
    These difficulties constrain the scale and diversity of manually-constructed DVU dataset.
    To address these, we previously introduced StoryMind to automatically construct DVU datasets with balanced fine-grained topics. Though it can generate high-quality question-answer pairs (QAs) for TV series, it suffers significant performance degradation when handling longer and more complex movies.
    In this paper, we further design StoryMindv2, an enhanced multi-agent collaboration framework to generate high-quality DVU datasets for both TV series and movies. By integrating a novel supervisor-guided generation mechanism and a refined multi-reviewer voting strategy, the framework is utilized to construct StoryVideoQA, the largest DVU dataset to date, featuring over 363K QAs on 393.2 hours diverse story videos including TV series (avg. 1,635 seconds) and movies (avg. 7,878 seconds). 
    Comprehensive evaluations of 20 state-of-the-art VideoQA methods on this large-scale benchmark reveal that they cannot fully maintain long-range character associations or construct a coherent understanding of complex storylines.
    To bridge this gap, we propose PlotTree, a novel video understanding agent, re-organizing long-range video content into a hierarchical plot structure, 
    enabling efficient storyline reasoning on StoryVideoQA. Project page: \url{https://github.com/nercms-mmap/StoryVideoQA/}
}





\maketitle
\section{Introduction}\label{Introduction}
\begin{figure*}[t]
	\centering  
    \includegraphics[width=\linewidth]{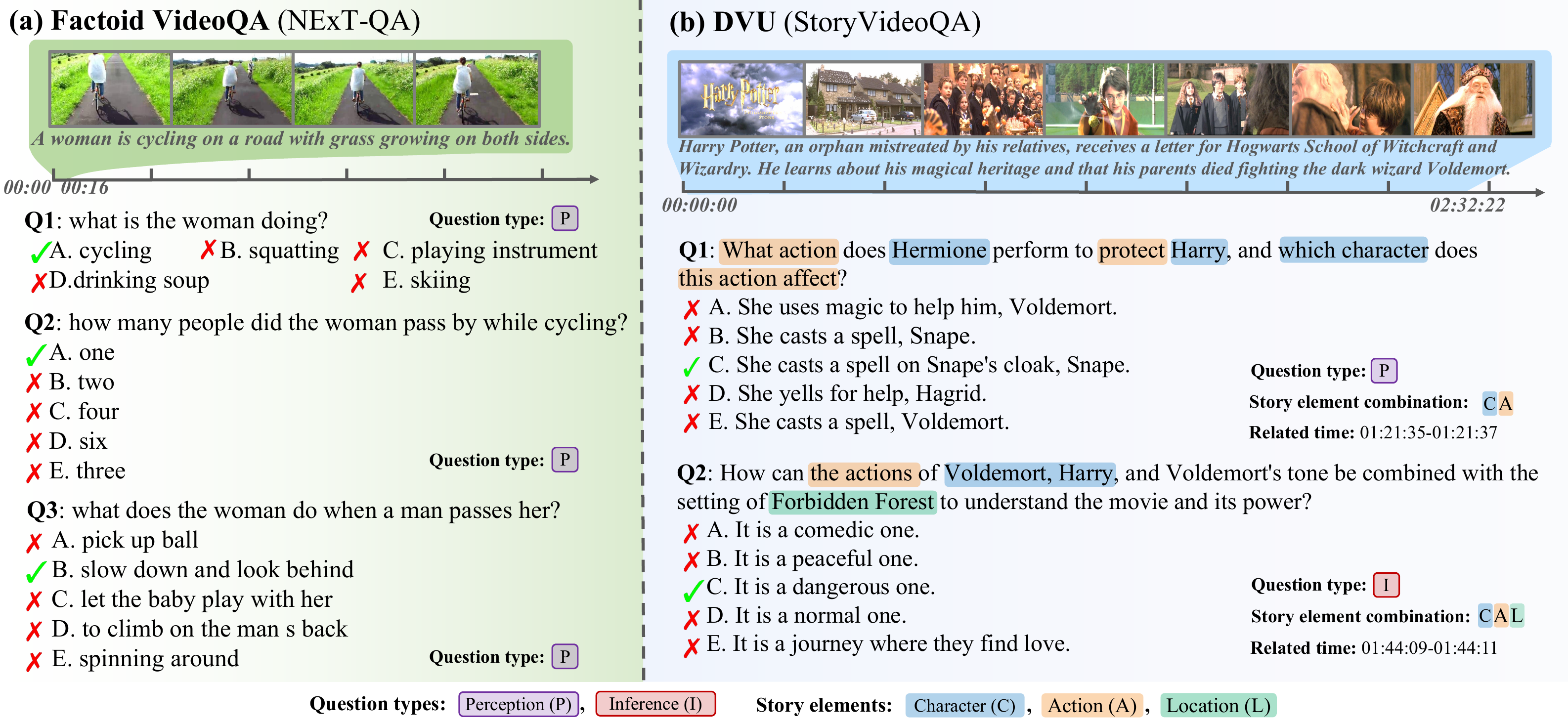}
    \caption{Comparisons of factoid VideoQA and DVU.}
    \label{fig:example}
\end{figure*}

\begin{figure}[t]
	\centering  
    \includegraphics[width=0.98\linewidth]{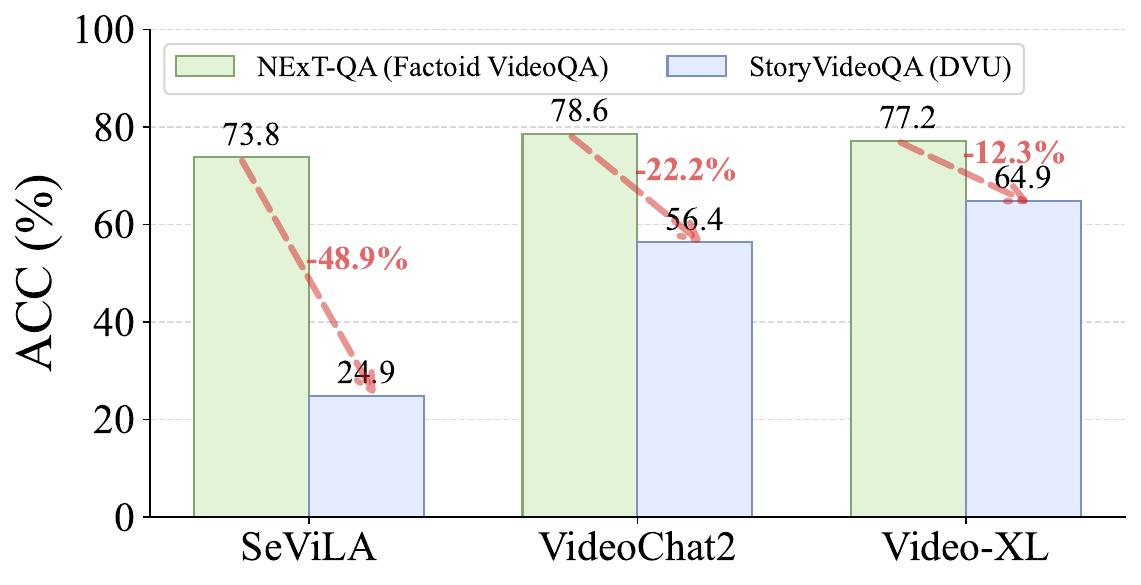}
    \caption{VideoQA methods' performance declines from factoid VideoQA (NExT-QA \cite{nextqa}) to DVU Dataset (StoryVideoQA).}
    \label{fig:factoid_dvu}
\end{figure}

Video question answering (VideoQA) aims to answer questions about given videos, supporting advanced applications like video grounding \cite{timechat, LITA-eccv24, icml2024-momentor}, video summarization \cite{cvpr25-videosummarization, cvpr24-scalingvideosummarization,cvpr23-alignsummarize}, interactive video recommendation system \cite{icmr25-recommendation1, mmm24-talksee, Galanopoulos_2025_CVPR}, and video chatbots \cite{chat-univi,videochatgpt,videochat1}. Early researches mainly focus on factoid VideoQA, which involves answering questions about observable elements such as objects or actions within a short video clip \cite{activitynet-qa} (Figure \ref{fig:example}). However, as the field evolves to more complex deep video understanding (DVU) tasks that require comprehending complex storylines in long story videos, the performance of existing methods significantly declines (Figure \ref{fig:factoid_dvu}).

The performance degradation stems from the complexity of understanding storylines in DVU but absent in factoid VideoQA. Firstly, storylines are inherently long-range. As illustrated in Figure \ref{fig:example}(b), DVU requires reasoning over hours of video content, a stark contrast to the short clips (e.g., 16 seconds in Figure \ref{fig:example}(a)) in factoid VideoQA. Consequently, methods must connect events across vast temporal spans to grasp the storyline's development.
Secondly, storylines are built upon various story elements at the instance level \cite{3w,3wjournal}, involving specific characters (C), their actions (A) and locations (L). For example, question-answer pairs (QAs) in DVU often use specific names like `Hermione' in Figure \ref{fig:example}(b), rather than `Woman' in Figure \ref{fig:example}(a).
Thirdly, comprehending storylines requires moving from perception (P) to complex inference (I). As shown in Figure \ref{fig:example}, unlike factoid VideoQA's perception QAs that ask about observable facts (e.g., ``What is the woman doing?''), DVU includes inference QAs of reasoning about abstract relationships and causality (e.g., ``How can ... be combined to understand the movie?''). This leap from perceiving events to inferring their meaning poses a challenge for current methods.

\begin{figure}[t]
\centering
\includegraphics[width=\linewidth]{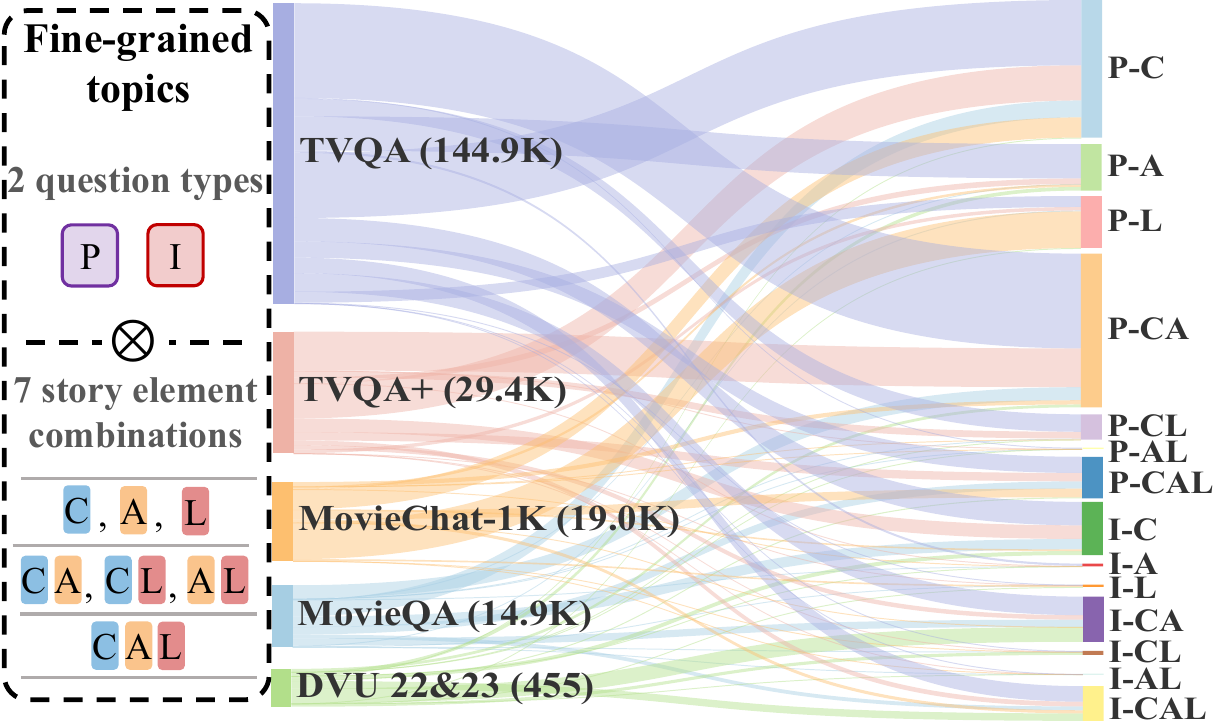}
\caption{The distribution of 5 datasets QAs across the 14 fine-grained topics, based on 2 question types (perception (P) and inference (I)) and 7 story element combinations (character (C), action (A), location (L) and their combinations)}
\label{fig:sankey}
\end{figure}

The inherent complexity of storylines, as mentioned above, poses a significant challenge to the manual construction of DVU datasets, which in turn leads to two primary shortcomings in existing DVU datasets. Firstly, the long-range nature of story videos makes the generation of QAs about storyline laborious and time-consuming, resulting in datasets that are typically small in scale (Table \ref{tab:datasetComparisons}). Secondly, the various story elements and question types inherent in storylines make it difficult for human constructors to ensure a balanced distribution. As shown in Figure \ref{fig:sankey}, we introduce a storyline taxonomy of 14 fine-grained topics by combining 2 question types (perception and inference) with 7 story element combinations (character, action, location, and their combinations). We find most DVU datasets are highly imbalanced across 14 fine-grained topics, hindering a comprehensive evaluation of method's capabilities.

\begin{figure}
    \centering
    \includegraphics[width=\linewidth]{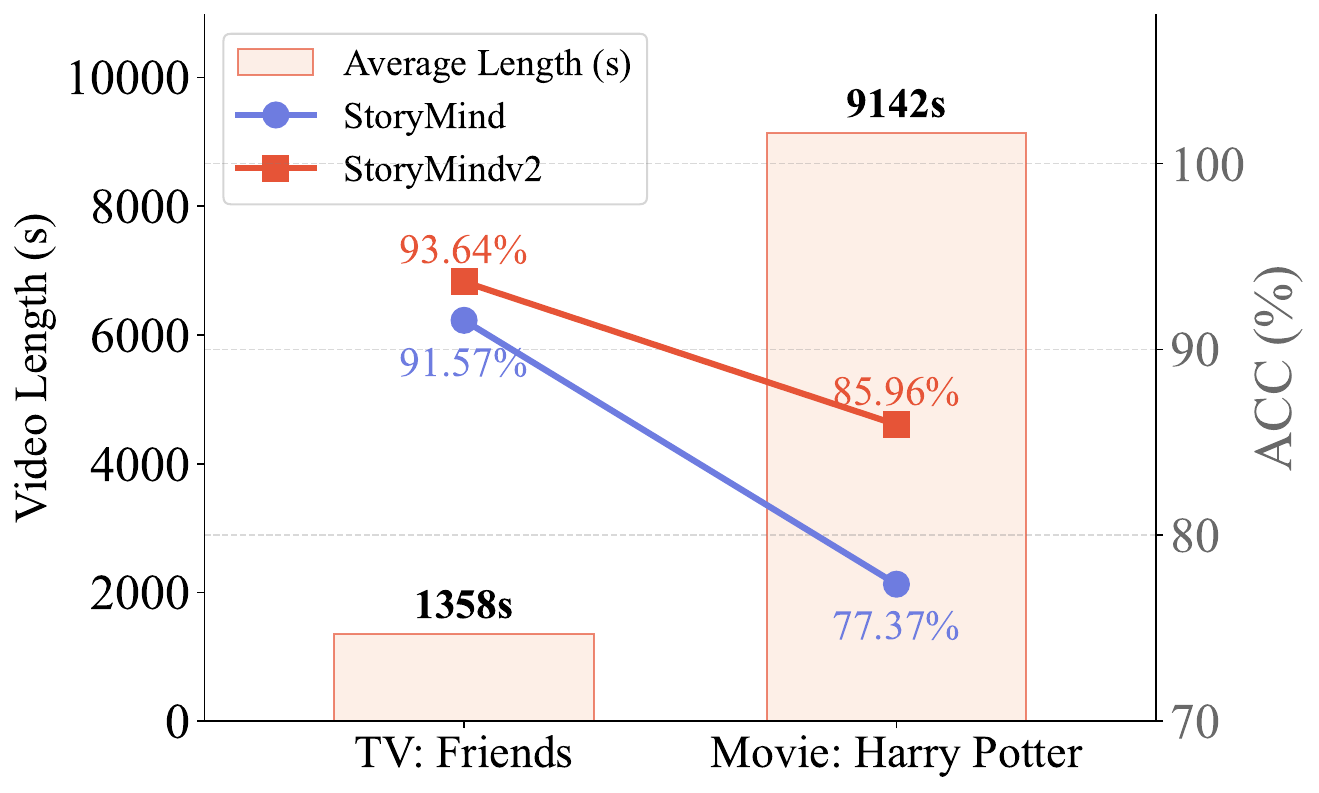}
    \caption{Comparison of the accuracy in automatic QAs generation between StoryMind and StoryMindv2.}
    \label{storymindv1v2}
\end{figure}

To overcome the limitations of manual design of QAs on complex storyline, we previously proposed StoryMind \cite{friendsqa25}, a multi-agent collaboration framework designed to automatically construct DVU datasets, which enables the creation of FriendsQA, a large-scale DVU dataset with balanced distribution across fine-grained topics  (Table \ref{tab:datasetComparisons}). However, StoryMind is primarily applied to episodic TV shows. As shown in Figure \ref{storymindv1v2}, the accuracy of the automatically generated QAs show a notable decline when applied to longer, more intricate movies. It reveals the new challenge of scaling automated data generation to handle more complex storylines and longer-range video content.

\begin{table*}[t]
    \centering
    \caption{Comparisons of existing DVU datasets. Scale compares the number of QAs (\# QAs), the total length (Len.(h)) of all videos, the average duration (Dur.(s)) of videos, QAs density (Den.(h$^{-1}$)) in terms of (\# QAs)/(Len.) and dataset scale (Sca.(h)) in terms of (\# QAs)$\times$(Len.). Fine-grained topic considers the number of fine-grained topics exceeding 5\% of the dataset (\# Fin.) and the balance degree of fine-grained topic distribution. The Gini index (Gin.) and entropy (Ent.) are employed to measure the distribution's balance. The figures around the ``/'' corresponds to TV series and movies, respectively. }
    \label{tab:datasetComparisons}
    \begin{threeparttable} 
        \resizebox{\textwidth}{!}{%
            \begin{tabular}{ccrrcrrcrrcc}
                \toprule
                \multirow{2}{*}[-3pt]{\textbf{Dataset}} & \multirow{2}{*}[-3pt]{\textbf{Venue}} & \multicolumn{5}{c}{\textbf{Scale}}                                                                                           & \multicolumn{3}{c}{\textbf{Fine-grained topic}} & \multirow{2}{*}[-3pt]{\textbf{Type}} & \multirow{2}{*}[-3pt]{\textbf{\begin{tabular}[c]{@{}c@{}}Difficulty \\ Measure\end{tabular}}} \\
                \cmidrule(lr){3-7} \cmidrule(lr){8-10}
                &                                 & \multicolumn{1}{l}{\textbf{Len. (h)}}    & \textbf{\# QAs}  & \textbf{Dur. (s)}    & \textbf{Den.(h$^{-1}$)} & \textbf{Sca. (h)} & \textbf{\# Fin.}     & \textbf{Gin.}     & \textbf{Ent.}     &                                &                                                                                         \\
                \midrule\midrule
                MovieQA \cite{movieqa}            & CVPR'16                         & 381.0         & ~14.9K                                   & 202.7                & 39.11                   & 5.68M             & 6                    & 0.819             & 2.713             & Movie                          & \textcolor{red}{\ding{55}}                                                              \\
                TVQA \cite{tvqa}                  & EMNLP'18                        & \mybold{461.2}       & 144.9K                                         & 76.2                 & 314.18                  & 66.83M            & 8                    & 0.821             & 2.873             & TV                             & \textcolor{red}{\ding{55}}                                                              \\
                TVQA+ \cite{tvqaplus}             & ACL'20                          & 71.7        & 29.4K                                   & 61.5                 & 410.04                  & 2.11M             & 5                    & 0.789             & 2.660             & TV                             & \textcolor{red}{\ding{55}}                                                              \\
                HLVU (DVU 22\&23) \cite{hlvu}     & ICMR'20                         & 24.8        & 455                                 & 106/4,907            & 18.35                   & 0.01M             & 6                    & 0.773             & 2.548             & Movie                          & \textcolor{red}{\ding{55}}                                                              \\
                DramaQA \cite{dramaqa}            & AAAI'21                         & 20.5        & ~17.9K                                  & 3.6/91.8             & 873.17                  & 0.37M             & -                    & -                 & -                 & TV                             & \textcolor{green}{\ding{51}}                                                            \\
                DeepMovieQA \cite{deepmaven}      & EACL'23                         & 41.3        & 1K                                 & 3,102                & 24.21                   & 0.04M             & -                    & -                 & -                 & Movie                          & \textcolor{red}{\ding{55}}                                                              \\
                CinePile \cite{cinepile}          & CVPRW'24                        & 417.6       & ~305K                                 & 160                  & 730.36                  & 127.37M           & -                    & -                 & -                 & Movie                          & \textcolor{red}{\ding{55}}                                                              \\
                MovieChat-1K \cite{moviechat}     & CVPR'24                         & 156.7       & 19.0K                                & 564                  & 121.25                  & 2.98M             & 4                    & 0.701             & 2.203             & Movie                          & \textcolor{red}{\ding{55}}                                                              \\
                
                \rev{LvBench} \cite{zhang2025lvbench}    & \rev{IJCV'25}                  & \rev{209.5}         & \rev{20.0K}                                   & \rev{948}                & \rev{95.76}                   & \rev{4.20M}             & \rev{-}                    & \rev{-}                    & \rev{-}              & \rev{Movie}                          & \textcolor{red}{\ding{55}}                                                              \\
                \midrule
                FriendsQA \cite{friendsqa25}      & AAAI'25                         & 89.6        & 44.6K                                  & 1,358                & 497.77                  & 4.00M             & \mybold{14}          & \mybold{0.927}    & 3.794             & TV                             & \textcolor{green}{\ding{51}}                                                            \\
                StoryVideoQA                      & Ours                            & 393.2       & \mybold{363K}                                 & \mybold{1,635/7,878} & \mybold{923.19}         & \mybold{142.73M}  & \mybold{14}          & \mybold{0.927}    & \mybold{3.795}    & \mybold{TV/Movie}             & \textbf{\textcolor{green}{\ding{51}}}                                                  \\
                \bottomrule
            \end{tabular}
        }
        
        \begin{tablenotes}[flushleft]
            \footnotesize
            \item  \parbox{\textwidth}{ 
                \rev{Note: For a comprehensive comparison involving broader general-purpose long video understanding benchmarks \cite{videomme,LongVideoBench,Video-mmmu,CG-Bench,vrbench}, e.g., Video-MME \cite{videomme}, LVBench \cite{wang2024lvbench}, LongVideoBench \cite{LongVideoBench}, please refer to Table A1 and Section A in the Appendix.}
            }
        \end{tablenotes}
    \end{threeparttable} 
\end{table*}

In this paper, we design StoryMindv2, an enhanced multi-agent collaboration framework designed to enhance the quality of QAs generation for long-range story videos including both TV series and movies. Specifically, it builds on its predecessor's architecture by introducing \rev{three} key innovations.
Firstly, to mitigate accuracy degradation of QA generation from complex storylines, we integrate a novel supervisor-guided mechanism. Equipped with an fault archive, the supervisor can actively identify and rectify generation failures, subsequently providing targeted feedback to the generator. This allows the generator to learn from past faults and enhance its accuracy (Table \ref{tab:wwosupervisor}).
Secondly, to overcome the dataset scale constrains caused by StoryMind's strict consistency filtration strategy, we implement a refined multi-reviewer voting strategy. Under this strategy, a QA pair is accepted if it passes a majority vote. This approach ensures high-quality QAs filtration while simultaneously enabling the construction of a large-scale dataset (Table \ref{tab:consivote}). 
\rev{Lastly, we introduce a novel difficulty measure to evaluate question complexity, candidate answer divergence, and question-answer concordance.}

On this basis, we build the largest DVU dataset to date, StoryVideoQA, featuring over 363K QAs on 393.2 hours of diverse, long-range story videos with balanced coverage across 14 fine-grained topics. Compared to FriendsQA, StoryVideoQA significantly broadens video sources, including 3 TV series (\textit{Friends, The Big Bang Theory, Game of Thrones}) and 78 top-rated movies like \textit{The Shawshank Redemption} from the IMDB\footnote{https://www.imdb.com/} and Douban\footnote{https://www.douban.com/} Top 250 lists, with average video lengths of 1,635s and 7,878s, respectively. With this new benchmark, we comprehensively evaluate the DVU capbility of 20 state-of-the-art (SOTA) VideoQA methods, encompassing video language models (VLMs)-based methods, multimodal large language models (MLLMs)-based methods and video understanding agents methods.

Our evaluation on this new benchmark reveals existing VideoQA methods cannot fully maintain long-range character associations and construct a coherent understanding of storylines. To bridge this gap, we devise PlotTree, a novel large language models (LLMs)-driven video understanding agents. It first converts a video into textual plot nodes, and then, recursively organizes them into a hierarchical PlotTree via node clustering and plot condensation, with the root node encapsulating the entire storyline. This transforms the DVU task into efficient reasoning upon the most relevant nodes across multiple abstraction levels in the tree structure, enabling PlotTree to achieve superior performance in comprehending the long-range evolution of storylines.

\rev{
Compared to the previous conference version \cite{friendsqa25}, this journal version represents a significant expansion in methodological depth, data scale, and evaluative breadth: (1) a enhanced multi-agent framework StoryMindv2 (Section \ref{StoryMindv2}); (2) the StoryVideoQA dataset, which scales the volume to 363K QAs and broadens the genre diversity (Section \ref{StoryVideoQA}); (3) a novel PlotTree method that recursively organizes long videos into hierarchical plot structures for deep reasoning (Section \ref{PlotTree}); and (4) a more comprehensive evaluation on 20 VideoQA methods to reveal the structural limitations of current paradigms (Section \ref{experiments}). 

In summary, our contributions are as follows:
}
\begin{itemize}
    \item We design StoryMindv2, an enhanced multi-agent collaboration framework, featuring a novel supervisor-guided generation mechanism and a refined multi-reviewer voting strategy to enable high-quality, large-scale QAs generation for complex story videos.
    \item We construct StoryVideoQA, the largest and most diverse dataset for DVU to date. It features over 363K QAs on 393.2 hours diverse, long-range story videos (3 TV series and 78 top-rated movies) with balanced coverage across 14 fine-grained topics. We use this as a new benchmark to provide a comprehensive analysis of 20 SOTA methods.
    \item We propose a novel video understanding agents PlotTree. It re-organizes video content into a hierarchical plot structure for efficient reasoning,  achieves superior performance in comprehending the long-range evolution of storylines.
\end{itemize}

\section{Related Work}\label{Related Work}

This section reviews related work in VideoQA, with a focus on datasets and methods. The former surveys existing VideoQA benchmarks, including factoid VideoQA and DVU datasets. The latter discusses recent VideoQA methods, categorizing them into video language models, multimodal large language models and the emerging paradigm of video understanding agents. \rev{For a more comprehensive survey of VideoQA studies, we recommend \cite{vqa20_zhuwenwu,emnlp22vqasurvey,acl24-vlmsurvey,ijcv2025videoqa} to the readers.}

\subsection{VideoQA Datasets}

\vspace{2mm}
\noindent {\textbf{Factoid VideoQA Datasets}.
Early Factoid VideoQA datasets \cite{msvdqa,msrvtt-mc,activitynet-qa,how2qa, liu2024tempcompass,li2024videovista, wu2024star,egoschema} primarily focus on simple visual facts within short-range video clips, such as object recognition \cite{how2qa,nextqa, liu2024tempcompass}, action recognition \cite{activitynet-qa, how2qa, li2024videovista}, and spatial-temporal understanding \cite{li2024videovista, wu2024star,nextqa, egoschema}. For instance, ActivityNet-QA \cite{activitynet-qa} centers on recognizing actions and their temporal relationships, while NEXT-QA \cite{nextqa} delves into videos featuring object interactions. MVBench \cite{mvbench} constructs a unified benchmark for video understanding from existing VideoQA datasets, categorizing its tasks into spatial understanding and temporal understanding. }

Recently, factoid datasets begin to incorporate longer videos. EgoSchema \cite{egoschema} consists of over 5,000 multiple-choice QAs, each QAs corresponds to a 180 seconds daily-life video clip from Ego4D \cite{2022ego4d}. However, their primary limitation remains: they do not focus on the long-range evolvement of a complex storyline. The QA pair rarely require a method to track and understand specific characters, actions, and locations with specific name as they develop within an complex storyline, which is the core requirement of DVU. This is a major factor contributing to the significant performance gap between factoid VideoQA and DVU.

\vspace{2mm}
\noindent \textbf{DVU Datasets}.
Story videos, including TV series and movies, are characterized by intricate interactions and long-range evolvement of story elements in storyline \cite{storyvideo1,storyvideo2,storyvideo3}. 
It requires methods to achieve a deep understanding of the evolvement of storyline (i.e., DVU). However, these unique characteristics of DVU present
significant challenges for manual design of QAs, which gives rise to two primary shortcomings, i.e., limited scale and imbalanced fine-grained topic distribution.

Firstly, manual design of QAs for long story video are laborious and time-consuming. Different from constructing DVU datasets for short video clips \cite{tvqa,tvqaplus, movieqa,dramaqa, moviechat, cinepile,zhang2025lvbench}, constructing DVU dataset for long story videos is particularly challenging. Therefore, most of the DVU dataset for TV series and movies are relatively small in scale \cite{hlvu,deepmaven}. For example, HLVU \cite{hlvu} include only 455 QAs across the DVU 2022 \cite{dvu22} and DVU 2023 \cite{dvu23} grand challenges\footnote{https://sites.google.com/view/dvuchallenge2023/home}. \rev{Even recently proposed benchmarks like LvBench \cite{zhang2025lvbench} are limited to 20K QAs due to the heavy reliance on manual annotation}, hindering comprehensive evaluations of complex storylines.
Secondly, as illustrated in Figure \ref{fig:sankey}, the majority of current DVU datasets \cite{movieqa, tvqa, tvqaplus, moviechat, hlvu} lack a balanced distribution of fine-grained topics. Most of the QAs are focused on perception QAs, particularly regarding the perception of character and action story elements. This is mainly due to early DVU tasks limiting QAs to the prediction of  interactions and relations \cite{Kukleva_2020_CVPR, hlvu, dvu23, whu23}. \rev{While recent LvBench categorize questions into six question types that evaluate various perceptual and cognitive capabilities, they still lack a specialized focus on the story elements inherent in movies.}

This phenomenon indicates existing datasets are difficult to provide a thorough and detailed analysis of the VideoQA methods' capability in DVU. 
Recently, leveraging the powerful comprehension and generation capabilities of LLMs, StoryMinds \cite{friendsqa25}  introduces a multi-agent collaboration framework with a generator and two reviewers to automatically construct the FriendsQA dataset.

However, StoryMind primarily focuses on the TV series \textit{Friends}, and suffers a significant performance degradation when applied to longer movies (Figure \ref{storymindv1v2}). To this end, we propose a upgraded StoryMindv2 framework with a novel supervisor-guided generation mechanism and a refined multi-reviewer voting strategy to ensure high-quality automated QAs generation and filtration for both TV series and movies. On this basis, we construct StoryVideoQA, the largest DVU dataset to date with over 363K QAs on 393.2 hours diverse and long-range story videos including  both TV series and movies.

\subsection{VideoQA Methods}
\vspace{2mm}
\noindent \textbf{Video Language Models}.
Before the advent of LLMs, VLMs is the primary foundation for VideoQA approaches \cite{alpro22,Singularity, mplug2,vid-tldr,violetv2,acl24-vlmsurvey}. Their shared process usually involves two steps: firstly, achieving robust video-text alignment, and secondly, feeding the aligned video and text features into a text decoder for text generation training. ALPRO \cite{alpro22} proposes a video-and-language pre-training framework that utilizes a video-text contrastive loss and prompting entity modeling to facilitate effective cross-modal alignment, enabling the pre-trained model to achieve excellent performance on VideoQA. mPLUG2 \cite{mplug2} introduces a multi-module composition network to address modality entanglement during multi-modal pretraining. 
VIOLETv2 \cite{violetv2} introduces masked visual modeling during pre-training to enhance cross-modal alignment on video and text by randomly masking video frames and predicting target features.

Despite these advances, a significant limitation behind these VLMs-based methods is that their text generation capabilities require extensive pre-training on large-scale video-text datasets. This makes it challenging for these methods to handle open-ended QAs effectively.

\vspace{2mm}
\noindent \textbf{Multimodal Large Language Models}. The recent emergence of LLMs \cite{chatGPT,geminiv1,geminiv1.5, gpt4v,llama,vicuna,llama2,flant5-new} has revolutionized this landscape by replacing the traditional text decoder, evolving VLMs into MLLMs \cite{mllmsurvey}. It first achieves remarkable success in visual question answering \cite{blip2,llava,flamingo} and subsequently extends to the VideoQA \cite{videochat1, videollama1,videollava, mvbench, timechat, chat-univi, videollama2, videochatgpt, videollama3}. A typical approach Video-LLaVA \cite{videollava} maps the visual representations of video frames into the language feature space of an LLMs, allowing the LLMs to comprehend video content. However, the inherent context limitations of LLMs pose a significant challenge for long-range videos \cite{LongVA,longvu}. To mitigate this, a common strategy is to compress visual tokens \cite{moviechat,malmm, adacm2, vilamp,videoxl}. ViLAMP \cite{vilamp} introduces a differential distillation to preserve task-relevant information. Similarly, AdaCM$^2$ \cite{adacm2} employs a cross-modality attention module and layer-wise video memory reduction to decrease the memory footprint of the key-value (KV) cache.

Despite their powerful video comprehension capabilities, MLLMs struggle to follow the narrative development within story videos. A critical weakness is their difficulty in identifying specific characters by name \cite{3wjournal} and tracking their evolution throughout a storyline. Therefore, directly applying MLLMs to DVU tasks still faces significant challenges and difficulties.

\vspace{2mm}
\noindent \textbf{Video Understanding Agents}.
Besides MLLMs that directly fuse modalities, another emerging paradigm utilizes LLMs-driven agents \cite{videoagentsurvey, acl24-vlmsurvey, wang2024videoagent, Liu_2025_CVPR, cvpr25drvideo} for video understanding. 
Early works like LLoVi \cite{LloVi23} first employs VLMs, e.g. LaViLa \cite{lavila} or BLIP2 \cite{blip2}, as captioners to generate textual captions, which are then fed into an LLMs for VideoQA tasks. 
MM-VID \cite{mmvid} utilizes powerful vision models like GPT-4V \cite{gpt4v} to achieve superior performance. Recent researches focus on developing more sophisticated and efficient agentic workflows. DoraemonGPT \cite{yang2024doraemongpt} employs the MCTS planner to decompose a question into an action sequence to guides the reasoning process, while OmAgent \cite{omagent} introduces a divide-and-conquer loop to tackle complex problems. To enhance detail perception, VideoAgent \cite{wang2024videoagent} integrates an object detector for explicit object data. To improve efficiency on long videos, VideoTree \cite{wang2025videotree} transforms the Retrieval-Augmented Generation (RAG) process for captions into a three-layer tree node retrieval.

However, these agents methods essentially optimize RAG over a flat collection of video captions. It is misaligned with the primary challenge of DVU: understanding the long-range evolvement of a complex storyline. This makes it difficult for existing RAG-based video understanding agents to adapt to the unique challenges of DVU.

\begin{figure*}[t]
\centering
\includegraphics[width=\linewidth]{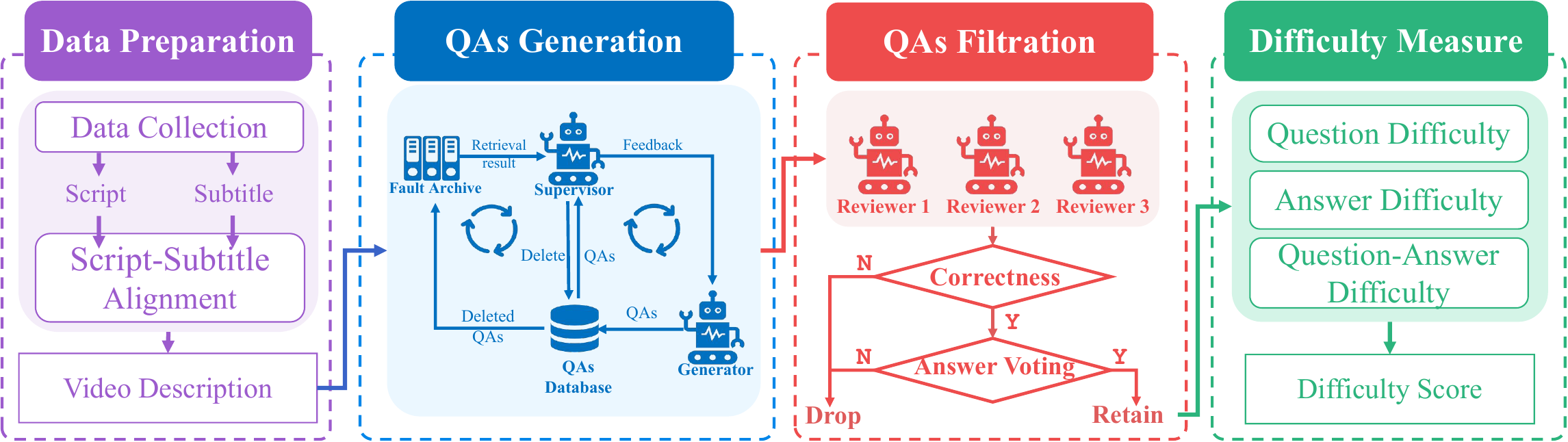}
\caption{The workflow diagram of the enhanced multi-agent collaboration framework StoryMindv2.}
\label{fig:arch}
\end{figure*}

\section{StoryMindv2}\label{StoryMindv2}

In this paper, we propose StoryMindv2, an enhanced multi-agent collaboration framework designed to automatically generate large-scale, high-quality QAs with balanced fine-grained topics for long-range videos. As shown in Figure \ref{fig:arch}, our framework consists of four main stages: data preparation, QAs generation, QAs filtration, and difficulty measure.

\subsection{Data Preparation} \label{datapreparation}
The foundation of StoryMindv2 is a large-scale, high-quality corpus of time-aligned scripts and subtitles. The data is repared through a meticulous two-stage process: data collection and script-subtitle alignment.

\begin{figure}[t]
    \centering
    \includegraphics[width=\linewidth]{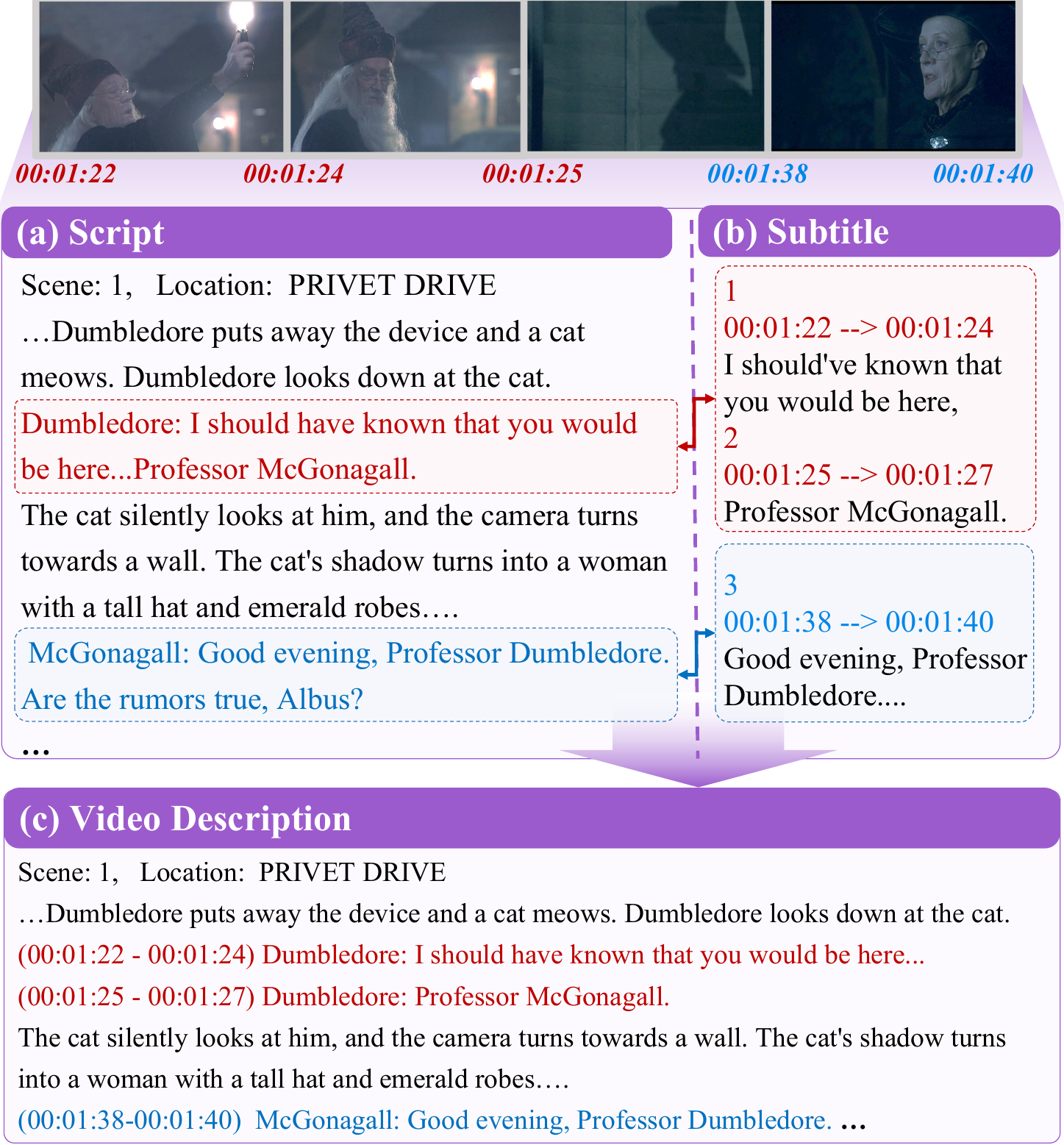}
    \caption{The flowchart of data preparation.}
    \label{fig:data_pre}
\end{figure}

\vspace{2mm}
\noindent \textbf{Data Collection}. 
Our data collection is centered on creating aligned script-subtitle pairs that leverage the complementary strengths of each source. Scripts provide the rich contextual details essential for generating high-quality, context-aware QAs, such as scene locations, character names, and action descriptions (Figure \ref{fig:data_pre}(a)). In contrast, subtitles supply the precise temporal backbone, offering dialogue timestamps crucial for generating QAs time spans and evaluating QA's difficulty (Figure \ref{fig:data_pre}(b)). To build such a resource, we collect both scripts and subtitles, which complement each other by providing contextual richness and precise temporal information.

Our raw data is sourced from two main channels. One is public PAINS dataset \cite{TVCSINS}, which provides automatically aligned and manually verified script-subtitle pairs for all episodes of \textit{Friends} and the first eight seasons of \textit{The Big Bang Theory}. The other is from online sources, comprising all 73 episodes of \textit{Game of Thrones}\footnote{https://genius.com/artists/Game-of-thrones}  and 78 top-rated movies\footnote{https://screenplays.io/}  like \textit{The Shawshank Redemption} from the IMDB and Douban Top 250 lists. For more details on data source, please refer to Section B.1 and Table A2 of the Appendix.

\begin{figure*}[t]
    \centering
    \includegraphics[width=\linewidth]{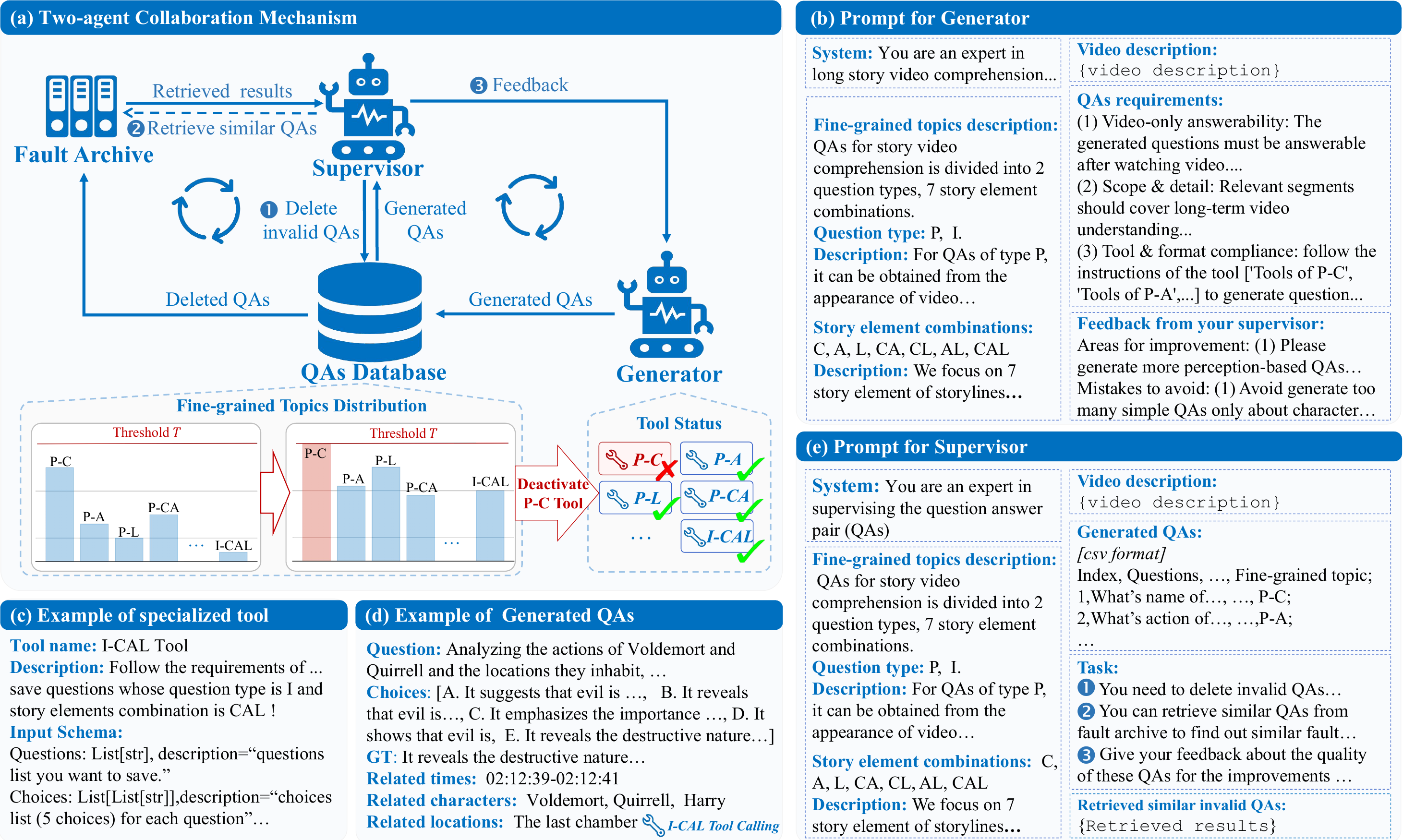}
    \caption{The flowchart of QAs generation, red workflows details the tool deactivation for topics balance. }
    \label{fig:qg}
\end{figure*}
\vspace{2mm}
\noindent \textbf{Script-Subtitle Alignment}.
To ensure the quality of the scripts and subtitles collected from the internet, we first apply the dynamic time warping (DTW) algorithm, used in PAINS dataset \cite{TVCSINS}, to align the script with the subtitle (Figure \ref{fig:data_pre}(b)). Subsequently, each aligned script-subtitle pair would be manually verified to correct any mismatched content (Refer to Appendix B.2 for manual alignment details.). 
To account for deviations between scripts and final video content \cite{movidescriptioncvpr, movidescriptionijcv}, script material (e.g., scene description or dialogue) is discarded during manual verification if inconsistent with the released product. This ensures high fidelity to the source.

Through this meticulous process of collection, alignment, and verification, StoryMindv2 obtains a large-scale corpus of high-fidelity, time-aligned scripts for a diverse range of story videos. As shown in Figure \ref{fig:data_pre}(c), this becomes a crucial foundation for the QAs generation prompts, which we refer to as the $\texttt{video description}$.

\subsection{QAs Generation} \label{questiongeneration}
StoryMindv2 introduces a two-agent collaboration mechanism (Figure \ref{fig:qg}(a)) powered by Gemini-2.0-flash\footnote{https://aistudio.google.com/} to produce a large-scale and high-quality QAs covering all 14 fine-grained topics in balance. It includes a generator with a QAs database and a superior with fault archive.

\vspace{2mm}
\noindent \textbf{Generator}. Following the approach of StoryMind \cite{friendsqa25}, StoryMindv2 provides the generator with comprehensive context (Figure \ref{fig:qg}(b)), including the $\texttt{video description}$ as prepared in Section \ref{datapreparation} and $\texttt{fine-grained topics description}$. The latter contains descriptions for the 14 fine-grained topics, where each is a combination of one of 2 question types (P, I) and one of 7 story element combinations (C, A, L, CA, CL, AL and CAL).
It's used to prompt the generator to generate QAs that are relevant to the specific story video and aligned with different fine-grained topics and save to QAs database (Refer to Appendix B.3 for the prompt of generator).

To ensure a balanced distribution across all fine-grained topics, StoryMindv2 designs a dynamic control mechanism that leverages a suite of specific tools for each fine-grained topic. Unlike the single-tool approach in StoryMind, StoryMindv2 designs a specialized tool (Figure \ref{fig:qg}(c))  for each fine-grained topic's QAs generation (e.g., the generated multiple-choice QA example with five options shown in Figure \ref{fig:qg}(d)), which allows for individual moderation. The system tracks the QAs count for each fine-grained topic and deactivates the corresponding tool upon reaching a threshold $T$, forcing the generator to select from the remaining active tools and thus preventing overproduction (red workflows in Figure \ref{fig:qg}(a)). This control mechanism prevents the generator from overproducing QAs belonging to specific fine-grained topics and is crucial for achieving a balanced final dataset.

\begin{table}[t]
    \caption{Comparison of QAs generation quality and mean time cost per QA pair without (w/o) and with (w/) supervisor.}
    \label{tab:wwosupervisor}
    \setlength{\tabcolsep}{1.3mm}{
    \begin{tabular}{ccc}
    \toprule
    \mybold{Quality} & {\mybold{\begin{tabular}[c]{@{}c@{}} StoryMind \\ (w/o supervisor)\end{tabular}}} & {\mybold{\begin{tabular}[c]{@{}c@{}}  StoryMindv2 \\ (w/ supervisor)\end{tabular}}} \\
    \midrule
    ACC (\%) $\uparrow$        & 49.50                                                            & \mybold{62.30}                                                            \\
    Self-BLEU-2 (\%) $\downarrow$ & 80.10                                                             & \mybold{79.30}                                                              \\
    Self-BLEU-4 (\%) $\downarrow$ & 50.10                                                             & \mybold{48.90}                                                             \\
    Time (min) $\downarrow$ & \mybold{0.081} & 0.365  \\
    \bottomrule
    \end{tabular}%
    }
\end{table}

\vspace{2mm}
\noindent \textbf{Supervisor}. Unlike StoryMind, StoryMindv2 introduces a supervisor whose purpose is to inspect the generator's output, delete erroneous QAs, and provide targeted feedback. As shown in Figure \ref{fig:qg}(e), the supervisor is prompted with the same context as the generator, along with the full set of \texttt{generated QAs} (Figure \ref{fig:qg}(d)) just produced by the generator (Refer to Appendix B.4 for the prompt of supervisor). It then performs a comprehensive check on quality and validity and deletes any QAs it deems incorrect. These flawed QAs are then stored in fault archive to serve as an fault archive. Crucially, to generate targeted feedback, the supervisor employs a RAG step powered by Multilingual-E5 model \cite{wang2024multilingual} to retrieve top-10 similar QAs from this story video's fault archive and synthesize similar past mistakes from this memory. This process allows it to provide targeted guidance on the generator's weaknesses, thus improving generated QAs' quality even before the final filtration stage. To enhance generation efficiency, both the generator's QAs generation and the supervisor's QAs checking are performed in batches.

To demonstrate the effectiveness of this improvement, we design a comparative experiment on 2,000 QAs generation by manual verification. We compare the quality of QAs generated by the generator with and without a supervisor, as shown in Table \ref{tab:wwosupervisor}. The supervisor significantly improves both the accuracy (ACC $\uparrow$) and diversity (Self-BLEU-2 $\downarrow$ and Self-BLEU-4 $\downarrow$) of the generated QAs. This suggests that the supervisor, by inspecting faults, deleting invalid QAs, and providing targeted feedback, effectively prevents the accumulation of faults, thereby continuously enhancing QAs quality and balance. Meanwhile, the generation time per QA pair increases to 0.365 minutes. Nevertheless, this cost remains far lower than manually QAs construction ($>$ 2 minutes).

\begin{figure}[t]
    \centering
    \includegraphics[width=\linewidth]{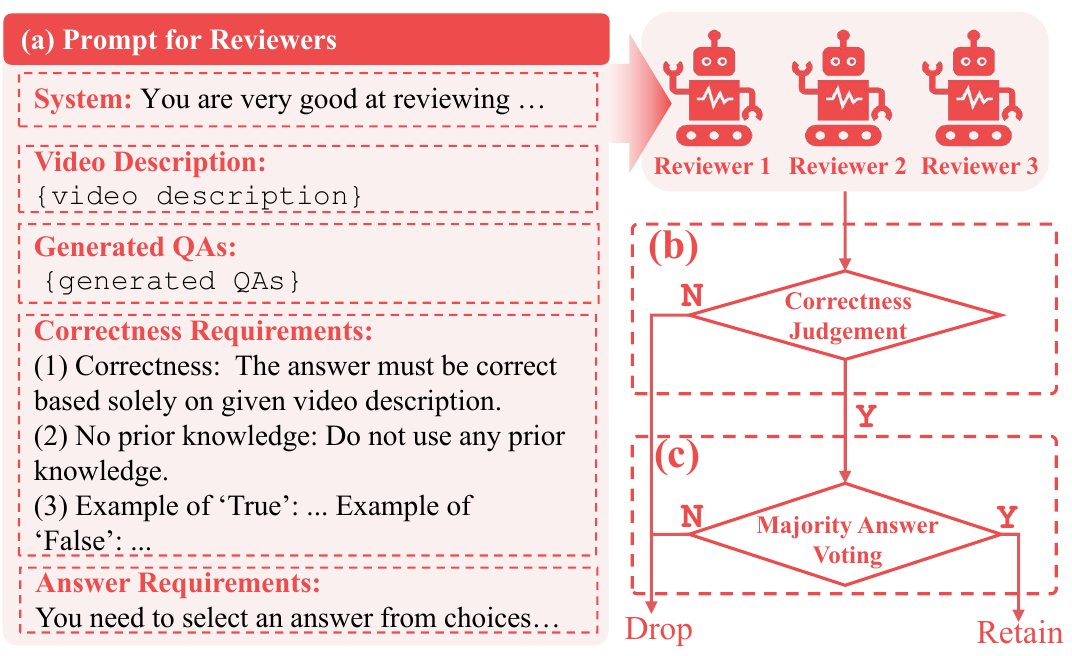}
    \caption{The flowchart of QAs filtration.}
    \label{fig:reviewer}
\end{figure}

\subsection{QAs Filtration}\label{sec:filtartion}
As the final quality control step, the QAs filtration stage employs  three reviewers to independently assess each QA's correctness (Figure \ref{fig:reviewer}). Their collective voting removes flawed items, ensuring high dataset accuracy and reliability.

\vspace{2mm}
\noindent \textbf{Reviewers}. StoryMindv2 incorporates three independent reviewers engined by GPT-4.1\footnote{https://chatgpt.com/}, Claude-3.7-Sonnet\footnote{https://www.anthropic.com/claude} and Gemini-2.0-flash to filter incorrect QAs.
To provide each reviewer with sufficient context, StoryMindv2 prompts it with the complete \texttt{video description} in Section \ref{datapreparation} and the \texttt{generated QAs} in Section \ref{questiongeneration} (Refer to Appendix B.5 for the prompt of reviewer). 
As shown in Figure \ref{fig:reviewer}(a), StoryMindv2 then tasks each reviewer with two primary requirements. The former is correctness requirement, where the reviewer must judge if a QA pair is logically sound, answerable from the given video context, and does not require external prior knowledge. It's used for correctness judgement. The latter is answer requirement which prompt the reviewer to select a correct answer for majority answer voting. 
Based on these, each reviewer outputs its assessment.

\vspace{2mm}
\noindent \textbf{Filtration Process}. Firstly, each reviewer output a binary judgment: `True' for a correct QA pair and `False' otherwise. Secondly, for all QAs marked `True', each reviewer independently selects the correct answer from the candidate choices, a selection crucial for the final answer voting stage. The outputs from the three reviewers are aggregated by a rigorous two-step voting process to ensure the final dataset's fidelity:
\begin{itemize}
    \item \textbf{Correctness judgement}. Following StoryMind \cite{friendsqa25}, a QA pair is retained only if it receives a unanimous `True' judgment for correctness from all three reviewers (Figure \ref{fig:reviewer}(b)). 
    \item \textbf{Majority answer voting}. For the QAs that pass the first check, StoryMindv2 introduces an majority answer voting mechanism, by comparing the answers selected by the three reviewers against the original ground-truth answer provided by the generator (Figure \ref{fig:reviewer}(c)).
\end{itemize}

To demonstrate the effectiveness of our multi-reviewer voting strategy, we design a comparative experiment. By comparing the quality of QAs filtered by a strict consistency in StoryMind \cite{friendsqa25} versus refined answer voting strategy (as shown in Table \ref{tab:consivote}), we observe that the answer voting mechanism achieves significantly higher Recall (54.15 $\to$ 67.02\%) while maintaining comparable Precision (90.12\%). This indicates that the traditional consistency strategy, which demands unanimous agreement, leads to a considerable number of valid QAs being discarded. Such a loss becomes particularly pronounced when constructing large-scale datasets. Therefore, by employing a majority voting approach, StoryMindv2 can effectively ensure high QAs quality while simultaneously enabling the construction of extensive DVU datasets.

\begin{table}[]
\caption{Comparison of QAs filtration quality between consistency strategy and answer voting strategy.}
\label{tab:consivote}
\begin{tabular}{ccc}
\toprule
\mybold{Quality} & {\mybold{\begin{tabular}[c]{@{}c@{}}StoryMind\\ (Consistency)\end{tabular}}} & {\mybold{\begin{tabular}[c]{@{}c@{}}StoryMindv2\\ (Answer Voting)\end{tabular}}} \\
\midrule
Precision (\%) $\uparrow$   & {\mybold{91.05}}                                                             & {90.12}                                                                 \\
Recall (\%) $\uparrow$     & {54.15}                                                                      & {\mybold{67.02}}                                                                 \\
F1 score (\%)$\uparrow$ & {67.91}                                                                      & {\mybold{76.87}}  \\
\bottomrule
\end{tabular}
\end{table}

\subsection{Difficulty Measure}
StoryMindv2 evaluates the difficulty of each generated QA pair by measuring question complexity, candidate answer divergence, and question-answer concordance, respectively. 

\vspace{2mm}
\noindent \textbf{Question Difficulty}.  
Following StoryMind \cite{friendsqa25}, question difficulty is measured from two dimensions, including the length of the relevant video segment and the number of involved story elements (e.g., characters and locations).  

\begin{itemize}
    \item \textbf{Segment length.} Shorter segments provide fewer temporal cues, making it harder for models to correctly localize and understand the video content \cite{timechat, Mu_2024_CVPR, aaai_grounding_2025}.  
    \item \textbf{Story elements.} Fewer relevant instances reduce semantic grounding. Although LLMs/MLLMs can leverage prior knowledge to infer answers from partial information, such sparsity still increases reasoning difficulty.  
\end{itemize}

Thus, questions associated with shorter segments and fewer story elements are regarded as more difficult.
For each QA pair, we consider the length of its relevant video segment $\vert L \vert$ and the number of relevant semantic instances $\vert S \vert$. 
Both features are normalized by z-score \cite{prml06} within QAs  with the same video: 
\begin{equation}
z_l = \frac{\lvert L\rvert - \lvert \overline{L}\rvert}{\sigma_{L}}, \qquad
z_s = \frac{\lvert S\rvert - \lvert \overline{S}\rvert}{\sigma_{S}}
\label{eq:zscore}
\end{equation}
where $\lvert \overline{L} \rvert$ and $\lvert \overline{S} \rvert $ denote the mean length and the mean number of semantic instances within the video, and $\sigma_{L}$ and $\sigma_{S}$ are their corresponding standard deviations.
To obtain a standardized score in (0,1), we use a sigmoid-based mapping:
\begin{equation}
D_q = \frac{1 - \text{sigmoid}(z_l)}{2} + \frac{1-\text{sigmoid}(z_s)}{2}
\label{eq:sigmoid}
\end{equation}

\vspace{2mm}
\noindent \textbf{Answer Difficulty}. Answer difficulty measures how easily models can distinguish the correct answer from distractors. Empirically, both high and low similarity between the correct answer and distractors correspond to greater difficulty: high similarity implies distractors are semantically close to the correct answer, while low similarity prevents the model from using effective exclusionary reasoning.
This non-monotonic relationship can be naturally captured by an entropy \cite{shannon,entropy1,entropymeasures} formulation. Specifically, for the correct answer $a_g$ ($g \in \{1,2,..,5\}$) in QA pair, we compute the BERTScore similarity \cite{bertscore} between $a_g$ and each distractor $a_i$ ($i \neq g$), and average them as:
\begin{equation}
B_a = \frac{1}{4} \sum_{\substack{i=1,i \neq g}}^5 \text{BERTScore}(a_g, a_i)
\end{equation}

Similarly, we apply z-score normalization withine all QAs in StoryVideoQA and sigmoid scaling to normalize $B_a$ into $\hat{B}_a \in (0,1)$. Finally, the entropy-based difficulty score is defined as:
\begin{equation}
D_a = 1 - \Big[-\hat{B}_a\log(\hat{B}_a) - (1-\hat{B}_a)\log(1-\hat{B}_a)\Big]
\end{equation}
where larger $D_a$ indicates higher answer difficulty.

\begin{figure}[t]
    \centering
    \includegraphics[width=0.8\linewidth]{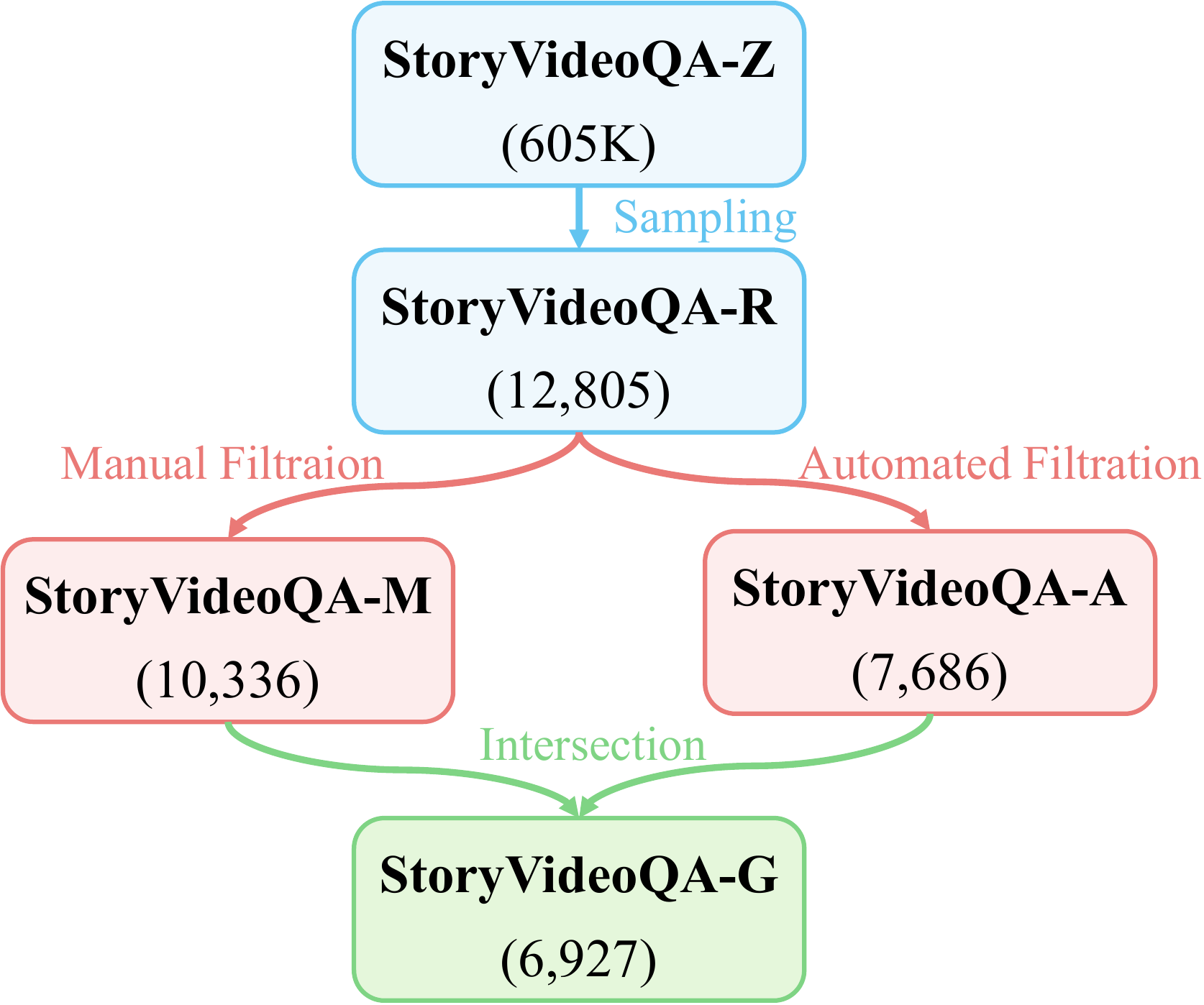}
    \caption{The validation process of StoryVideoQA.}
    \label{fig:subset}
\end{figure}

\vspace{2mm}
\noindent \textbf{Question-Answer Difficulty}.  
This factor evaluates the semantic gap between the question and its correct answer. A larger semantic gap indicates higher difficulty, as the model must perform deeper reasoning rather than relying on surface-level associations. For each QA pair, we compute the BERTScore similarity between the question $q$ and its correct answer $a_g$
\begin{equation}
B_{qa} = \text{BERTScore}(q, a_{g})
\end{equation}
Since $B_{qa}$ are distinctly differentiated in the dataset, StoryMindv2 directly applies min-max normalization to ensure comparability across all QA pairs, and define question-choice difficulty as:

\begin{equation}
D_{qa} = 1 - \frac{B_{qa}-\text{min}(B_{qa})}{\text{max}(B_{qa})-\text{min}(B_{qa})}
\end{equation}

\vspace{2mm}
\noindent \textbf{Overall Difficulty}.  
The overall difficulty is the sum of three equally weighted factors:
\begin{equation}
D(\{q, \{a_{i}\}_{i=1}^5\}) = \frac{D_q + D_a + D_{qa}}{3}
\end{equation}

Figure \ref{fig:diff_dis}(a) presents a statistical analysis of the difficulty distribution of questions and answers across the entire dataset.

\section{StoryVideoQA}\label{StoryVideoQA}
This section begins by evaluating the quality of the auto-constructed StoryVideoQA via a manual assessment of a sampled subset, followed by a statistical analysis of the dataset's scale, composition, and characteristics.

\begin{table}[t!]
\caption{Ablation study of StoryMindv2 on generation accuracy (\%). `Cor.' indicates the correctness judgement, and `Vot.' indicates the answer voting.}
\label{tab:ablate_storymind}
    \setlength{\tabcolsep}{3mm}{
    \begin{tabular}{ccccc}
    \toprule
    \textbf{Cor.} & \textbf{Vot.} & \textbf{TV} & \textbf{Movie} & \textbf{Total} \\
    \midrule
    \midrule
    \multicolumn{5}{c}{\cellcolor[HTML]{D1EBFF} \textit{QAs Generation Stage}} \\
    \textcolor{red}{\ding{55}}            &   \textcolor{red}{\ding{55}}            & 85.94       & 68.35          & 80.71           \\
    \midrule
    \multicolumn{5}{c}{\cellcolor[HTML]{D1EBFF} \textit{QAs Filtration Stage}} \\
    \textbf{\textcolor{green}{\ding{51}}}             & \textcolor{red}{\ding{55}}              & 90.03        & 77.70           & 86.16           \\
    \textbf{\textcolor{green}{\ding{51}}}            & \textbf{\textcolor{green}{\ding{51}}}             & \mybold{91.98}        & \mybold{85.87}           & \mybold{90.12}       \\
    \bottomrule   
    \end{tabular}
    }
\end{table}

\subsection{Dataset Quality}
To validate the generation quality of our StoryMindv2 framework, we conduct manual verification at both the QAs generation and filtration stages, demonstrating its effectiveness and ultimately leading to the construction of the largest DVU dataset to date, StoryVideoQA.

\vspace{2mm}
\noindent \textbf{Validation Process}. As illustrated in Figure \ref{fig:subset}, we design a systematic validation process to evaluate both the QAs generation (Section \ref{questiongeneration}) and QAs filtration (Section \ref{sec:filtartion}) stages of StoryMindv2.
Initially, during the QAs generation phase, StoryMindv2 automatically constructs the StoryVideoQA-Z dataset, comprising 605K QAs. We first sample a raw subset StoryVideoQA-R (12,805 QAs) for detailed validation. This subset is then processed along two parallel branches:
\begin{itemize}
    \item \textbf{Manual Filtration}. Through rigorous manual review on StoryVideoQA-R, we filter out StoryVideoQA-M, a high-quality dataset of 10,336 QAs. This subset represents the manually annotated QAs and serves as the primary reference for evaluating filtration strategy.
    \item \textbf{Automated Filtration}. We apply our proposed automated QAs filtration approach on StoryVideoQA-R, generating automated subset StoryVideoQA-A, containing 7,686 QAs. This subset provides the automatically filtered results, allowing us to quantify the performance.
\end{itemize}

Finally, the intersection of the two subsets yields a gold-standard subset, StoryVideoQA-G (6,927 QAs). 
It's used as the benchmark for video understanding agents in Section \ref{sec:plottree_exp} to reduce the high API cost for full dataset on external LLMs.
Notably, even the smallest StoryVideoQA-G still surpasses existing DVU datasets, e.g., HLVU \cite{hlvu} and DeepMovieQA \cite{deepmaven}, with movie-length story videos in both scale and quality.

\begin{table}[]
\caption{Performance difference on accuracy (\%) between StoryMind-A and StoryMind-G.}
\label{tab:perdiff}

\setlength{\tabcolsep}{2mm}{
    \begin{tabular}{cccc}
    \toprule
                        & \multicolumn{2}{l}{\textbf{StoryVideoQA}}                                  &                              \\\cmidrule{2-3}
\multirow{-2}{*}{\textbf{Method}} & \multicolumn{1}{c}{\textbf{A}} & \multicolumn{1}{c}{\textbf{G}} & \multirow{-2}{*}{$|\Delta|$}      \\

    \midrule\midrule
    SINGULARITY \cite{Singularity}    & 20.47                & 20.44                & 0.03                        \\
    VIOLETv2 \cite{violetv2}       & 15.86                & 15.78                & 0.08                        \\
    Vid-TLDR \cite{vid-tldr}      & 22.21                & 22.39                & 0.18 \\
    SeViLA \cite{sevila}        & 23.72                & 23.66                & 0.06 \\
    VideoLLaMA2 \cite{videollama2}    & 69.52                & 70.13                & 0.61 \\
    VideoChat2 \cite{mvbench}    & 58.51                & 59.23                & 0.72 \\
    Chat-UniVi \cite{chat-univi}      & 30.39                & 30.71                & 0.32 \\
    MA-LMM \cite{malmm}         & 64.22                & 64.69                & 0.47 \\
    TimeChat \cite{timechat}      & 36.79                & 37.36                & 0.57 \\
    Video-ChatGPT \cite{videochatgpt}  & 18.79                & 18.95                & 0.16 \\
    VideoLLaMA3 \cite{videollama3}   & 79.35                & 80.09                & 0.74 \\
    ViLAMP \cite{vilamp}        & 76.75                & 77.34                & 0.59 \\
    Video-XL \cite{videoxl}      & 67.89                & 68.50                & 0.61 \\
    \bottomrule
    \end{tabular}
}
\end{table}

\vspace{2mm}
\noindent \textbf{Manual vs. Automated}. Using manually annotated StoryVideoQA-M as the ground truth, we first examine StoryVideoQA-R, obtaining the result of QAs generation stage without any filtration (The first data line in Table \ref{tab:ablate_storymind}). It reveals that the initial QAs generation accuracy for complex movie QAs is only 68.35\%, significantly lower than 85.94\% for TV series. Then we inspect StoryVideoQA-A to evaluate the utilities of correction judgement and answer voting steps in QAs filtration (Last two lines in Table \ref{tab:ablate_storymind}). The combination of them achieves the best results (90.12\%) and the generation accuracy for movie increases to 85.87\%. It confirms StoryMindv2's effectiveness.

Furthermore, We also compare the performance of 13 SOTA methods (Refer to Section \ref{sec:detail} for more details) on automated subset StoryVideoQA-A and golden standard subset StoryVideoQA-G (Table \ref{tab:perdiff}), observing a maximum performance difference of only 0.8\%. This minimal discrepancy indicates StoryMindv2's QA generation and QA filtration successfully generates and filters a high-quality DVU dataset automatically.

Based on the above validation, StoryMindv2 applies the complete QAs filtration process to the initial 605K QAs (StoryVideoQA-Z) and yield the final high-quality StoryVideoQA dataset, which comprises 363K QAs (Table \ref{tab:scale}).

\subsection{Dataset Statistics}\label{sec:diff_analysis}
Dataset statistics include dataset scale and composition, topic distribution and QA difficulty.

\begin{table}[t]
\caption{Detailed statistics of our proposed StoryVideoQA dataset.}
\centering
\label{tab:scale}
    \setlength{\tabcolsep}{1.5mm}{
    \begin{tabular}{crrrr}
    \toprule
     \multirow{2}{*}[-3pt]{\textbf{Dataset}} & \multicolumn{2}{c}{\textbf{TV}}                        & \multirow{2}{*}[-3pt]{\textbf{Movie}} & \multirow{2}{*}[-3pt]{\textbf{Total}} \\ \cmidrule(lr){2-3}
                     & Sitcom & Drama &                        &                        \\
    \midrule
    \midrule
    \multicolumn{5}{c}{\cellcolor[HTML]{D1EBFF} \textit{Full Set}} \\
    StoryVideoQA-Z    & 328K   & 107K   & 168K   & 605K              \\
    StoryVideoQA      & 202K   & 65K    & 95K    & 363K              \\
    \midrule
    \multicolumn{5}{c}{\cellcolor[HTML]{D1EBFF} \textit{Subset}} \\
    StoryVideoQA-R    & 6,000  & 3,000  & 3,805  & 12,805            \\
    StoryVideoQA-M    & 5,128  & 2,607  & 2,601  & 10,336            \\
    StoryVideoQA-A    & 3,663  & 1,686  & 2,337  & 7,686             \\
    StoryVideoQA-G    & 3,346  & 1,574  & 2,007  & 6,927             \\
    \bottomrule
    \end{tabular}%
    }
\end{table}

\vspace{2mm}
\noindent \textbf{Dataset Scale and Composition}. 
By applying our StoryMindv2 framework to a range of TV series (2 sitcoms series \textit{Friends} and \textit{The Big Bang Theory}, and 1 drama series \textit{Game of Thrones}) and movies, we construct StoryVideoQA, a massive DVU dataset containing over 363K QAs. The dataset is built upon 412 TV episodes, with an average length of 1,635s, and 78 top-rated movies from the IMDB and Douban Top 250 lists, which feature a longer average duration of 7,878s.

\begin{figure*}[!t]
    \centering
    \includegraphics[width=\linewidth]{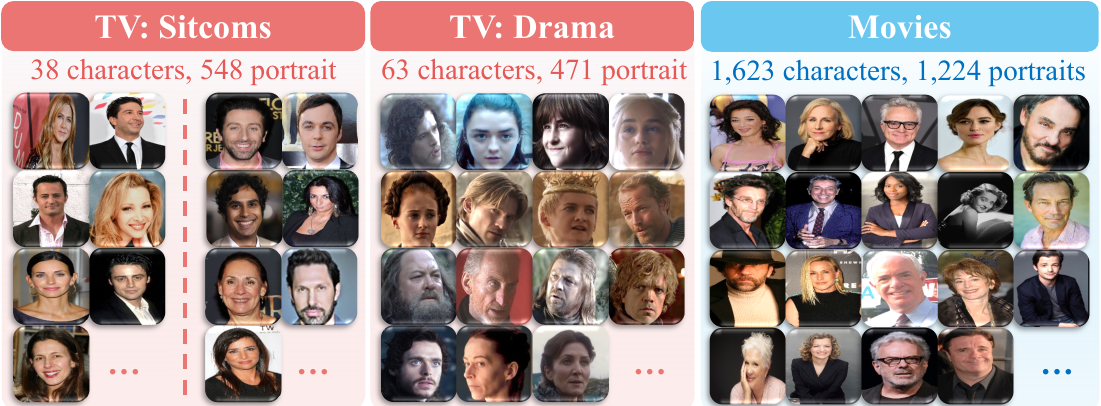}
    \caption{Characers library of StoryVideoQA.}
    \label{fig:char_lib}
\end{figure*}

\begin{figure}[!t]
    \centering
    \includegraphics[width=\linewidth]{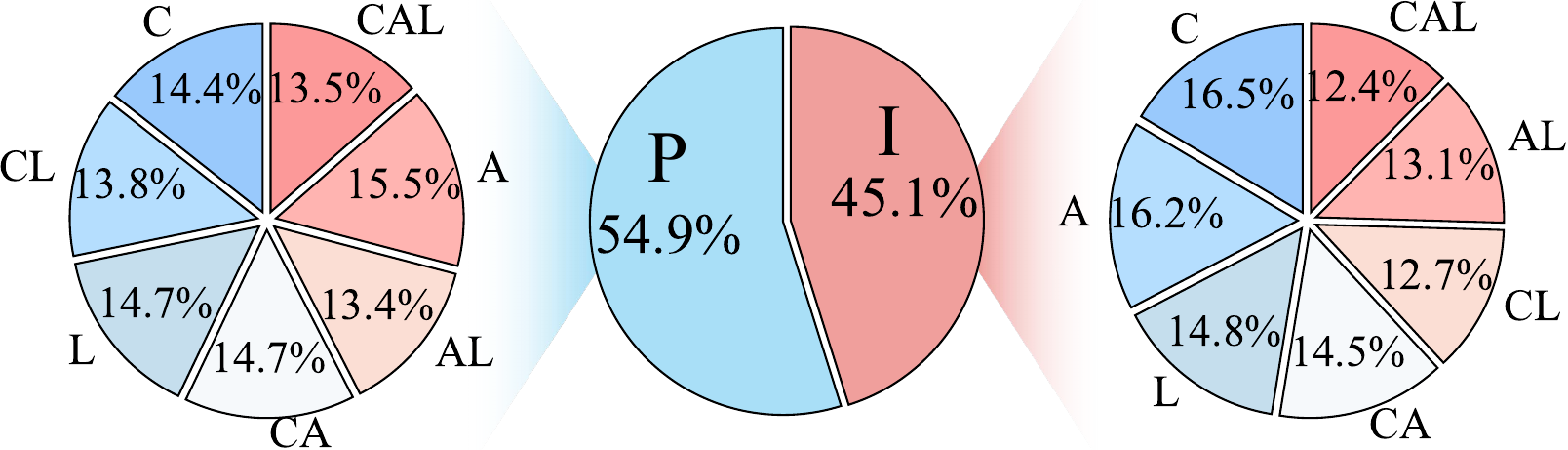}
    \caption{The ditsribution of fine-grained topics in StoryVideoQA.}
    \label{fig:topic_dis}
\end{figure}
To better demonstrate the dataset scale, we report two complementary indicators in Table \ref{tab:datasetComparisons}: QAs density (Den.), which measures the average number of QAs per hour of video (923.19 QAs/h in our dataset), and dataset scale (Sca.), defined as the product of the total number of QAs and the total video duration (142.73M in our dataset). It is worthy noting that such unprecedented density and scale would be nearly impossible to achieve without the automated StoryMindv2 framework.
For more statistics and examples, please refer to Section C and Figure A4 of the Appendix.

Furthermore, recognizing that current VideoQA methods struggle to identify characters \cite{friendsqa25} in story videos, we also construct a comprehensive characer library for the TV series and movies in StoryVideoQA to facilitate future research. As detailed in Table \ref{tab:character} and Figure \ref{fig:char_lib}, this library includes:

\begin{itemize}
\item \textbf{TV}. For \textit{Friends} and \textit{The Big Bang Theory} 2 sitcoms series, the characer library is derived from the PAINS dataset \cite{TVCSINS}, encompassing 38 characters with 548 portrait photos. For \textit{Game of Thrones} drama series, we manually crop a library of 471 portrait photos for the 63 main characters directly from the videos.
\item \textbf{Movie}. For the movie collection, we gather photos of actors corresponding to their movie roles from IMDB\footnote{https://www.imdb.com/}, resulting in a library of 1,623 characters represented by 1,224 actor portraits as actors may appear in multiple movies..
\end{itemize}

\vspace{2mm}
\noindent \textbf{Topic Distribution}.
As illustrated in Figure \ref{fig:topic_dis}, the QAs in StoryVideoQA are distributed between the 2 question types (perception (P) and inference (I)) and 7 story element combinations (C, A, L, and their combinations).  
Though it's generally easier to construct perception QAs, the split shows that percetion QAs (54.9\%) are only slightly more numerous than inference QAs (45.1\%). 
In addition, It is evident that the fine-grained topics are relatively balanced within both the perception and inference, ensuring that our dataset provides comprehensive coverage for evaluating deep video understanding capabilities.

\begin{table}[t]
\caption{Detailed character statistics of our proposed StoryVideoQA dataset.}
\centering
\label{tab:character}
    \setlength{\tabcolsep}{1.5mm}{
    \begin{tabular}{crrrr}
    \toprule
     \multirow{2}{*}[-3pt]{\textbf{Dataset}} & \multicolumn{2}{c}{\textbf{TV}}                        & \multirow{2}{*}[-3pt]{\textbf{Movie}} & \multirow{2}{*}[-3pt]{\textbf{Total}} \\ \cmidrule(lr){2-3}
                     & Sitcom & Drama &                        &                        \\
    \midrule
    \midrule
    \# Character    & ~38    & ~63    & 1,623  & 1,724             \\
    \# Portrait     & 548    & 471    & 1,224  & 2,243             \\
    \bottomrule
    \end{tabular}%
    }
\end{table}

\begin{figure*}[!t]
    \centering
    \begin{subfigure}[t]{0.32\linewidth}
        \centering
        \includegraphics[width=\linewidth]{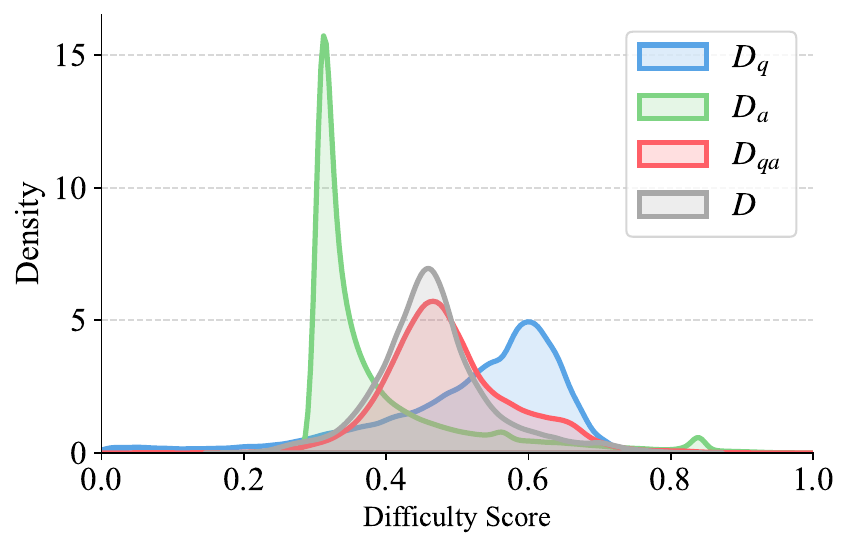} 
        \caption{Distributions of QA difficulties.}
        \label{fig:diff_dist}
    \end{subfigure}
    \hfill 
    \begin{subfigure}[t]{0.32\linewidth}
        \centering
        \includegraphics[width=\linewidth]{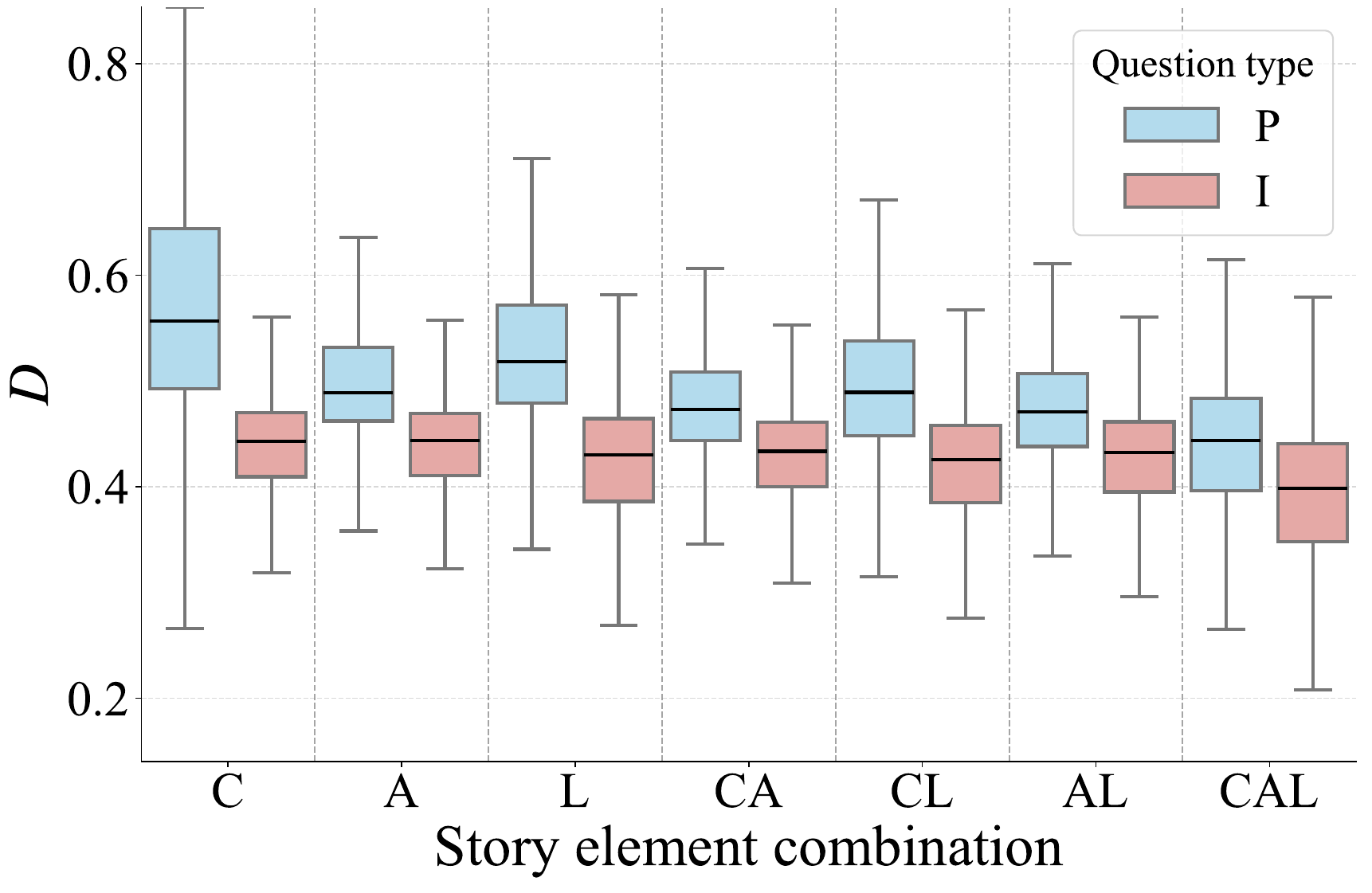} 
        \caption{Difficulty on fine-grained topics.}
        \label{fig:diff_topic}
    \end{subfigure}
    \begin{subfigure}[t]{0.32\linewidth}
        \centering
        \includegraphics[width=\linewidth]{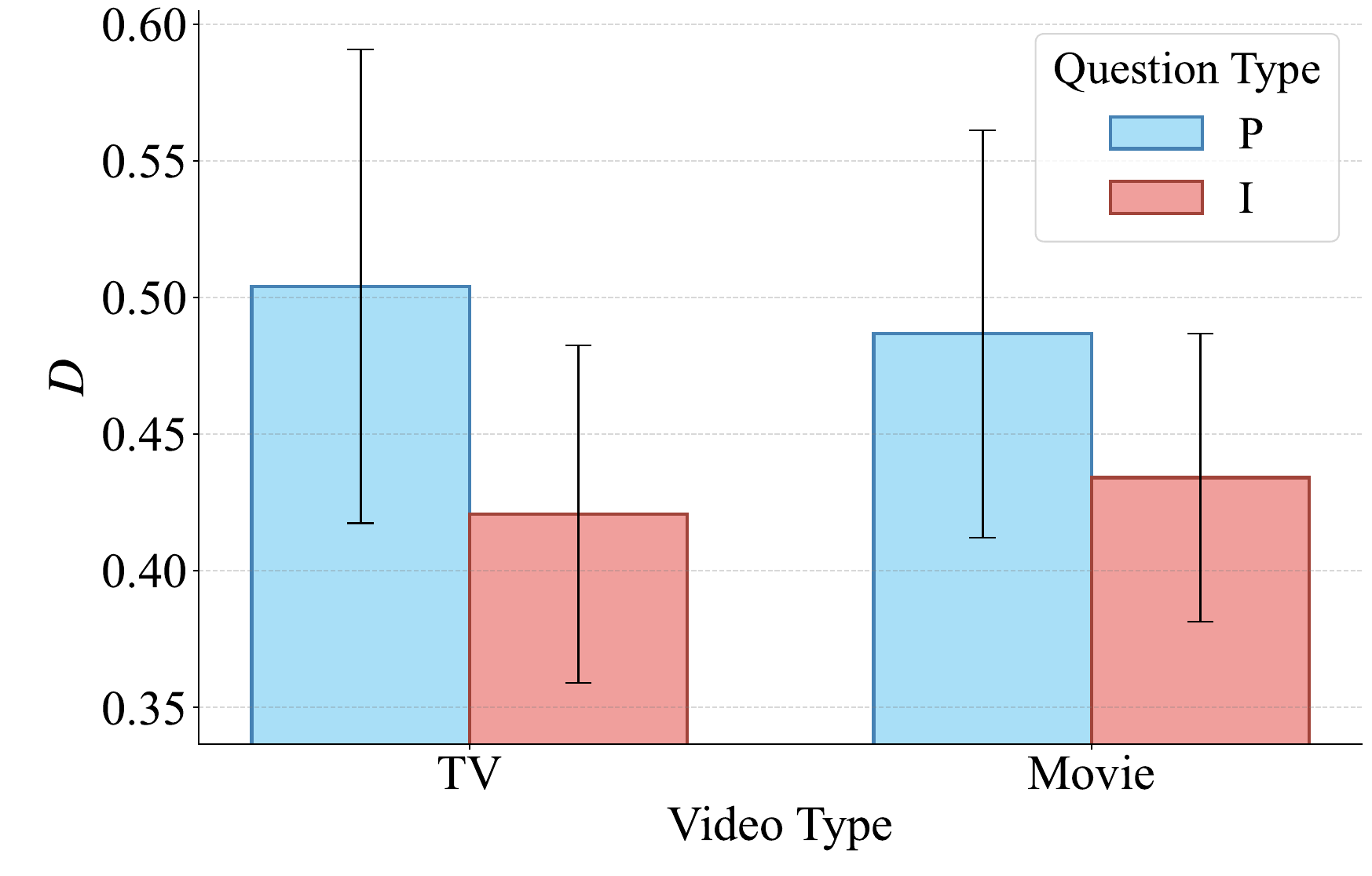} 
        \caption{Difficulty on video type.}
        \label{fig:diff_video_type}
    \end{subfigure}
    \caption{Analysis of QA difficulties from distribution, fine-grained topic and video type.}
    \label{fig:diff_dis}
\end{figure*}
\vspace{2mm}
\noindent \textbf{QA Difficulty}. As shown in Figure \ref{fig:diff_dis}(a), we analyze the distribution of different difficulty measures ($D_Q$, $D_A$, and $D_{QA}$) and overall difficulty $D$ in StoryVideoQA. 
The $D_Q$ distribution is centered around a higher difficulty of $0.60$, in contrast to $D_A$, which is concentrated at $0.35$. Both $D_{QA}$ and the overall difficulty $D$ are approximately normally distributed, with peaks centered at $0.45$. This suggests that the overall difficulty of the StoryVideoQA dataset is well-balanced, providing a sufficient range of challenging and easy questions.

\begin{figure*}[t]
\centering
\includegraphics[width=\linewidth]{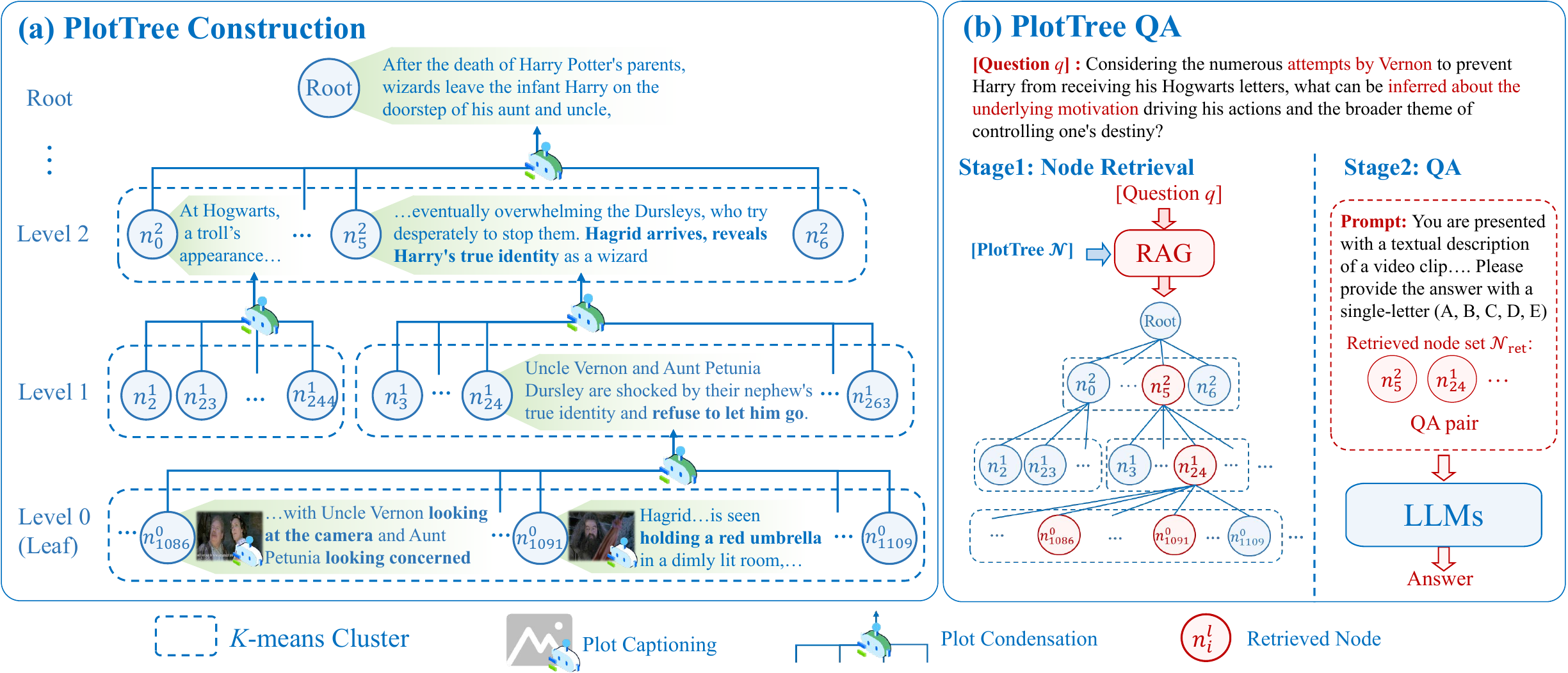}
\caption{The workflows of our proposed video understanding agents PlotTree (\textbf{Bold} texts track to difference abstract level of Nodes in PlotTree).}
\label{fig:plottree}
\end{figure*}

We also analyze the overall difficulty $D$ across 14 fine-grained topics (Figure \ref{fig:diff_dis}(b)). The observed difficulty aligns with our difficulty measure design: Perception QAs, which involve fewer segments and story elements, are generally more difficult than inference QAs. Similarly, QAs on single story element (C, A, or L) tend to exhibit higher difficulty compared to those on composite story elements (e.g., CAL) within perception QAs. However, in inference QAs, the difficulty across different story elements does not show significant variance. This is likely because the segments length and the number of story elements involved in inference questions consistently maintain a relatively large and stable size.

For different video types, we observe contrasting patterns in the mean and standard deviation of difficulty (Figure \ref{fig:diff_dis}(c)). Perception QAs are easier in movies compared to TV series, likely due to movies offering longer story segments for context. Conversely, inference QAs in movies are more difficult than their TV counterparts,  reflecting the increased complexity and longer-range storylines in movies.

\section{PlotTree}\label{PlotTree}

Existing video understanding approaches often represent a video as a flat sequence of discrete events, which struggle to capture the plot's long-range evolution and hierarchical structure. To address this, we propose PlotTree, a novel video understanding method including two phrases: PlotTree construction and PlotTree QA. The former constructs a multi-level representation that organizes plots into a tree structure. The latter effectively converts the DVU task into a RAG problem over the PlotTree to answer questions about story videos (Figure \ref{fig:plottree}).

\subsection{PlotTree Construction}
PlotTree construction consists of 2 steps, i.e., leaf node generation and hierarchical condensation.

\begin{figure}[t]
\centering
\includegraphics[width=\linewidth]{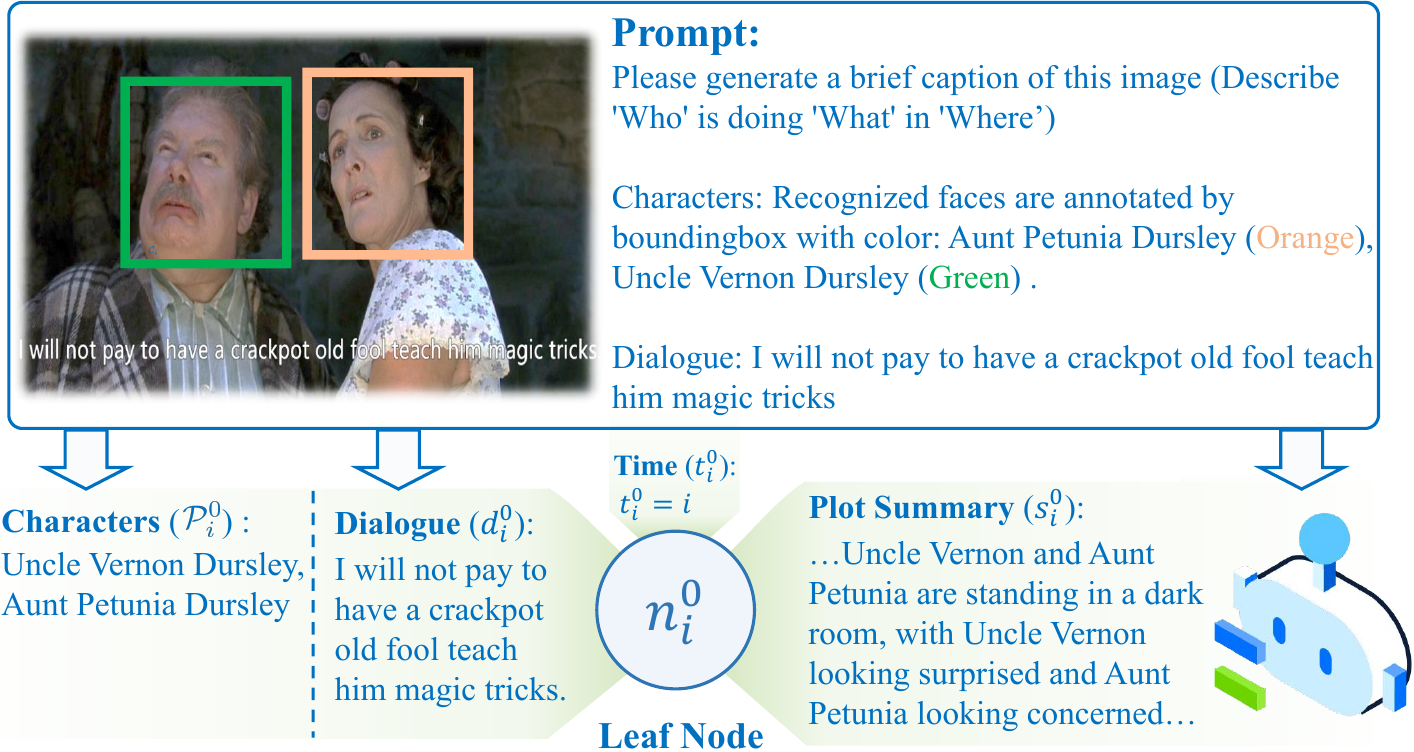}
\caption{Leaf node generation in PlotTree.}
\label{fig:plot_cap}
\end{figure}

\vspace{2mm}
\noindent \textbf{Leaf Node Generation}. This initial step aims to generate plot summary for video frames, explicitly grounding them with character identities and dialogues.
Recognizing that MLLMs exhibits limited ability to consistently link specific characters to their actions and dialogue \cite{han2023autoad1, han2023autoad2,han2024autoad3}, our process begins with explicit character identification. Following prior works \cite{autoad-zero,omagent}, we evenly sample $F$ keyframes and 
leverage InsightFace\footnote{https://github.com/deepinsight/insightface} for face recognition, tagging characters by matching them against character library of StoryVideoQA (Figure \ref{fig:char_lib}). Each character from set $\mathcal{P}_i^0$, identified at the $i$-th keyframe ($i \in \{1,2,...,F\}$) on level 0 of the PlotTree, is annotated with a colored bounding box.
Each annotated keyframe, along with its dialogue $d^0_i$ and a character-to-color map text (e.g.,`Aunt Petunia Dursley (Orange)'), are then used to prompt LLaVA-1.6\footnote{https://huggingface.co/liuhaotian/llava-v1.6-vicuna-7b} for plot captioning, generating a plot summary $s^0_i$ with specific character name, as illustrated in Figure \ref{fig:plot_cap}. 

The resulting data unit is a leaf node, denoted as a quadruple $n^0_i = (\mathcal{P}^0_i, d^0_i, s^0_i, t^0_i)$, where $\mathcal{P}^0_i$ represents character recognition results, $d^0_i$ is the dialogue from the subtitle, and $s^0_i$ is the generated plot summary. Crucially, $t^0_i$ is the corresponding time, defined by the sequential keyframe index $i$.
This structure provides a much richer description than simple captions. By explicitly binding character identity within $\mathcal{P}^0_i$ to dialogue in $d^0_i$ and behaviors in $s^0_i$, these nodes offer solid, unambiguous foundation for the subsequent hierarchical condensation.

\begin{figure}[t]
\centering
\includegraphics[width=0.95\linewidth]{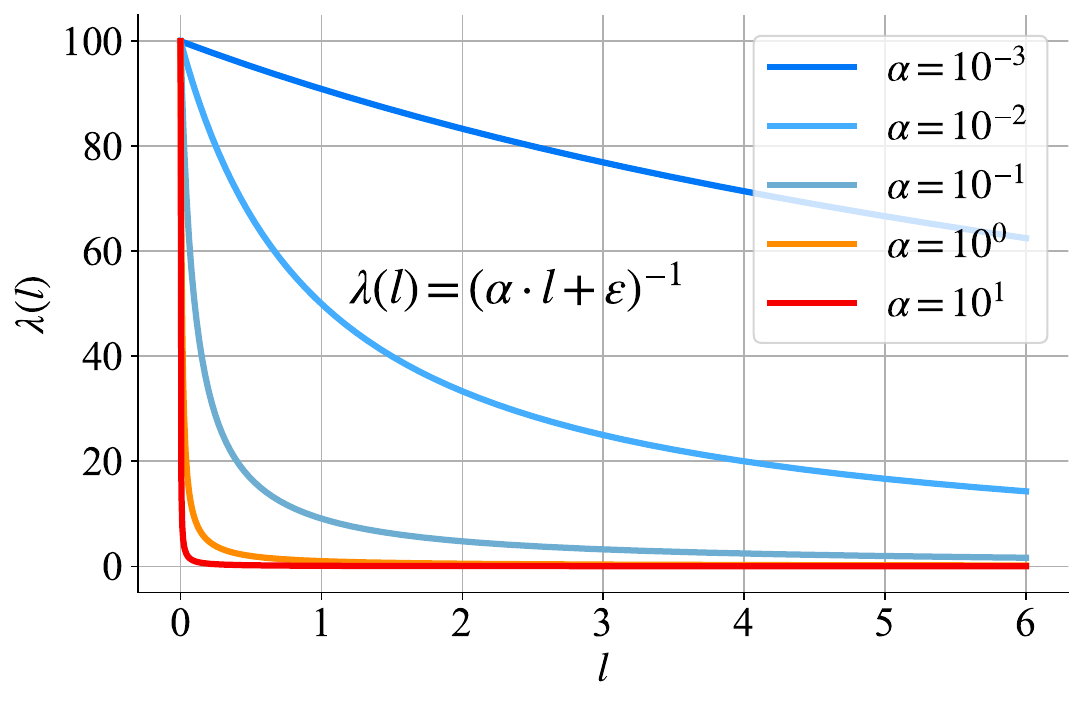}
\caption{$\lambda(l)$ ($\epsilon$=0.01) in different scaling rate $\alpha$.}
\label{fig:decay_func}
\end{figure}

\begin{figure}[t]
\centering
\includegraphics[width=\linewidth]{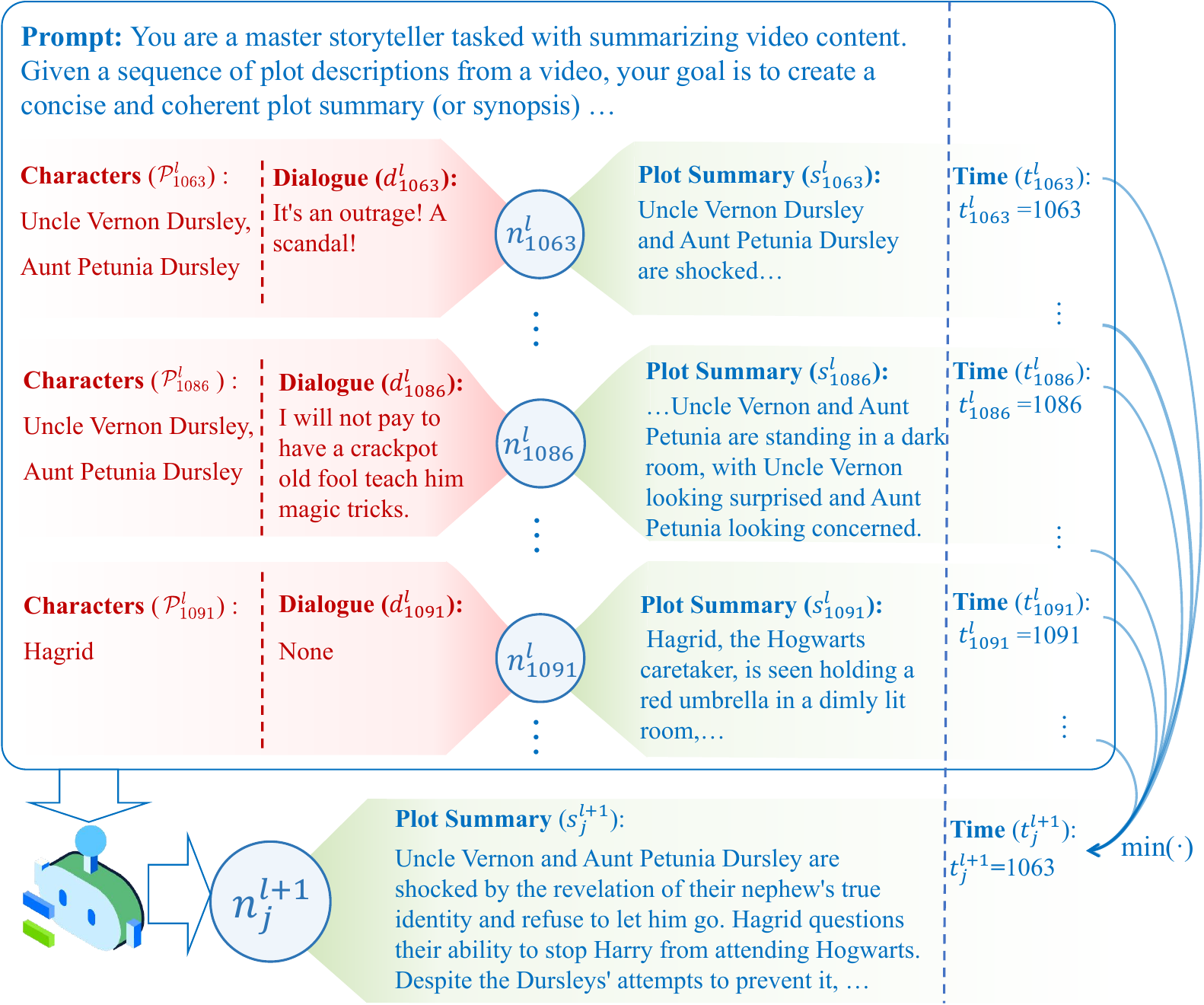}
\caption{Plot condensation of child nodes' textual content for parent node (Red parts exists only when $l=0$).}
\label{fig:nodemerging}
\end{figure}

\vspace{2mm}
\noindent \textbf{Hierarchical Condensation}. 
To transform the flat, linear sequence of leaf nodes $\mathcal{N}^0 = \{n_i^0\}$ into a meaningful plot hierarchy that reveals complex storyline, we employ an iterative, bottom-up condensation process. The core of condensation process is to progressive cluster a set of child nodes $\mathcal{N}^l $ into $K^{l}$ distinct clusters $\{\mathcal{C}^{l}_1, \mathcal{C}^{l}_2, \dots, \mathcal{C}^{l}_{K^{l}}\}$ and condense each cluster $\mathcal{C}^{l}_j$ into a parent node $n_j^{l+1}$ by plot condensation (Figure \ref{fig:nodemerging}), yielding a more abstract parent node set $\mathcal{N}^{l+1}=\{n_j^{l+1}\}$.

Specifically, we approach the above clustering problem using the $K$-Means algorithm. To ensure that higher-level clusters focus more on plot semantic similarity, we design a distance metric $\mathrm{D}^l$ that introduces a decay coefficient $\lambda(l)$ to the temporal distance, shifting the emphasis from temporal to semantic proximity as the hierarchy deepens.
\begin{equation} 
\mathrm{D}^l(n_i^l, n_j^l) =
\lambda(l)\cdot\frac{|t_i^l - t_j^l|}{F} +\left(
1 - \text{cos}(\boldsymbol{e}_i^l, \boldsymbol{e}_j^l)\right)
\end{equation}
where decay function $\lambda(l)=({\alpha\cdot l + \epsilon})^{-1}$ progressively reduces temporal influence at higher hierarchy levels (Figure \ref{fig:decay_func}). Here, $\alpha>0$ is a scaling factor controlling the decay rate, and $\epsilon$ is a small constant ($1 \times 10^{-2}$) to prevent division by zero. $t_i^l$ denotes the time of node $n_i^l$, and $\boldsymbol{e}_i^l$ represents the normalized semantic embedding of its textual content extracted by Qwen3 embedding \cite{qwen3embedding} model $\phi_{\textrm{emb}}$:
\begin{equation}
\boldsymbol{e}_i^l = \begin{cases}
\phi_{\textrm{emb}}\left((\mathcal{P}_i^0, d_i^0, s_i^0)\right), & l=0 \\
\phi_{\textrm{emb}}(s_i^l), & l\ge 1
\end{cases}
\end{equation}
The two branches separately model low-level visual details (e.g., characters and dialogues) and high-level plot semantics.

At each level, for every cluster $\mathcal{C}_j^l$, all nodes $\{n_i^l\}$ are first sorted chronologically by $t_i^l$. Their textual content are fed into Gemini-2.0-flash to generate a single, more abstract plot summary $s_j^{l+1}$. 
The resulting parent node is defined as $n_j^{l+1}=(s_j^{l+1}, t_j^{l+1})$, where $t_j^{l+1}=\min(t_i^l | n_i^l\in \mathcal{C}_j^l)$ is the latest time among child nodes belonging to the cluster $\mathcal{C}_j^l$. The collection of all such nodes constitutes the next level of the tree $\mathcal{N}^{l+1}$.

The number of clusters $K^{l}$ is determined by a compression rate $\beta \in (0,1)$, which dynamically adjusts to video content and controls the compression ratio between successive layers, thus determining the overall depth of the PlotTree $\mathcal{N}$:
\begin{equation} \label{eq:k_clusters}
K^{l} = \max(1, \lfloor \beta \cdot |\mathcal{N}^l| \rfloor)
\end{equation}
where $\lfloor \cdot \rfloor$ denotes the floor operation.

\subsection{PlotTree QA}
To enable a deep understanding of the plot in long videos, we reframe the VideoQA task as a RAG problem over the PlotTree $\mathcal{N}$. Once the PlotTree is constructed, it can be reused for multiple incoming questions without reconstruction. As illustrated in Figure \ref{fig:plottree}(b). This process consists of two main steps: Node retrieval and QA.

\vspace{2mm}
\noindent \textbf{Node Retrieval}. Node retrieval stage aims to capture both macro-level themes and micro-level details simultaneously. 
Given a question $q$, we first extract and encode question $q$ from QA pair into semantic embeddings $\boldsymbol{e}_q$ using Qwen3 embedding model. 
Subsequently, we calculate the semantic similarity between $\boldsymbol{e}_q$ and all nodes' semantic embedding $\{\boldsymbol{e}^l_i | n^l_i \in \mathcal{N}\}$ and select the top-$M$ matching nodes. This forms a retrieval set $\mathcal{N}_{ret}$, that spans multiple levels of the plot hierarchy.

\vspace{2mm}
\noindent \textbf{QA}. The QA stage is responsible for synthesizing the retrieved information and performing reasoning. We extract the plot summaries from each node in the retrieved node set $\mathcal{N}_{ret}$ and sort them chronologically\footnote{Higher level nodes have priority when times are identical.}. These sorted summaries are then concatenated into a single, coherent context paragraph. Finally, this context is combined with the original QA pair to construct a prompt, which is fed to Gemini-2.0-flash to generate the final answer.

\section{Experiments}\label{experiments}

\begin{table*}[t]
\caption{Evaluation results (\%) on StoryVideoQA datasets between VLMs-based and MLLMs-based methods across 14 fine-grained topics (2 question types (P, I) and 7 story element combinations (C, A, L and their combinations)). }
\label{tab:comparison}
\resizebox{\textwidth}{!}{%
\begin{tabular}{cc|rrrrrrr|rrrrrrr|c}
    \toprule
{}       &                          & \multicolumn{7}{c|}{{P}}                                                                                                                                                                                                                                                                                                                                                                                                                                                                                                                                                     & \multicolumn{7}{c|}{{I}}                                                                                                                                                                                                                                                                                                                                                                                                                                                                                                                                                              & {}                                                           \\
\cmidrule{3-9}\cmidrule{10-16}
\multirow{-2}{*}{\textbf{Method}} & \multirow{-2}{*}{\textbf{Venue}} & {C}                                                          & {A}                                                          & {L}                                                          & {CA}                                                         & {CL}                                                         & {AL}                                                         & {CAL}                                                        & {C}                                                          & {A}                                                          & {L}                                                          & {CA}                                                         & {CL}                                                         & {AL}                                                         & {CAL}                                                        & \multirow{-2}{*}{\textbf{Avg.}}                                    \\
\midrule\midrule
\multicolumn{17}{c}{\cellcolor[HTML]{D1EBFF} \textit{VLMs}}                                                                                                                                                                                                                                                                                                                                                                                                                                                                                                                                                                                                                                                                                                                                                                                                                                                                                                                                                                                                                                                                                                                                                                                                                                                                                               \\
{SINGULARITY \cite{Singularity} }                             & {ACL'23}                           & {21.44}                              & {20.59}                              & {20.20}                              & {20.96}                              & {20.59}                              & {20.31}                              & {19.85}                              & {20.53}                              & {20.84}                              & {20.37}                              & {20.61}                              & {19.92}                              & {20.24}                              & {19.90}                              & {20.48}                              \\
{VIOLETv2 \cite{violetv2} }                               & {CVPR'23}                          & {19.27}                              & {16.32}                              & {18.18}                              & {16.86}                              & {20.32}                              & {18.45}                              & {18.30}                              & {12.03}                              & {12.69}                              & {14.15}                              & {12.54}                              & {14.70}                              & {14.68}                              & {14.09}                              & {16.06}                              \\
{Vid-TLDR \cite{vid-tldr} }                                & {CVPR'24}                          & {19.16}                              & {19.16}                              & {21.07}                              & {21.71}                              & {22.42}                              & {21.94}                              & {23.53}                              & {23.11}                              & {24.15}                              & {23.27}                              & {22.89}                              & {23.16}                              & {23.40}                              & {24.31}                              & {22.24}                              \\
\midrule
\multicolumn{17}{c}{\cellcolor[HTML]{D1EBFF} \textit{MLLMs}}  \\
{SeViLA \cite{Singularity}}                                   & {CVPR'23}                          & {30.32}                              & {27.38}                              & {35.44}                              & {24.31}                              & {30.74}                              & {26.65}                              & {22.05}                              & {20.78}                              & {21.08}                              & {22.61}                              & {19.47}                              & {21.69}                              & {21.21}                              & {19.18}                              & {24.89}                              \\
{VideoLLaMA2 \cite{videollama2} }                              & {ArXiv'24}                         & {44.77}                              & {47.03}                              & {52.26}                              & {58.76}                              & {58.96}                              & {62.24}                              & {71.17}                              & {79.31}                              & {77.87}                              & {81.40}                              & {79.69}                              & {79.74}                              & {78.32}                              & {80.22}                              & {66.68}                              \\
{VideoChat2 \cite{mvbench}  }                              & {CVPR'24}                          & {35.53}                              & {39.15}                              & {45.04}                              & {49.53}                              & {50.58}                              & {52.91}                              & {62.95}                              & {65.67}                              & {65.03}                              & {69.59}                              & {66.41}                              & {66.91}                              & {65.98}                              & {69.80}                              & {56.37}                              \\
{Chat-UniVi \cite{chat-univi}}                                & {CVPR'24}                          & {29.35}                              & {41.00}                              & {25.37}                              & {37.98}                              & {28.89}                              & {38.94}                              & {31.31}                              & {28.14}                              & {32.36}                              & {19.55}                              & {28.18}                              & {19.88}                              & {29.47}                              & {22.92}                              & {30.03}                              \\
{MA-LMM \cite{malmm} }                                   & {CVPR'24}                          & {40.95}                              & {45.68}                              & {48.06}                              & {54.26}                              & {55.03}                              & {58.01}                              & {64.45}                              & {72.34}                              & {71.43}                              & {74.49}                              & {71.59}                              & {72.35}                              & {71.78}                              & {74.10}                              & {61.32}                              \\
{TimeChat \cite{timechat}   }                             & {CVPR'24}                          & {20.97}                              & {17.47}                              & {29.83}                              & {26.28}                              & {30.84}                              & {28.24}                              & {30.07}                              & {44.24}                              & {44.13}                              & {47.08}                              & {41.19}                              & {41.69}                              & {39.93}                              & {37.72}                              & {33.49}                              \\
{Video-ChatGPT \cite{videochatgpt} }                            & {ACL'24}                           & {7.41}                               & {18.16}                              & {4.34}                               & {16.94}                              & {9.52}                               & {17.05}                              & {14.00}                              & {31.12}                              & {30.26}                              & {24.91}                              & {29.19}                              & {24.78}                              & {27.51}                              & {23.47}                              & {19.30}                              \\
{Video-XL \cite{videoxl}}                                 & {CVPR'25}                          & {46.75}                              & {58.94}                              & {51.35}                              & {60.91}                              & {57.10}                              & {61.75}                              & {66.67}                              & {75.09}                              & {73.21}                              & {75.98}                              & {73.08}                              & {73.51}                              & {71.71}                              & {73.63}                              & {64.88}                              \\
{ViLAMP \cite{vilamp} }                                 & {ICML'25}                          & {54.19}                              & {66.26}                              & {58.04}                              & {70.70}                              & {67.24}                              & {71.63}                              & {76.03}                              & {84.60}                              & {82.83}                              & {84.91}                              & {83.04}                              & {83.07}                              & {82.22}                              & {83.44}                              & {73.97}                              \\
{VideoLLaMA3  \cite{videollama3} }                            & {ArXiv'25}                         & {\mybold{55.47}}                     & {\mybold{71.00}}                     & {\mybold{62.85}}                     & {\mybold{74.06}}                     & {\mybold{68.88}}                     & {\mybold{74.21}}                     & {\mybold{78.99}}                     & {\mybold{85.56}}                     & {\mybold{83.91}}                     & {\mybold{87.03}}                     & {\mybold{84.22}}                     & {\mybold{84.60}}                     & {\mybold{83.63}}                     & {\mybold{85.53}}                     & {\mybold{76.32}}                     \\
\bottomrule
\end{tabular}%
}
\end{table*}
In this section, we evaluate the proposed StoryVideoQA dataset through a series of experiments. We first describe the experimental setup, followed by benchmarking VLMs-based and MLLMs-based SOTA methods on StoryVideoQA to reveal the key challenges of DVU. We then use our PlotTree and more SOTA methods to further evaluate on StoryVideoQA-G.

\subsection{Experimental Setup}\label{sec:detail}

\vspace{2mm}
\noindent \textbf{Baselines}. We totally benchmark 20 SOTA methods on StoryVideoQA and StoryVideoQA-G, and categorize these methods into 3 groups.
\begin{itemize}
    \item \textbf{VLMs}. VLMs-based method includes Singularity \cite{Singularity}, VIOLETv2 \cite{violetv2} and Vid-TLDR \cite{vid-tldr}. We use the weights finetuned on MSRVTT-QA, the most commonly used VideoQA dataset, to ensure fairness.
    \item \textbf{MLLMs}. MLLMs-based method includes SeViLA \cite{sevila}, Chat-Univi \cite{chat-univi}, MA-LMM \cite{malmm}, TimeChat \cite{timechat}, Video-ChatGPT \cite{videochatgpt}, VideoChat2 \cite{mvbench}, VideoLLaMA2\cite{videollama2}, VideoLLaMA3 \cite{videollama3}, ViLAMP \cite{vilamp}, Video-XL \cite{videoxl}, \rev{VideoChat-Flash \cite{videochat-flash}, Long-VITA \cite{Long-vita}, and Qwen3-VL \cite{qwen3vltechnicalreport}. With the exception of SeViLA, which uses the FlanT5-XL 3B model \cite{flant5-new} as its backbone, all other MLLMs-based methods in our evaluation utilize their officially recommended LLMs ranging from 7B to 14B as their backbone. This includes various versions from both the LLaMA \cite{llama,llama2} and Qwen series \cite{qwen2, qwen2.5, qwen3vltechnicalreport} models. In addition, we also evaluate two frontier MLLMs,  Gemini-3-Flash\footnote{https://deepmind.google/models/gemini/} and GPT-5.2\footnote{https://openai.com/} to establish the current performance ceiling of the StoryVideoQA benchmark.}
    \item \textbf{Agents}. Agents methods includes VideoTree \cite{videotree} and Video2RAG \cite{omagent}. 
\end{itemize}

\vspace{2mm}
\noindent \textbf{Implementation Details}. Following prior works \cite{egoschema, longvu,LongVideoBench, videomme, friendsqa25}, We set methods in zero-shot VideoQA settings on StoryVideoQA, with official default configurations to ensure fairness. 
For VLMs-based and MLLMs-based methods, all experiments are run on 4 $\times$ NVIDIA RTX A6000 GPUs. We use the default or officially recommended number of input frames for each method to ensure fair and reproducible comparisons.
For agents methods, we standardize the core components to isolate the performance of the agentic workflow itself. To maintain consistency across all agents, we employ LLaVA-1.6 as the captioning tool, Qwen3 embedding model for embedding, and Gemini-2.0-Flash as the LLMs. For PlotTree, we set the sample rate of keyframe as 1 fps, scaling factor $\alpha$ as 10, compression rate $\beta$ as 1/36, and extract top-32 matching nodes in PlotTree QA process. Notably, only the Video2RAG and PlotTree methods can utilize the provided character library for enhanced character identification.   
For more implementation details of baselines, please refer to Section D of the Appendix.

\vspace{2mm}
\noindent \textbf{Evaluation Metrics}. We use accuracy \cite{whu23} as
metrics, calculating by dividing the number of correct answered QAs by the total number of QAs.

\subsection{Experiment Result}
Our evaluation employs a dual setting for feasibility and breadth. Only VLMs-based and main MLLMs-based methods are benchmarked on the full set StoryVideoQA. Some methods are assessed on a representative subset StoryVideoQA-G due to the prohibitive API costs and time consuming of evaluating on the full 363K QAs set.

\begin{figure*}[t]
    \centering
    \resizebox{\linewidth}{!}{ 
    \begin{subfigure}[t]{0.32\linewidth}
        \centering
        \includegraphics[width=\linewidth]{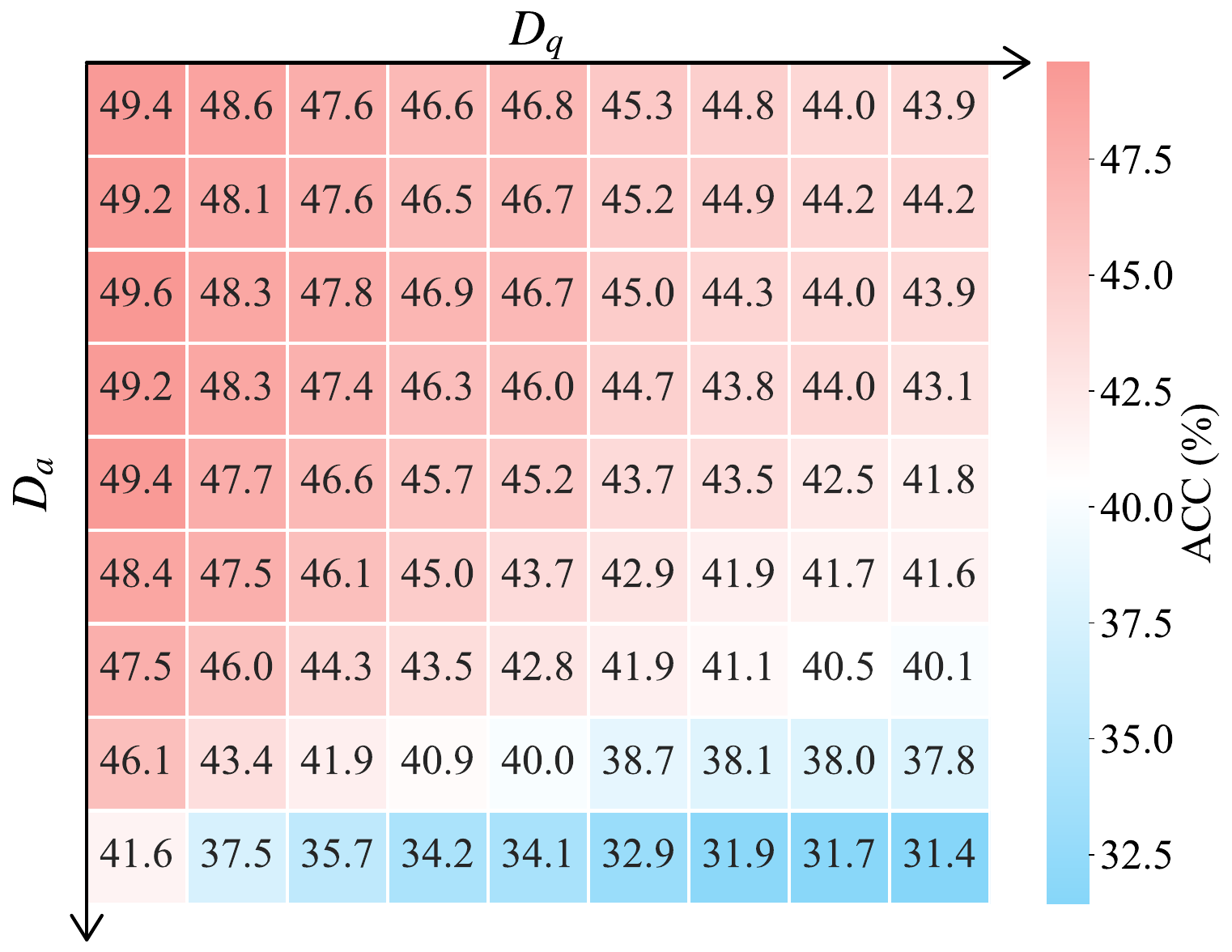} 
        \caption{$D_q$ and $D_a$ }
        \label{fig:diff_factor1}
    \end{subfigure}
    \hfill 
    \begin{subfigure}[t]{0.32\linewidth}
        \centering
        \includegraphics[width=\linewidth]{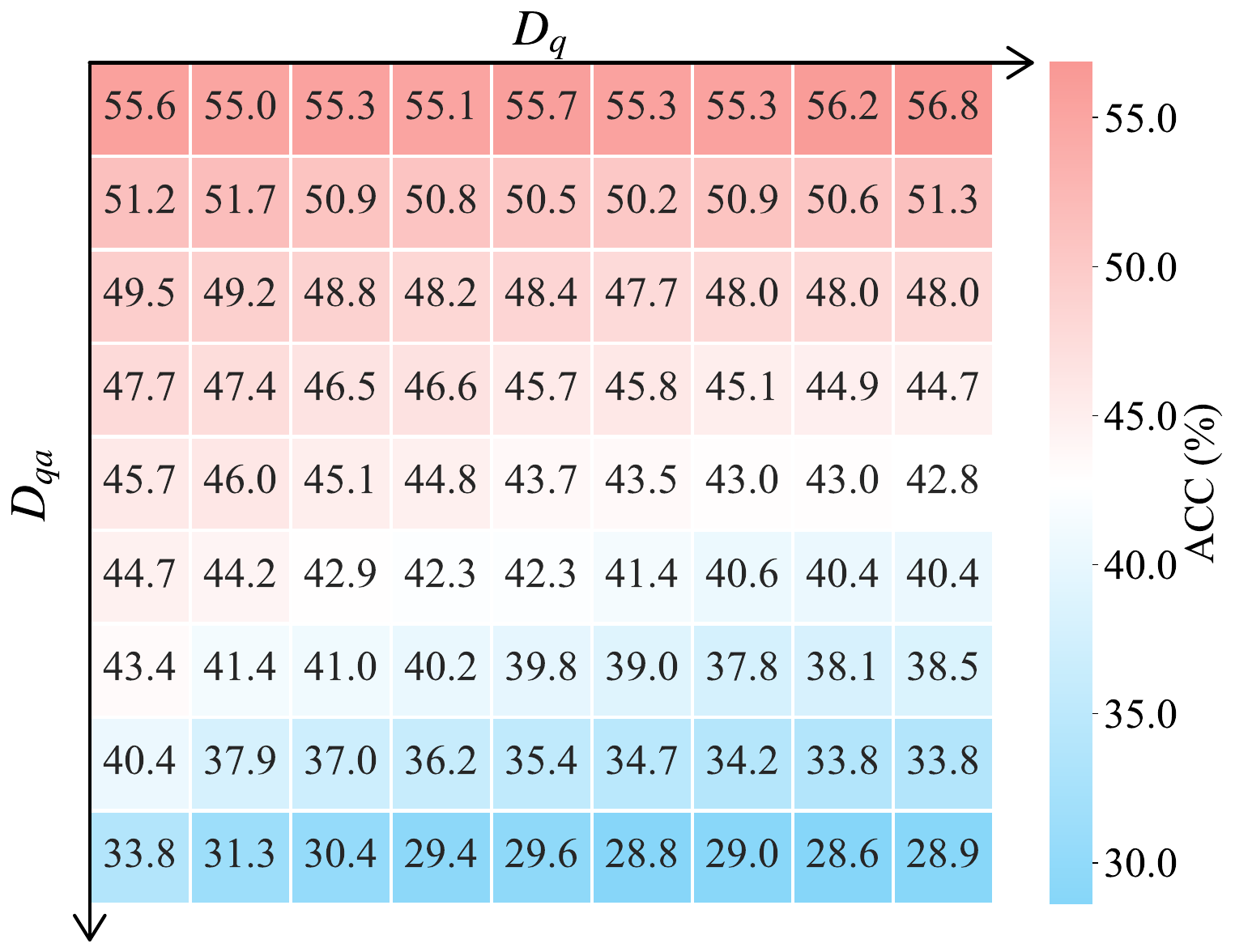} 
        \caption{$D_q$ and $D_{qa}$ }
        \label{fig:diff_factor2}
    \end{subfigure}
    \hfill
    \begin{subfigure}[t]{0.32\linewidth}
        \centering
        \includegraphics[width=\linewidth]{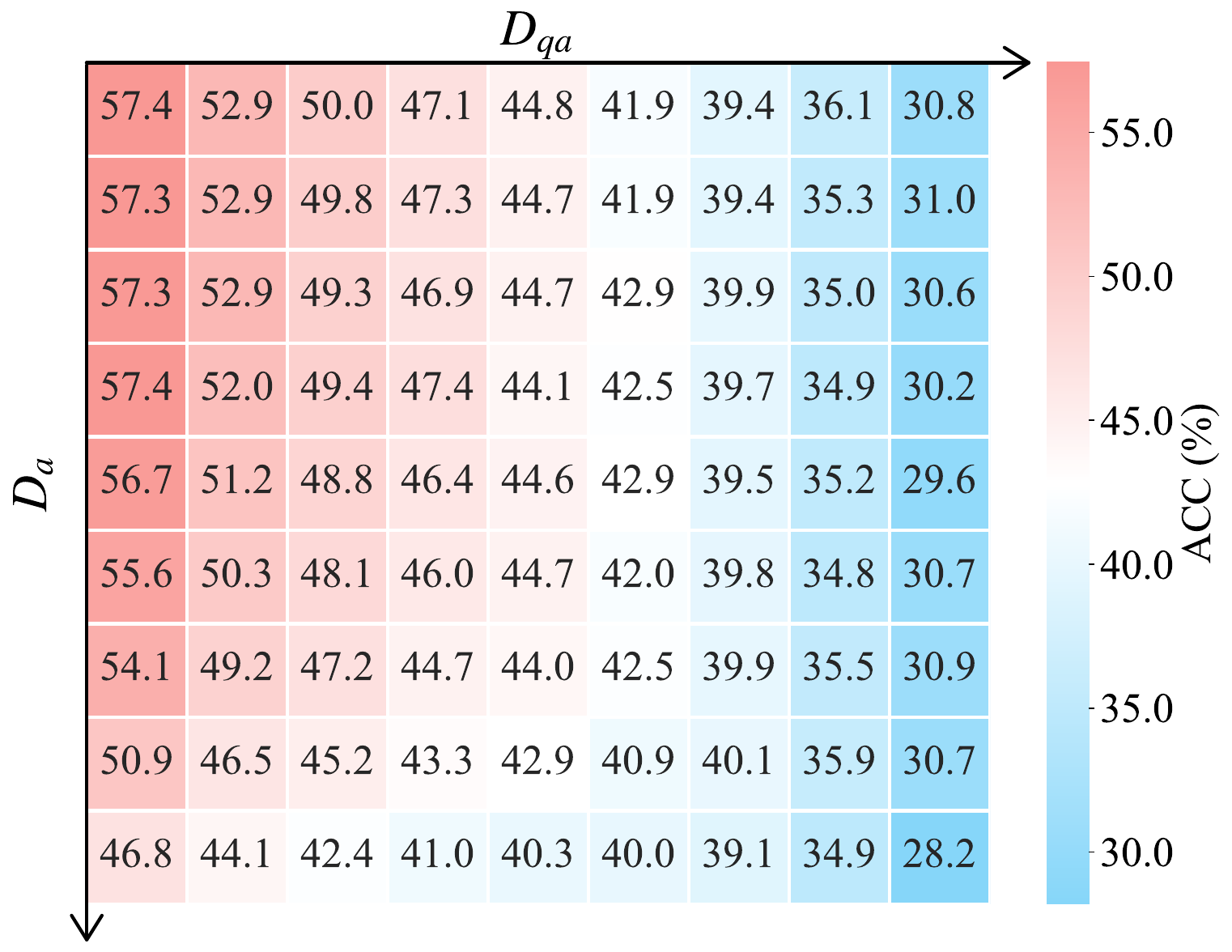} 
        \caption{$D_{qa}$ and $D_a$ }
        \label{fig:diff_factor3}
    \end{subfigure}
    }
    \caption{Average QA accuracy across different difficulty levels on StoryVideoQA.}
    \label{fig:diff_factor}
\end{figure*}

\subsubsection{Evaluations on StoryVideoQA}\label{sec:evaluation}
We conduct comprehensive evaluations of VLMs-based and MLLMs-based methods on the full StoryVideoQA dataset. Table \ref{tab:comparison} shows detailed results categorized by all 14 fine-grained topics.

\begin{figure}[!t]
    \centering
    \includegraphics[width=\linewidth]{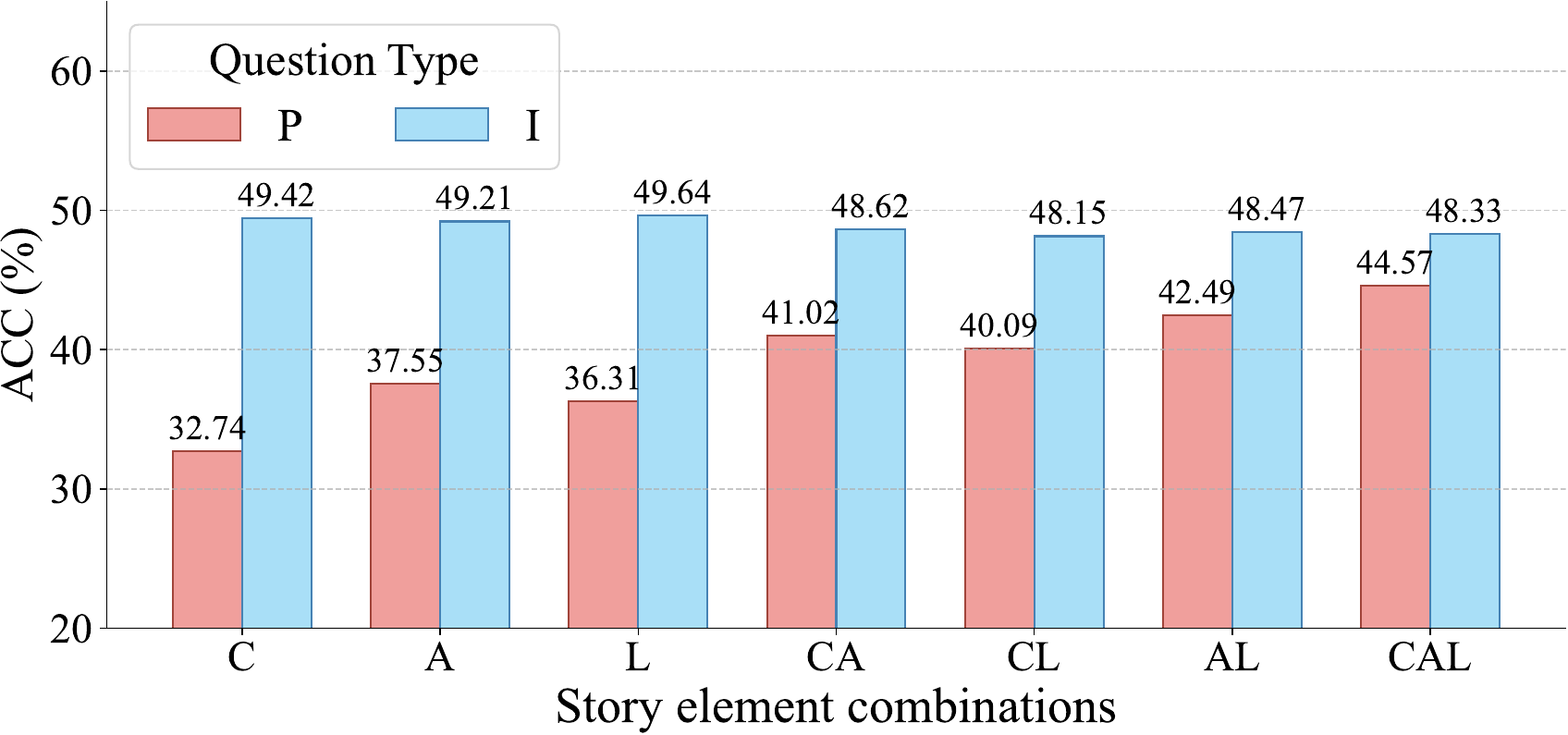}
    \caption{Average performance of VLMs and MLLMs on fine-grained topic.}
    \label{fig:acc_fgtopic}
\end{figure}

\vspace{2mm}
\noindent \textbf{VLMs vs. MLLMs}.
The results in Table \ref{tab:comparison} reveal a distinct performance gap between the two main architectural paradigms, MLLMs-based methods achieve up to 76.32\% accuracy, while traditional VLMs-based methods remain below 23\%. Across fine-grained topics, VLMs-based methods perform poorly on both perception and inference QAs (none exceeding 30\%), whereas MLLMs benefit from pre-trained knowledge and show clear gains on inference QAs. The advantage of MLLMs mainly stems from integrating LLMs as their core, which brings stronger language understanding and enhanced reasoning ability.

\begin{figure}[!t]
    \centering
    \includegraphics[width=\linewidth]{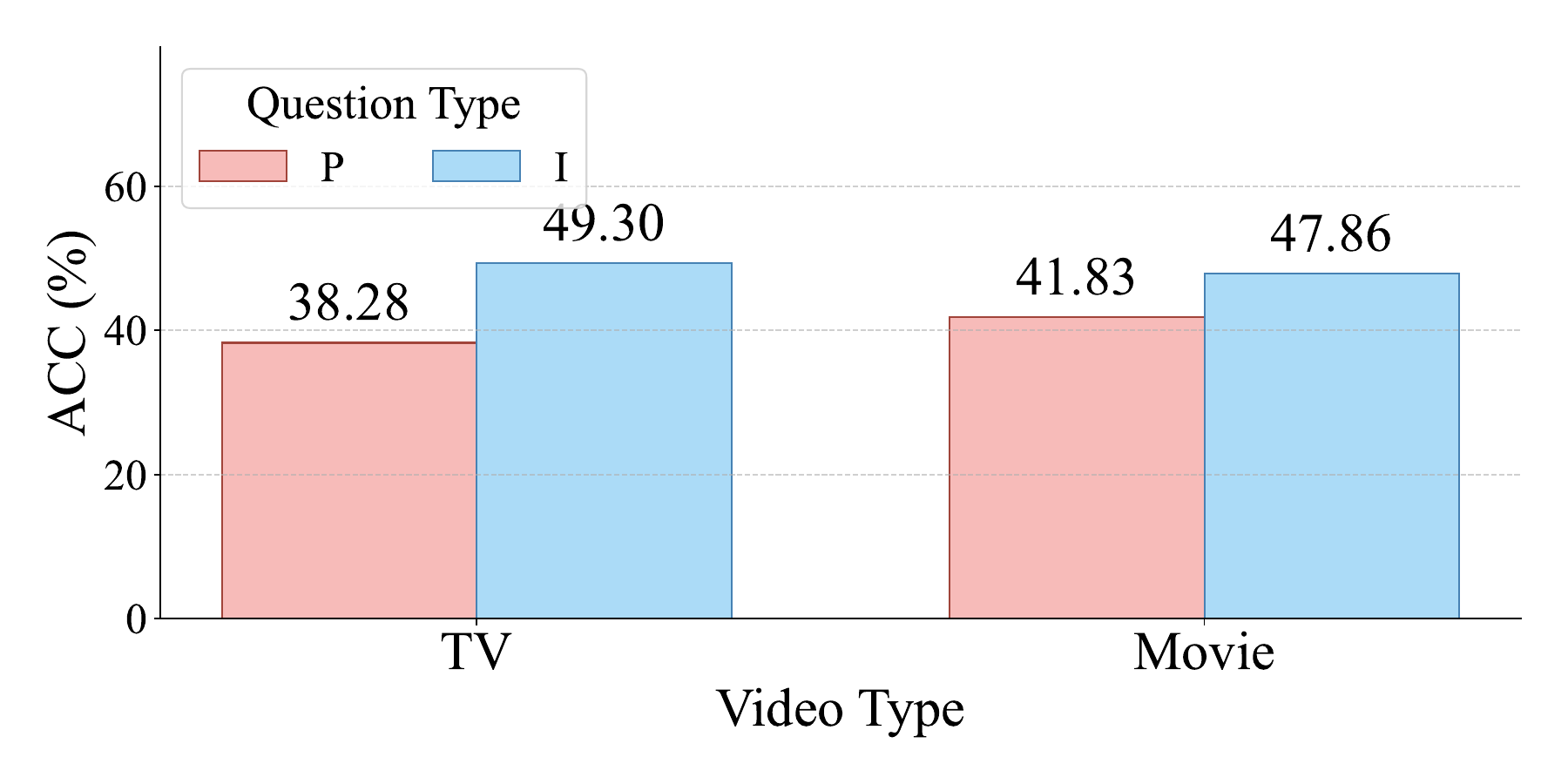}
    \caption{Average performance analysis across different video types and question types.}
    \label{fig:tv_movie_diff}
\end{figure}

\vspace{2mm}
\noindent \textbf{Fine-grained Topic Analysis}.  
For the average performance of VLMs and MLLMs on fine-grained topic (Figure \ref{fig:acc_fgtopic}), we find perception QAs show notable performance degradation on single story element QAs compared to those of composite story elements. Take P-CAL as an example, the average performance drops are substantial: 11.83\% for character (P-C), 7.02\% for action (P-A), and 8.26\% for location (P-L). Conversely, performance on inference QAs does not exhibit such a clear trend. This result is highly consistent with our finding in Figure \ref{fig:diff_dis}(b).

\vspace{2mm}
\noindent \textbf{Video Type}. Furthermore, we compare performance across two video types (TV and Movie) and two question types (perception and inference). For perception QAs, performance is lower on TV (38.28\%) than movies (41.83\%), while for inference QAs, performance is lower on movies (47.86\%) than TV videos (49.30\%), as illustrated in Figure \ref{fig:tv_movie_diff}. This suggests that despite MLLMs' strong prior knowledge, they encounter significant challenges when performing long-range inference on the more complex and long contexts movies (Refer to Appendix E for more experiments of disentangling priority knowledge).

\vspace{2mm}
\noindent \textbf{Difficulty Influence}.  
To validate our difficulty metrics, we analyze methods' accuracy across discretized bins of question ($D_q$), answer ($D_a$), and question-answer difficulty ($D_{qa}$) on the full StoryVideoQA dataset. As shown in Figures \ref{fig:diff_factor}, average performance declines as difficulty increases, confirming the reliability of our metrics.

\begin{figure}[!t]
    \centering
    \includegraphics[width=\linewidth]{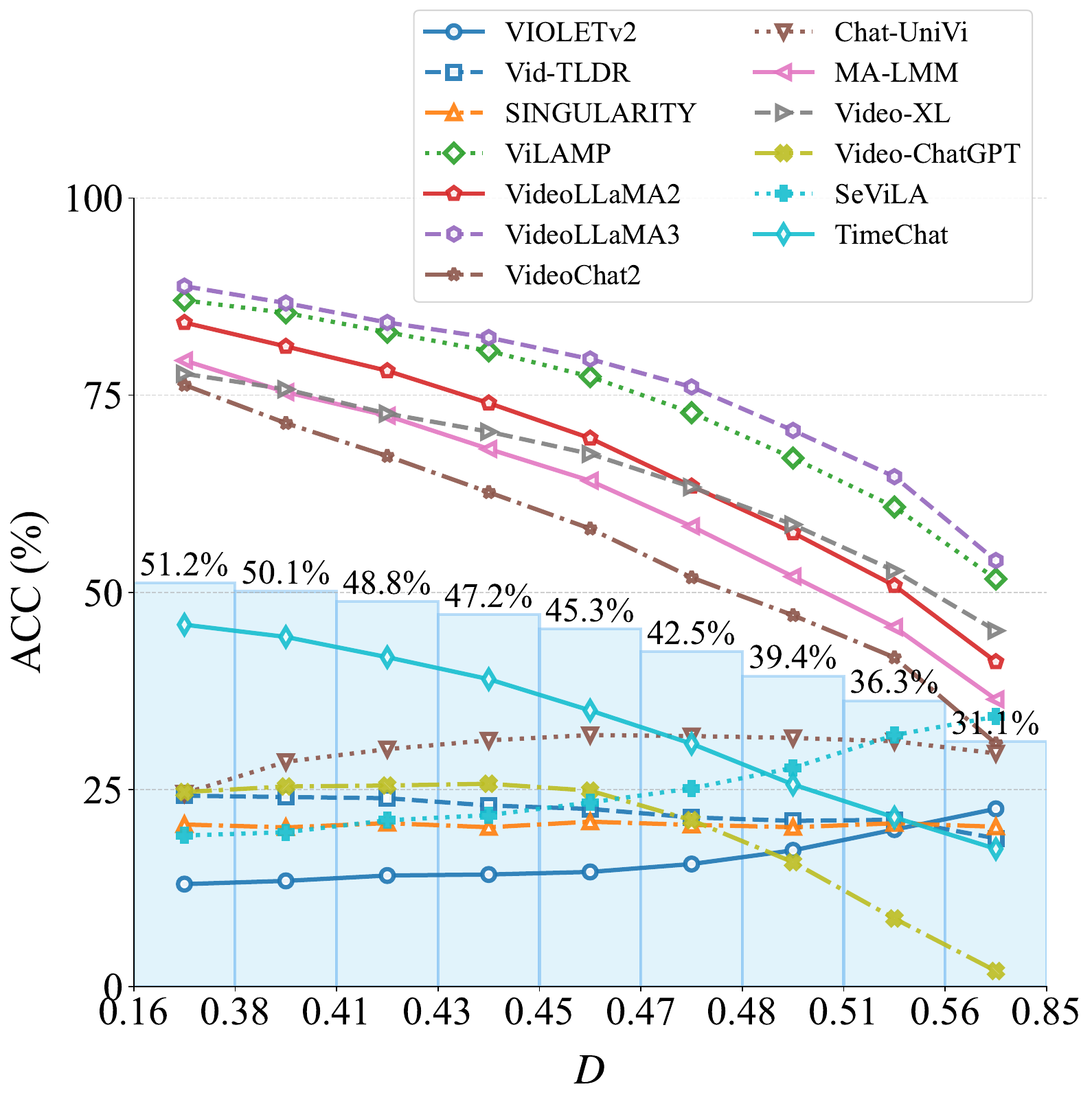}
    \caption{Average (bars) and individual performance (curves) analysis of $D$ on StoryVideoQA.}
    \label{fig:diff_full}
\end{figure}

\begin{table*}[t]
\caption{Comparisons (\%) on StoryVideoQA-G across 14 fine-grained topics. The baseline marked with * is implemented by us due to the lack of official code.}
\label{tab:subcompare}
\resizebox{\textwidth}{!}{%
\begin{tabular}{cc|rrrrrrr|rrrrrrr|c}
\toprule
                      &                    & \multicolumn{7}{c|}{{P}}                                                                                                                                                                                                                                                                                                                                                                                                                                                                                                                                                     & \multicolumn{7}{c|}{{I}}                                                                                                                                                                                                                                                                                                                                                                                                                                                                                                                                                              & {}                                                           \\
        \cmidrule{3-9}\cmidrule{10-16}
        \multirow{-2}{*}{\textbf{Method}} & \multirow{-2}{*}{\textbf{Venue}}  & {C}                                                          & {A}                                                          & {L}                                                          & {CA}                                                         & {CL}                                                         & {AL}                                                         & {CAL}                                                        & {C}                                                          & {A}                                                          & {L}                                                          & {CA}                                                         & {CL}                                                         & {AL}                                                         & {CAL}                                                        & \multirow{-2}{*}{\textbf{Avg.}}                                    \\
        \midrule\midrule
        \multicolumn{17}{c}{\cellcolor[HTML]{D1EBFF} \textit{VLMs}}  \\
        SINGULARITY \cite{Singularity}           & ACL'23                                    & 22.63                          & 20.07                         & 18.45                         & 21.36                  & 18.73                  & 21.89                  & 19.76                   & 21.63                 & 21.00                 & 20.91                 & 25.11                  & 18.73                  & 17.00                  & 17.70                   & 20.44                                      \\
        VIOLETv2 \cite{violetv2}                 & CVPR'23                                   & 18.83                          & 14.97                         & 18.28                         & 16.64                  & 19.48                  & 18.53                  & 16.33                   & 12.77                 & 12.12                 & 15.19                 & 11.09                  & 13.16                  & 18.50                  & 12.92                   & 15.78                                      \\
        Vid-TLDR \cite{vid-tldr}                 & CVPR'24                                   & 18.83                          & 20.23                         & 22.07                         & 20.04                  & 26.59                  & 23.79                  & 25.60                   & 22.34                 & 20.56                 & 24.46                 & 18.33                  & 24.30                  & 25.50                  & 22.19                   & 22.39                                      \\
        \midrule
        \multicolumn{17}{c}{\cellcolor[HTML]{D1EBFF} \textit{MLLMs}}  \\
        SeViLA \cite{sevila}                     & CVPR'23                                   & 27.12                          & 22.70                         & 32.07                         & 20.98                  & 32.40                  & 28.63                  & 20.97                   & 22.70                 & 18.61                 & 23.67                 & 17.19                  & 19.49                  & 21.25                  & 17.42                   & 23.66                                      \\
        VideoLLaMA2 \cite{videollama2}           & ArXiv'24                                  & 48.36                          & 52.47                         & 58.45                         & 62.76                  & 63.48                  & 68.63                  & 77.62                   & 79.08                 & 79.22                 & 83.63                 & 79.41                  & 83.04                  & 82.25                  & 82.58                   & 70.13                                      \\
        VideoChat2 \cite{mvbench}                & CVPR'24                                  & 38.00                          & 44.74                         & 51.90                         & 48.96                  & 54.12                  & 61.05                  & 72.78                   & 67.20                 & 63.20                 & 70.22                 & 64.03                  & 68.35                  & 70.50                  & 69.94                   & 59.23                                      \\
        Chat-UniVi \cite{chat-univi}              & CVPR'24                                   & 31.78                          & 40.79                         & 25.34                         & 41.40                  & 25.66                  & 38.95                  & 31.85                   & 30.32                 & 35.06                 & 19.33                 & 28.05                  & 21.27                  & 31.25                  & 23.88                   & 30.71                                      \\
        MA-LMM \cite{malmm}                       & CVPR'24                                   & 46.98                          & 49.18                         & 54.31                         & 53.88                  & 58.80                  & 65.89                  & 67.54                   & 75.89                 & 70.78                 & 74.75                 & 73.30                  & 78.23                  & 76.25                  & 77.53                   & 64.69                                      \\
        TimeChat \cite{timechat}                 & CVPR'24                                   & 24.35                          & 21.05                         & 37.76                         & 28.36                  & 41.01                  & 33.05                  & 31.25                   & 44.68                 & 43.51                 & 52.27                 & 42.08                  & 47.34                  & 43.00                  & 43.82                   & 37.36                                      \\
        Video-ChatGPT \cite{videochatgpt}         & ACL'24                                    & 9.33                           & 18.26                         & 5.34                          & 19.66                  & 8.80                   & 14.32                  & 14.31                   & 28.55                 & 29.87                 & 26.63                 & 26.92                  & 23.29                  & 28.00                  & 19.66                   & 18.95                                      \\
        Video-XL \cite{videoxl}                  & CVPR'25                                   & 51.30                          & 61.84                         & 60.17                         & 66.16                  & 60.86                  & 69.05                  & 70.77                   & 73.76                 & 72.29                 & 77.71                 & 73.76                  & 80.51                  & 75.00                  & 78.93                   & 68.50                                      \\
        ViLAMP \cite{vilamp}                     & ICML'25                                  & 58.55                          & 69.08                         & 64.83                         & 74.29                  & 72.28                  & 77.47                  & 77.62                   & 87.41                 & 85.28                 & 85.80                 & 82.35                  & 87.59                  & 87.50                  & 86.52                   & 77.34                                      \\
        VideoLLaMA3 \cite{videollama3}           & ArXiv'25                         & 60.45                          & 72.04                         & 68.62                         & 79.02                  & 72.66                  & 82.53                  & 86.29                   & 86.88                 & 85.28                 & 88.56                 & 86.43                  & 89.11                  & 88.25                  & 88.76                   & 80.09                                      \\
        VideoChat-Flash \cite{videochat-flash}     & ArXiv'25                         & 60.45 & 71.88 & 68.79 & 76.37 & 76.40 & 80.00 & 85.89 & 87.77 & 83.77 & 89.74 & 84.62 & 90.89 & 88.50 & 88.20 & 80.01 \\
        Long-VITA \cite{Long-vita}                 & ArXiv'25                         & 63.39 & 71.38 & 66.21 & 73.91 & 67.42 & 77.26 & 82.46 & 73.76 & 75.32 & 70.41 & 76.92 & 76.20 & 80.25 & 83.71 & 73.52 \\
        Qwen3-VL \cite{qwen3vltechnicalreport}     & ArXiv'25                         & 58.03 & 68.91 & 62.59 & 76.56 & 68.16 & 77.89 & 83.06 & 86.52 & 87.01 & 87.38 & 88.69 & 91.14 & 89.50 & 91.01 & 78.48 \\
        \midrule
        \multicolumn{17}{c}{\cellcolor[HTML]{D1EBFF} \textit{Frontier MLLMs}}     \\
        \rev{Gemini-3-Flash}      & \rev{Google}                           & \rev{85.32}                  & \rev{84.21}                  & \rev{84.14}                  & \rev{88.09}                  & \rev{85.77}                  & \rev{83.37}                  & \rev{87.70}                  & \rev{93.79}                  & \rev{88.53}                  & \rev{88.95}                  & \rev{88.91}                  & \rev{\mybold{93.16}}                  & \rev{92.75}                  & \rev{90.73}                  & \rev{87.96}                  \\
        \rev{GPT-5.2}             & \rev{OpenAI}                          & \rev{73.23} & \rev{73.85} & \rev{80.17} & \rev{84.88} & \rev{78.28} & \rev{82.11} & \rev{89.11} & \rev{92.91} & \rev{90.04} & \rev{\mybold{93.29}} & \rev{\mybold{90.72}} & \rev{91.39} & \rev{92.75} & \rev{92.98} & \rev{85.38} \\
        \midrule
        \multicolumn{17}{c}{\cellcolor[HTML]{D1EBFF} \textit{Agents (Powered by Gemini-2.0-Flash)}}     \\
        {Video2RAG* \cite{omagent}}        &    {EMNLP'24}                           & 77.72                          & 71.05                         & 73.45                         & 80.34                  & 79.59                  & 78.11                  & 85.89                   & 90.78                 & 91.99                 & 91.12                 & 89.37                  & 90.13                  & 90.50                  & 91.57                   & 83.63                                      \\
        {VideoTree  \cite{videotree}}      &    {CVPR'25}                           & 56.30                          & 53.78                         & 56.55                         & 65.78                  & 61.24                  & 66.74                  & 73.19                   & 86.17                 & 81.60                 & 86.39                 & 82.13                  & 86.08                  & 85.75                  & 85.96                   & 72.02                                      \\
        {PlotTree}       &    {Ours}                           & {83.07} & {75.66} & {78.28} & {82.80} & {85.58} & {81.89} & {87.30} & {91.67} & \mybold{92.64} & {92.11} & {90.50} & {91.90} & \mybold{93.25} & \mybold{93.26} & {86.50}                          \\
        \midrule
        \multicolumn{17}{c}{\cellcolor[HTML]{D1EBFF} \textit{Agents (Powered by Gemini-3-Flash)}}     \\          
        
        \rev{Video2RAG* \cite{omagent}}         &    \rev{EMNLP'24}                       & \rev{85.84} & \rev{86.02} & \rev{86.03} & \rev{86.77} & \rev{86.33} & \rev{83.79} & \rev{81.85} & \rev{90.78} & \rev{83.77} & \rev{86.39} & \rev{84.16} & \rev{88.35} & \rev{90.75} & \rev{87.08} & \rev{86.24} \\
        \rev{VideoTree  \cite{videotree}}      &    \rev{CVPR'25}                        & \rev{81.87} & \rev{83.72} & \rev{79.31} & \rev{\mybold{89.22}} & \rev{81.84} & \rev{84.84} & \rev{\mybold{90.52}} & \rev{93.62} & \rev{88.31} & \rev{90.14} & \rev{87.56} & \rev{88.86} & \rev{93.00} & \rev{89.33} & \rev{86.98} \\
        \rev{PlotTree}                         &    \rev{Ours}                          & \rev{\mybold{89.12}} & \rev{\mybold{88.32}} & \rev{\mybold{88.28}} & \rev{\mybold{89.22}} & \rev{\mybold{87.45}} & \rev{\mybold{85.89}} & \rev{89.31} & \rev{\mybold{94.86}} & \rev{87.23} & \rev{89.35} & \rev{86.88} & \rev{88.61} & \rev{91.75} & \rev{85.39} & \rev{\mybold{88.80}} \\\bottomrule

\end{tabular}%
}
\end{table*}

In addition, we further analyze methods' performance across different difficulty levels by dividing the total difficulty score $D$ into 9 groups, each containing an equal number of QAs. The average performance of VLMs-based and MLLMs-based methods within each group is shown as blue bars, while the continuous curves depict individual method's performance (Figure \ref{fig:diff_full}).
Both average and most individual methods' performance declines as difficulty $D$ increases (a few exceptions, e.g., VIOLETv2, with near random performance). This trend confirms that our proposed difficulty measure effectively distinguishes QAs of varying challenge levels in the StoryVideoQA, enabling researchers to better understand how QAs difficulty impacts methods' performance and to develop more targeted improvements.

\subsubsection{Evaluations on StoryVideoQA-G}\label{sec:plottree_exp}
In this section, we conduct more extensive experiments on the manually-labeled, high-quality golden subset StoryVideoQA-G across three categories of methods: VLMs, MLLMs, and agents including PlotTree, as illustrated in Table \ref{tab:subcompare}.

\begin{table*}[h]
    \centering
    \caption{Effect (\%) of character identification on StoryVideoQA-G, Char. stands for character identification.}
    \label{tab:character_effect}
    \resizebox{\textwidth}{!}{%
    \begin{tabular}{ccc|rrrrrrr|rrrrrrr|c}
    \toprule
                &        &              & \multicolumn{7}{c|}{{P}}                                                                                                                                                                                                                                                                                                                                                                                                                                                                                                                                                     & \multicolumn{7}{c|}{{I}}                                                                                                                                                                                                                                                                                                                                                                                                                                                                                                                                                              & {}                                                           \\
    \cmidrule{4-10}\cmidrule{11-17}
    \multirow{-2}{*}{\textbf{Method}} & \multirow{-2}{*}{\textbf{Venue}} & \multirow{-2}{*}{\textbf{Char.}}  & {C}                                                          & {A}                                                          & {L}                                                          & {CA}                                                         & {CL}                                                         & {AL}                                                         & {CAL}                                                        & {C}                                                          & {A}                                                          & {L}                                                          & {CA}                                                         & {CL}                                                         & {AL}                                                         & {CAL}                                                        & \multirow{-2}{*}{Avg.}                                    \\
    \midrule\midrule
                                    &                                & \textcolor{red}{\ding{55}}                         &71.16        & 66.61        & 68.62        & 72.78         & 73.03         & 71.79         & 82.26          & 89.89        & 89.39        & 89.74        & 88.46         & 90.38         & 91.25         & 92.70          & 80.22          \\
    \multirow{-2}{*}{Video2RAG}     &   \multirow{-2}{*}{EMNLP'24}   & \textcolor{green}{\ding{51}}                        & 77.72                          & 71.05                         & 73.45                         & 80.34                  & 79.59                  & 78.11                  & 85.89                   & 90.78                 & 91.99                 & 91.12                 & 89.37                  & 90.13                  & 90.50                  & 91.57                   & 83.63                                      \\\midrule
                                     &                               &\textcolor{red}{\ding{55}}               & 56.30                          & 53.78                         & 56.55                         & 65.78                  & 61.24                  & 66.74                  & 73.19                   & 86.17                 & 81.60                 & 86.39                 & 82.13                  & 86.08                  & 85.75                  & 85.96                   & 72.02                                      \\

    \multirow{-2}{*}{VideoTree}      &    \multirow{-2}{*}{CVPR'25}  & \textcolor{green}{\ding{51}}                        & 59.76                     & 52.96                     & 60.69                     & 69.38                     & 67.98                     & 69.26                     & 79.84                      & 88.83                     & 86.80                      & 90.53                     & 87.56                     & 91.39                     & 89.75                     & 90.17                      & 75.99 \\ \midrule
                &       &           \textcolor{red}{\ding{55}}               & 74.27        & 68.59        & 70.52        & 75.99         & 77.72         & 76.84         & 87.50          & 90.78        & 88.96        & 91.32        & 88.01         & 92.41         & 91.75         & 91.85          & 82.37           \\
    \multirow{-2}{*}{PlotTree}       &   \multirow{-2}{*}{Ours}  & \textcolor{green}{\ding{51}}                         & \mybold{83.07} & \mybold{75.66} & \mybold{78.28} & \mybold{82.80} & \mybold{85.58} & \mybold{81.89} & \mybold{87.30} & \mybold{91.67} & \mybold{92.64} & \mybold{92.11} & \mybold{90.50} & \mybold{91.90} & \mybold{93.25} & \mybold{93.26} & \mybold{86.50}                          \\ \bottomrule
    \end{tabular}%
    }
\end{table*}

\vspace{2mm}
\noindent \textbf{Agents vs. VLMs \& MLLMs}. \rev{ The agent-based paradigm exhibits distinct advantages over MLLMs and VLMs. When powered by Gemini-2.0-Flash, agents like PlotTree (86.50\%) and Video2RAG (83.63\%) already surpass large-scale pre-trained VLMs and MLLMs. However, the frontier MLLMs Gemini-3-Flash (87.96\%) outperforms all Gemini-2.0-powered agents, demonstrating the immense potential of its native multimodal understanding over raw video frames. To explore the upper bounds of the agent paradigm, we upgraded their backbones to Gemini-3-Flash. Although all agents improved, e.g., VideoTree gains 14.96\% as the stronger Gemini-3-Flash compensates for its reasoning gaps. It most struggle to surpass the standalone Gemini-3-Flash because agents rely on transformed captions with inherent information loss. Remarkably, PlotTree (88.80\%) is the only agent that outperforms the frame-input Gemini-3-Flash, proving that a well-structured reasoning architecture can overcome the captioning bottleneck.
}

\vspace{2mm}
\noindent \textbf{PlotTree vs. other Agents}. \rev{PlotTree consistently outperforms both VideoTree and Video2RAG across all fine-grained topics under Gemini-2.0-Flash / Gemini-3-Flash setting (Table \ref{tab:subcompare}). 
Under the Gemini-2.0-Flash setting, PlotTree demonstrates significant superiority, outperforming VideoTree and Video2RAG by 14.48\% and 2.87\%, respectively. When upgrading to the Gemini-3-Flash frontier MLLMs, the performance gap between PlotTree and VideoTree narrows from 14.48\% to 1.82\%. It indicates that the enhanced reasoning and multimodal alignment of Gemini-3-Flash can partially offset the limitations of VideoTree's more generalized video abstraction. However, PlotTree (88.80\%) maintains its lead, proving that its specialized hierarchical plot modeling remains more effective for complex storylines' understanding than relying solely on the scaling of MLLMs backbone capacities. 
}

\rev{

\begin{figure}
     \centering
    \includegraphics[width=0.9\linewidth]{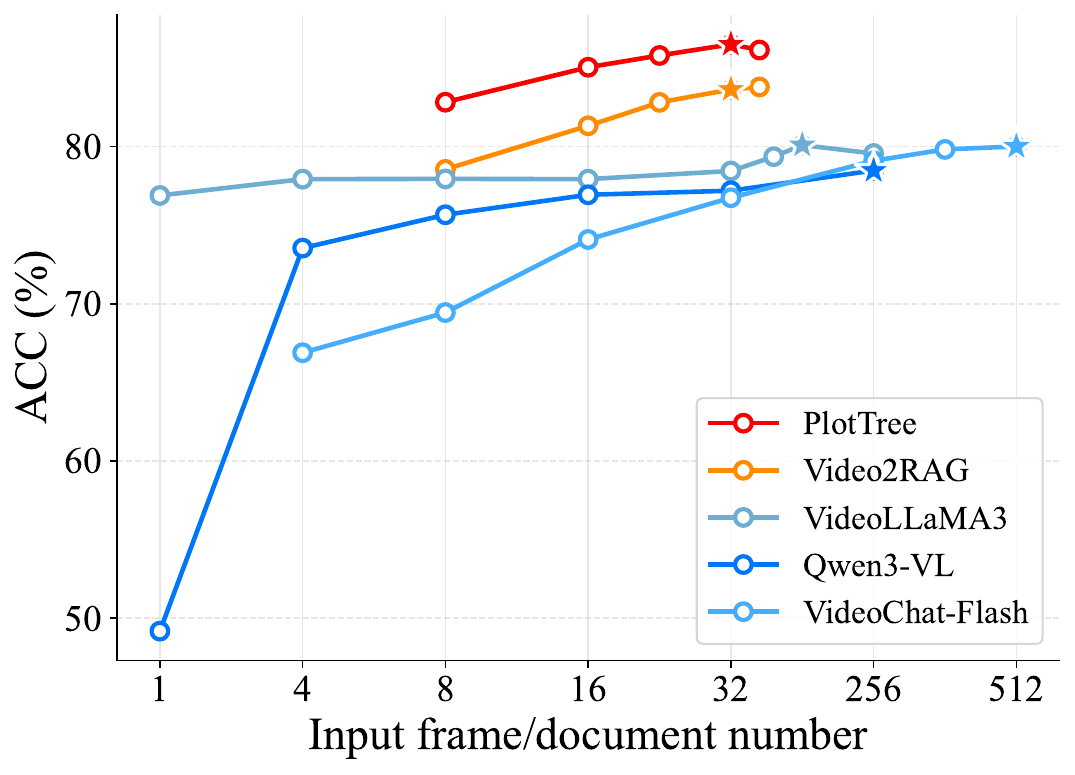}
    \caption{Dynamic performance of different models across varying input frame/document numbers. Stars ($\star$) indicate the maximum feasible/official configurations used in our main evaluations.}
    \label{fig:dynamic_frames}
\end{figure}

\begin{table*}[t]
    \centering
    \caption{Ablation Study (\%) for PlotTree on StoryVideoQA-G.}
    \label{tab:ablation}
    \resizebox{\textwidth}{!}{%
    \begin{tabular}{ccc|ccccccc|ccccccc|c}
    \toprule
    \multicolumn{3}{c|}{\textbf{Module}}              & \multicolumn{7}{c|}{{P}}                                                                                                                                                                                                                                                                                                                                                                                                                                                                                                                                                     & \multicolumn{7}{c|}{{I}}                                                                                                                                                                                                                                                                                                                                                                                                                                                                                                                                                              & {}                                                           \\
        \cmidrule{4-10}\cmidrule{11-17}
    \textbf{Ins.} & \textbf{Dia.} & \textbf{Tree}  & {C}                                                          & {A}                                                          & {L}                                                          & {CA}                                                         & {CL}                                                         & {AL}                                                         & {CAL}                                                        & {C}                                                          & {A}                                                          & {L}                                                          & {CA}                                                         & {CL}                                                         & {AL}                                                         & {CAL}                                                        & \multirow{-2}{*}{\textbf{Avg.}}                                    \\\midrule\midrule
    \textcolor{red}{\ding{55}}   & \textcolor{red}{\ding{55}}   & \textcolor{red}{\ding{55}}   & 71.16        & 66.61        & 68.62        & 72.78         & 73.03         & 71.79         & 82.26          & 89.89        & 89.39        & 89.74        & 88.46         & 90.38         & 91.25         & 92.70          & 80.22          \\
    \textcolor{green}{\ding{51}} & \textcolor{red}{\ding{55}}   & \textcolor{red}{\ding{55}}   & 78.58        & 68.75        & 72.59        & 79.40         & 80.71         & 77.89         & 86.49          & 90.78        & 89.18        & 91.91        & 89.59         & 91.39         & 91.25         & 93.54          & 83.57          \\
    \textcolor{red}{\ding{55}}   & \textcolor{green}{\ding{51}} & \textcolor{red}{\ding{55}}   & 80.48        & 68.26        & 73.97        & 80.91         & 80.52         & 75.37         & 85.48          & 90.78        & 92.21        & 90.34        & 89.59         & 92.15         & 92.00         & 92.70          & 83.79          \\
    \textcolor{red}{\ding{55}}   & \textcolor{red}{\ding{55}}   & \textcolor{green}{\ding{51}}& 74.27        & 68.59        & 70.52        & 75.99         & 77.72         & 76.84         & 87.50          & 90.78        & 88.96        & 91.32        & 88.01         & 92.41         & 91.75         & 91.85          & 82.37           \\\midrule
    \textcolor{green}{\ding{51}} & \textcolor{green}{\ding{51}} & \textcolor{red}{\ding{55}}   & 77.72        & 71.05        & 73.45        & 80.34         & 79.59         & 78.11         & 85.89          & 90.78        & 91.99        & 91.12        & 89.37         & 90.13         & 90.50         & 91.57          & 83.63          \\
    \textcolor{green}{\ding{51}} & \textcolor{red}{\ding{55}}   & \textcolor{green}{\ding{51}} & 83.42        & 75.33        & 78.45        & 82.99         & 83.15         & 80.42         & 88.10          & 92.02        & 90.91        & 92.50        & 88.24         & 90.38         & 91.50         & 94.94          & 86.00          \\
    \textcolor{red}{\ding{55}}   & \textcolor{green}{\ding{51}} & \textcolor{green}{\ding{51}} & 83.94        & 73.85        & 76.03        & 83.74         & 84.27         & 79.37         & 89.11          & 92.02        & 91.56        & 91.72        & 87.78         & 93.16         & 93.25         & 94.10          & 86.03          \\\midrule
    \textcolor{green}{\ding{51}} & \textcolor{green}{\ding{51}} & \textcolor{green}{\ding{51}} & 83.07        & 75.66        & 78.28        & 82.80         & 85.58         & 81.89         & 87.30          & 91.67        & 92.64        & 92.11        & 90.50         & 91.90         & 93.25         & 93.26          & \textbf{86.50} \\\bottomrule
    \end{tabular}%
    }
\end{table*}

\vspace{2mm}
\noindent \textbf{Characters' Effect}. To isolate the architectural gains from character identification, we ablate all agents methods with the same character module as PlotTree (i.e., plot captioning in PlotTree). As quantified in Table \ref{tab:character_effect}, the integration of explicit character identification consistently enhances performance across all models, with PlotTree exhibiting the most significant uplift of 4.13\% (82.37\% $\to$ 86.50\%) compared to VideoTree (+3.97\%) and Video2RAG (+3.41\%).  Moreover, PlotTree also outperforms baselines in the absence of character cues setting (82.37\% vs. 80.22\% for Video2RAG) and significantly broadens this gap when provided with identical character-level knowledge (Please refer to Appendix F for more robustness experiments of character recognition in PlotTree).

\begin{figure*}[t]
    \centering
    \resizebox{\linewidth}{!}{ 
    \begin{subfigure}[t]{0.32\linewidth}
        \centering
        \includegraphics[width=\linewidth]{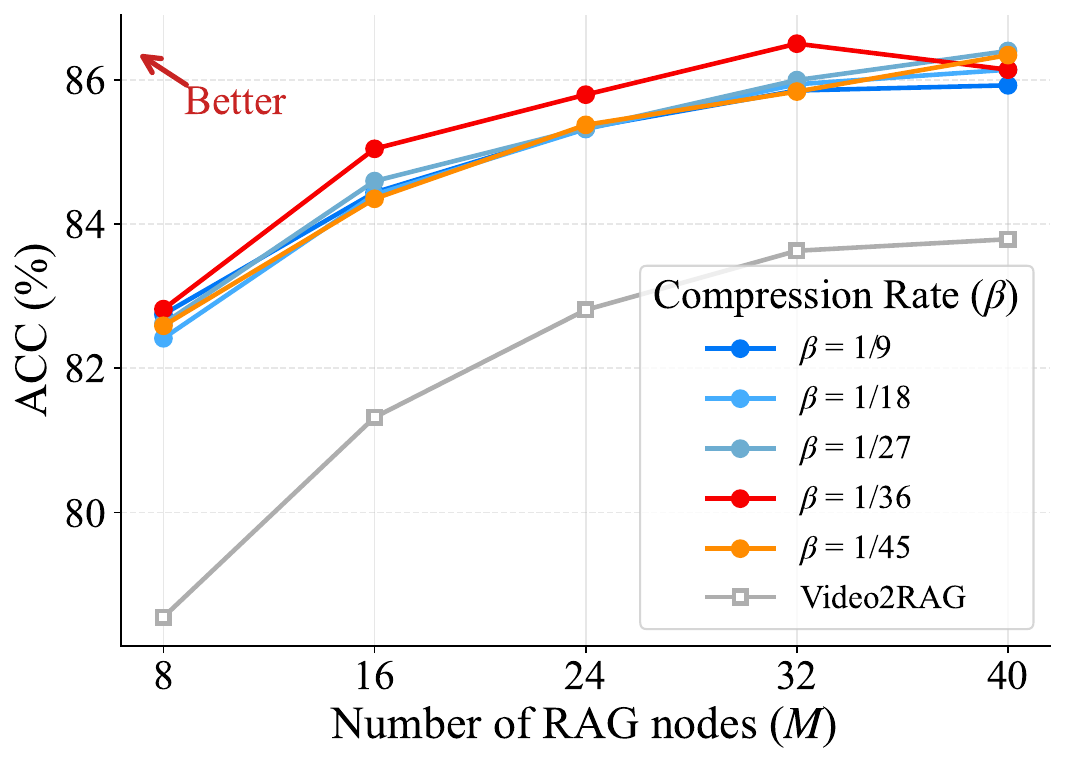} 
        \caption{$\beta$ and $M$ (while $\alpha$=10)}
        \label{fig:dynamic2}
    \end{subfigure}
    \hfill
    \begin{subfigure}[t]{0.32\linewidth}
        \centering
        \includegraphics[width=\linewidth]{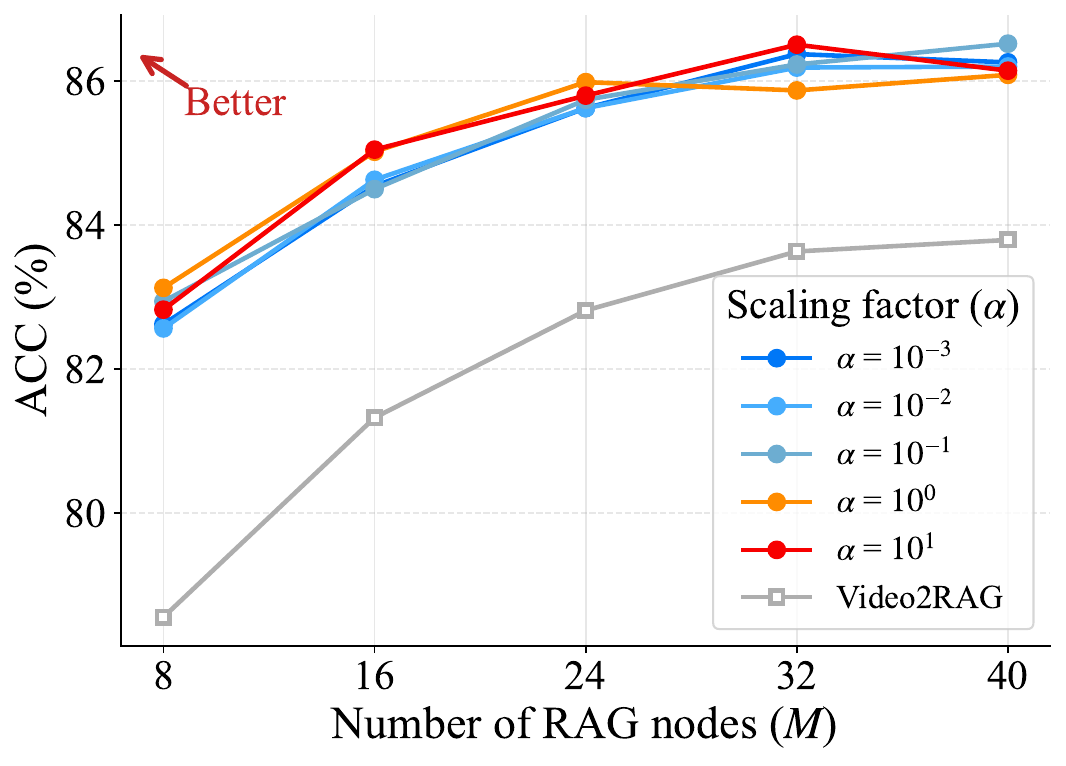} 
        \caption{$\alpha$ and $M$ (while $\beta$=1/36) }
        \label{fig:dynamic1}
    \end{subfigure}
    \hfill 
    \begin{subfigure}[t]{0.32\linewidth}
        \centering
        \includegraphics[width=\linewidth]{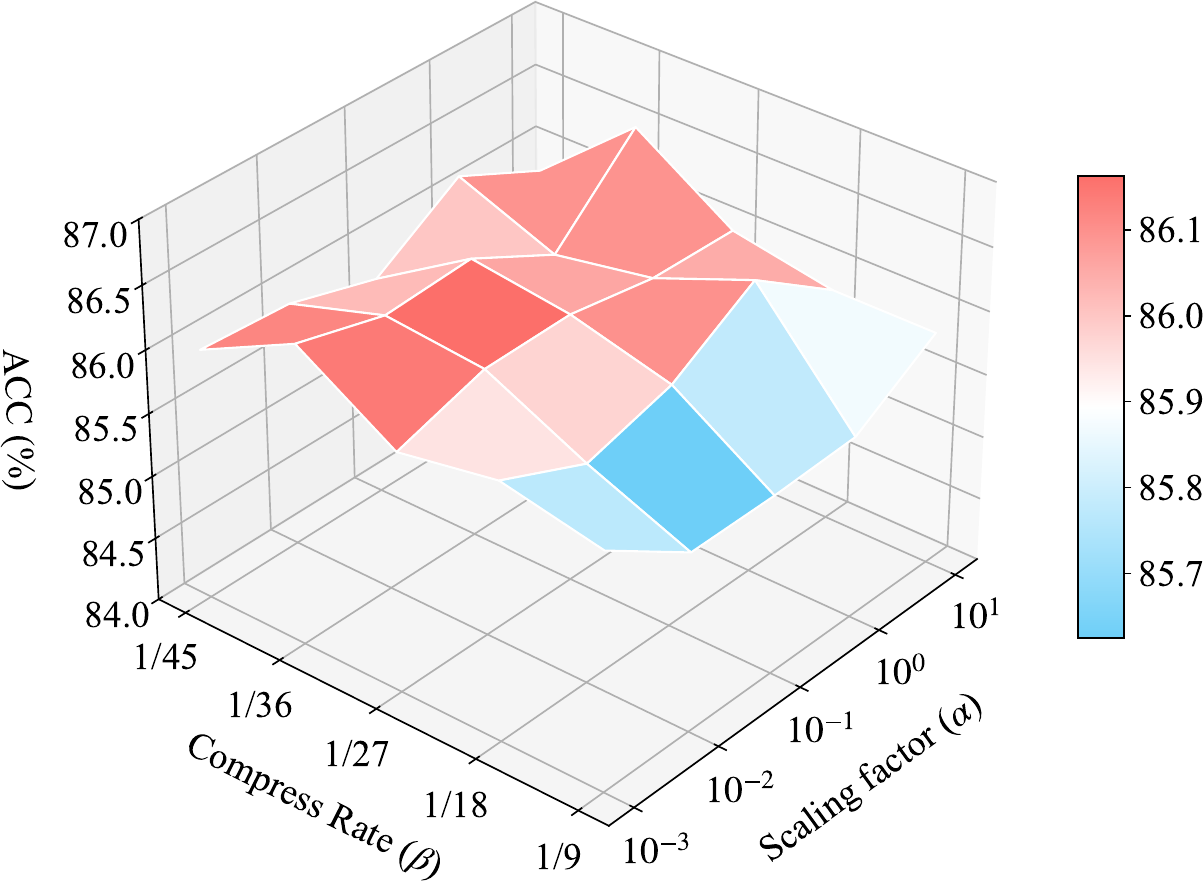} 
        \caption{$\alpha$ and $\beta$ (while $M$=32) }
        \label{fig:dynamic3}
    \end{subfigure}
    }
    \caption{Dynamic performance of different hyperparameters in PlotTree.}
    \label{fig:dynamic}
\end{figure*}

\vspace{2mm}
\noindent \textbf{Impact of Frame/Document Number}. To understand the relationship between the number of input frames/documents and performance, we conduct a dynamic analysis on differnt frames/documents number. As shown in Figure \ref{fig:dynamic_frames} , performance generally improves as M increases, the model performance exhibits a consistent upward trend as the number of input frame/document increases. It is noteworthy that both the official default settings and our hardware-constrained configurations (e.g., 256 frames for Qwen3-VL) reside near the performance saturation point. Specifically, agents that retrieve documents via RAG reach their performance plateau with fewer documents; this suggests that for deep video understanding, agent-based reasoning relies more on the precision of key information retrieval rather than the mere accumulation of context.
}

\subsubsection{Studies on PlotTree}
This section mainly evaluate the PlotTree on StoryVideoQA-G, covering ablation study, dynamic performance and qualitative analysis.

\vspace{2mm}
\noindent \textbf{Ablation Study}. \rev{
To evaluate the individual contributions of our components, we conduct an ablation study on the Ins. (face recognition), Dia. (dialogue), and Tree (PlotTree architecture) modules (Table \ref{tab:ablation}). Results show that all three modules independently yield performance gains over the baseline, with the Tree architecture alone achieving 82.37\%. This highlights its inherent capability to maintain narrative logic and filter noise even without external character cues. While merging modules consistently further boosts performance, the synergy between character-level data (Ins. and Dia.) and the Tree structure produces the peak result of 86.50\%. This confirms that while perceptual tools provide raw data, PlotTree acts as a critical reasoning hub that leverages global context to rectify local recognition failures. }

\begin{figure*}[!t]
    \centering
    \includegraphics[width=\linewidth]{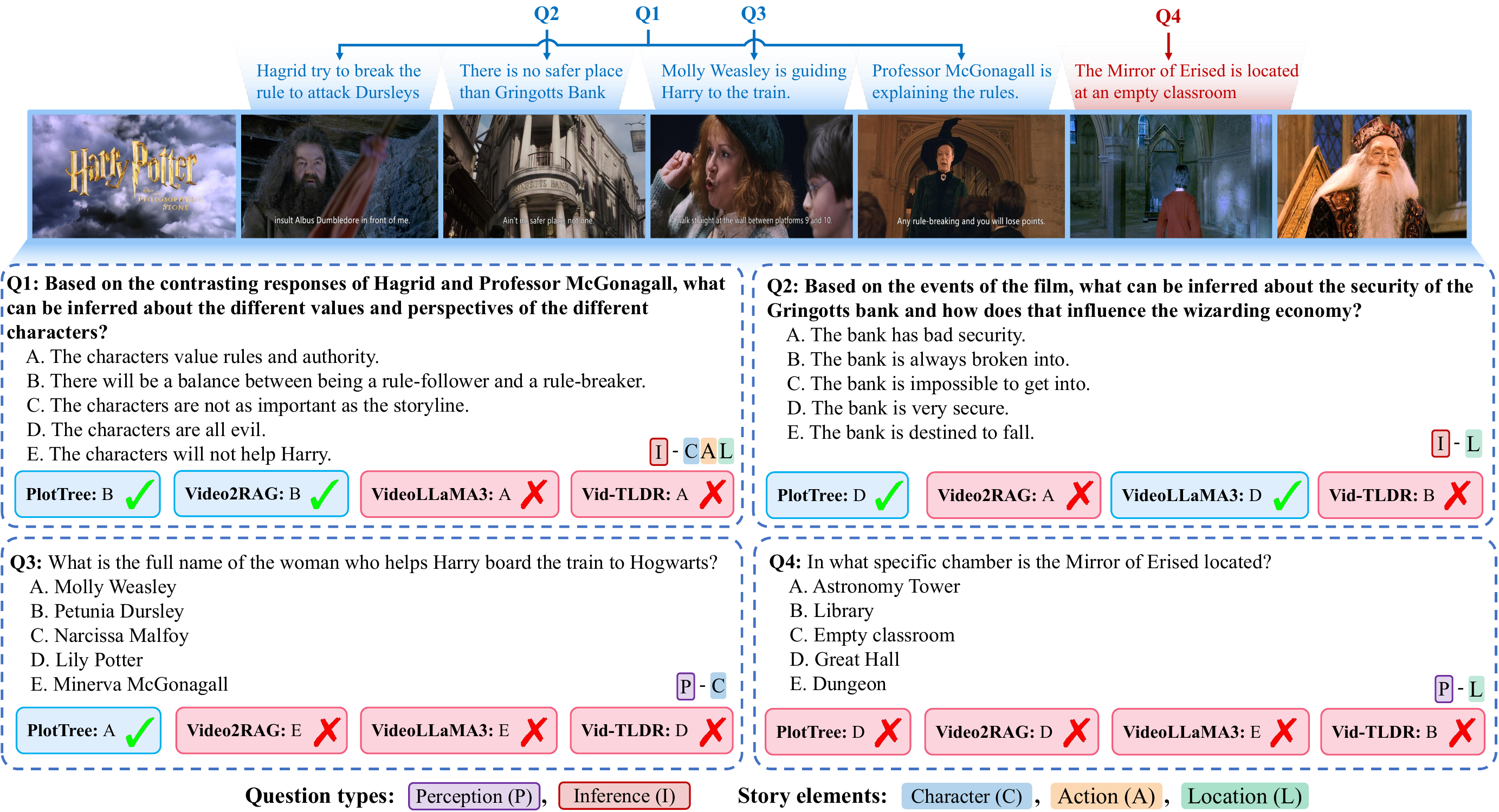}
    \caption{Qualitative study on StoryVideoQA-G.}
    \label{fig:qualitative}
\end{figure*}

\vspace{2mm}
\noindent \textbf{Dynamic Performance}.
We first analyze the impact of the number of RAG nodes $M$ on model performance. Under the same scaling factor (Figure \ref{fig:dynamic}(a)) and compression rate (Figure \ref{fig:dynamic}(b)), we evaluate PlotTree across different compression rates $\beta$ and scaling factors $\alpha$ between the number of RAG nodes $M$. As expected, expanding the range of RAG nodes allows PlotTree to access more information, thereby improving performance. However, when $M$ exceeds 32, the performance gain saturates and may even decline due to the introduction of irrelevant information. In addition, PlotTree consistently outperforms Video2RAG, demonstrating its superior robustness across various parameter settings.

Furthermore, we examine the influence of different scaling factors $\alpha$ and compression rates $\beta$ (Figure \ref{fig:dynamic}(c)). Larger decay factors $\alpha$ lead to better overall performance. Combined with the decay function illustrated in Figure \ref{fig:decay_func}, this indicates that lower-level (leaf) nodes should focus more on temporal consistency for visual understanding, while higher-level nodes should emphasize semantic similarity. Regarding compression rate $\beta$, PlotTree achieves optimal performance at a moderate level of compression. Overall, the best configuration is obtained with $\alpha$=10 and $\beta$=1/36.

\vspace{2mm}
\noindent \textbf{Qualitative Analysis}.
Beyond quantitative metrics, we also perform a qualitative comparison (Figure \ref{fig:qualitative}) between PlotTree and the best baseline from each method type in StoryVideoQA-G, namely Video2RAG in agents, VideoLLaMA3 in MLLMs, and Vid-TLDR in VLMs. 
The results highlight that PlotTree demonstrates stronger long-range reasoning ability, primarily owing to its hierarchical plot structure that preserves global narrative coherence. Conversely, other methods represent a video as a flat sequence of discrete events or visual embedding, struggling to capture the plot's long-range evolution and hierarchical structure (Q1-Q3 in Figure \ref{fig:qualitative}).
A shared limitation is in perception QAs, whici requires precise location understanding. The suboptimal performance of all methods on these questions indicates their lack of location recognition capability.

\section{Conclusion}

In this paper. we introduce StoryMindv2, an enhanced multi-agent framework featuring a novel supervisor-guided generation mechanism, a refined multi-reviewer voting strategy and a novel difficulty measure to evaluate question complexity, candidate answer divergence, and question-answer concordance. StoryMindv2 successfully enables high-quality, large-scale QA generation. Utilizing this framework, we construct StoryVideoQA, the largest and
most diverse dataset for DVU to date. It features over 363K QAs on 393.2 hours diverse,long-range story videos with balanced coverage across 14 fine-grained topics. We use this as a new benchmark to provide a comprehensive analysis of 20 SOTA methods. Finally, we propose PlotTree, which uses a hierarchical plot structure
for efficient comprehension, achieves superior performance in comprehending the long-range evolution of storylines.

\noindent \textbf{CRediT authorship contribution statement}
\\

\textbf{Zhengqian Wu}: Conceptualization, Methodology, Validation, Investigation, Data curation, Writing - original draft, Writing - review \& editing,
\textbf{Zhixian Liu}: Validation, Investigation, Data curation.
\textbf{Aodong Chen}: Validation, Investigation, Data curation.
\textbf{Jingyang Zhang}: Validation, Investigation, Data curation.
\textbf{Ruizhe Li}: Conceptualization, Validation, Investigation, Data curation.
\textbf{Hanlin Ge}: Investigation, Data curation.
\textbf{Zhongyuan Wang}: Investigation, Data curation, Funding acquisition.
\textbf{Chunxia Xiao}: Investigation, Data curation, Funding acquisition.
\textbf{Chao Liang}: Conceptualization, Methodology, Investigation, Data curation, Writing - original draft, Writing - review \& editing, Funding acquisition.

\quad

\noindent \textbf{Competing Interests}
\\

All authors certify that they have no affiliations with or involvement in any organization or entity with any financial interest or non-financial interest in the subject matter or materials discussed in this manuscript.

\quad

\noindent \textbf{Acknowledgements}
\\

This work is supported by the National Natural Science Foundation of China (No. 62372339, 62371350, and 62372336), Key Science and Technology Research Project of Xinjiang Production and Construction Corps (2025AB029), Hubei Provincial Science and Technology Plan Project (No. 2025BAB020, 2025CSA057) and the Ministry of Education Industry-University Cooperative Education Project (No. 240700006245501). The numerical calculations in this paper have been done on the supercomputing system in the Supercomputing Center of Wuhan University.

\quad

\noindent \textbf{Data Availability Statement }
\\

The authors confirm that the data supporting the findings of this study will be made publicly available upon acceptance of the manuscript. Specifically, this includes the StoryMindv2 dataset construction framework, the StoryVideoQA dataset, and the PlotTree video understanding method. Supplementary materials and instructions for access will be provided to ensure reproducibility.

\bibliography{storyvideoqa}

\setcounter{figure}{0} 
\renewcommand{\thefigure}{A\arabic{figure}} 

\setcounter{table}{0} 
\renewcommand{\thetable}{A\arabic{table}} 

\appendix 

\clearpage

\begin{center}
    {\LARGE\bfseries Appendix}\\
\end{center}

{\section{More Datasets Comparisons}}
\rev{
To further provide a comprehensive landscape of VideoQA, we compare our StoryVideoQA with both story-centric and general-purpose VideoQA datasets (detailed in Table \ref{tab:datasetComparisons}). 
Compared to general-purpose VideoQA datasets for long video such as Video-MME \cite{videomme}, LVBench \cite{lvbench}, and LongVideoBench \cite{LongVideoBench}, our dataset exhibits distinct characteristics: 
\begin{itemize}
    \item \textbf{Scale}: With 363K QAs and a density of 923.19 $h^{-1}$, StoryVideoQA is one to two orders of magnitude larger than recent general benchmarks (e.g., Video-MME, LongVideoBench).
    \item \textbf{Depth}: Unlike general-purpose datasets that include fragmented content (e.g., news, sports), StoryVideoQA focuses exclusively on structural story reasoning on story videos (TV series and movies). 
    \item \textbf{Difficulty}: StoryVideoQA is the a DVU dataset that incorporates a systematic difficulty measure, bridging the gap in existing benchmarks by enabling a granular analysis of model logic across varying complexity levels.
\end{itemize}
}

\begin{table*}[t]
    \centering
    \centering
    \caption{Comparisons of existing VideoQA datasets. Scale compares the number of QAs (\# QAs), the total length (Len.(h)) of all videos, the average duration (Dur.(s)) of videos, QAs density (Den.(h$^{-1}$)) in terms of (\# QAs)/(Len.) and dataset scale (Sca.(h)) in terms of (\# QAs)$\times$(Len.). Fine-grained topic considers the number of fine-grained topics exceeding 5\% of the dataset (\# Fin.) and the balance degree of fine-grained topic distribution. The Gini index (Gin.) and entropy (Ent.) are employed to measure the distribution's balance. The figures around the ``/'' corresponds to TV series and movies, respectively. }
    \label{tab:datasetComparisons}
    \begin{threeparttable} 
        \resizebox{\textwidth}{!}{%
            \begin{tabular}{ccrrcrrcrrcc}
                \toprule
                \multirow{2}{*}[-3pt]{\textbf{Dataset}} & \multirow{2}{*}[-3pt]{\textbf{Venue}} & \multicolumn{5}{c}{\textbf{Scale}}                                                                                           & \multicolumn{3}{c}{\textbf{Fine-grained topic}} & \multirow{2}{*}[-3pt]{\textbf{Type}} & \multirow{2}{*}[-3pt]{\textbf{\begin{tabular}[c]{@{}c@{}}Difficulty \\ measure\end{tabular}}} \\
                \cmidrule(lr){3-7} \cmidrule(lr){8-10}
                &                                 & \multicolumn{1}{l}{\textbf{Len. (h)}}    & \textbf{\# QAs}  & \textbf{Dur. (s)}    & \textbf{Den.(h$^{-1}$)} & \textbf{Sca. (h)} & \textbf{\# Fin.}     & \textbf{Gin.}     & \textbf{Ent.}     &                                &                                                                                         \\
                \midrule\midrule
                \multicolumn{12}{c}{\cellcolor[HTML]{D1EBFF} \textit{Story-centric VideoQA Datasets for story video}}     \\
                MovieQA \cite{movieqa}            & CVPR'16                         & 381.0         & ~14.9K                                   & 202.7                & 39.11                   & 5.68M             & 6                    & 0.819             & 2.713             & Movie                     & \textcolor{red}{\ding{55}}                                                                   \\
                TVQA \cite{tvqa}                  & EMNLP'18                        & 461.2       & 144.9K                                         & 76.2                 & 314.18                  & 66.83M            & 8                    & 0.821             & 2.873             & TV                    & \textcolor{red}{\ding{55}}                                                                    \\
                TVQA+ \cite{tvqaplus}             & ACL'20                          & 71.7        & 29.4K                                   & 61.5                 & 410.04                  & 2.11M             & 5                    & 0.789             & 2.660             & TV                           & \textcolor{red}{\ding{55}}                                                                \\
                HLVU (DVU 22\&23) \cite{hlvu}     & ICMR'20                         & 24.8        & 455                                 & 106/4,907            & 18.35                   & 0.01M             & 6                    & 0.773             & 2.548             & Movie                            & \textcolor{red}{\ding{55}}                                \\
                DramaQA \cite{dramaqa}            & AAAI'21                         & 20.5        & ~17.9K                                  & 3.6/91.8             & 873.17                  & 0.37M             & -                    & -                 & -                 & TV                           & \textcolor{green}{\ding{51}}                                  \\
                DeepMovieQA \cite{deepmaven}      & EACL'23                         & 41.3        & 1K                                 & 3,102                & 24.21                   & 0.04M             & -                    & -                 & -                 & Movie                             & \textcolor{red}{\ding{55}}                                                           \\
                CinePile \cite{cinepile}          & CVPRW'24                        & 417.6       & ~305K                                 & 160                  & 730.36                  & 127.37M           & -                    & -                 & -                 & Movie                          & \textcolor{red}{\ding{55}}                                                              \\
                MovieChat-1K \cite{moviechat}     & CVPR'24                         & 156.7       & 19.0K                                & 564                  & 121.25                  & 2.98M             & 4                    & 0.701             & 2.203             & Movie                           & \textcolor{red}{\ding{55}}                                                             \\
                LvBench \cite{zhang2025lvbench}    & IJCV'25                         & 209.5         & 20.0K                                   & 948                & 95.76                   & 4.20M             & -                    & -                    & -              & Movie                       & \textcolor{red}{\ding{55}}                                     \\
                \midrule
                \multicolumn{12}{c}{\cellcolor[HTML]{D1EBFF} \textit{General-purpose VideoQA Datasets for long video}}     \\
                LongVideoBench \cite{LongVideoBench}      & NeurIPS'24                         & 494.4         & 6,678                                   & 473                & 13.51                   & 3.30M             & -                    & -                    & -              & General long video$^*$         & \textcolor{red}{\ding{55}}                                                                               \\
                LVBench \cite{lvbench}            & ICCV'25                         & 117.0         & 1,549                                   & 4,101                & 13.24                   & 0.18M             & -                    & -                    & -              & General long video$^*$                  & \textcolor{red}{\ding{55}}                                                                      \\
                Video-MME \cite{videomme}            & CVPR'25                         & 254.5         & 2,700                                   & 1,017.9                & 10.61                   & 0.69M             & -                    & -                    & -              & General long video$^*$             & \textcolor{red}{\ding{55}}                                                                           \\
                CG-Bench \cite{CG-Bench}            & ICLR'25                         & 550.0         & 12,129                                   & 1,624.4                & 22.05                   & 6.67M             & -                    & -                    & -              & General long video$^*$             & \textcolor{red}{\ding{55}}                                                                           \\
                VRBench \cite{vrbench}            & arXiv'25                         & \textbf{1,545.6}         &  8,243                                  & 5,796                & 5.33                   & 12.74M             & -                    & -                    & -              & General long video$^*$      & \textcolor{red}{\ding{55}}                                                                                  \\
                Video-MMMU \cite{Video-mmmu}            & arXiv'25                        & 42.2         & 900                                   & 506.2                & 21.33                  & 0.04M             & -                    & -                    & -              & General long video$^*$                & \textcolor{red}{\ding{55}}                                                                       \\
                \midrule
                FriendsQA \cite{friendsqa25}      & AAAI'25                         & 89.6        & 44.6K                                  & 1,358                & 497.77                  & 4.00M             & \textbf{14}          & \textbf{0.927}    & 3.794             & TV                                                                                         \\
                StoryVideoQA                      & Ours                            & 393.2       & \textbf{363K}                                 & \textbf{1,635/7,878} & \textbf{923.19}         & \textbf{142.73M}  & \textbf{14}          & \textbf{0.927}    & \textbf{3.795}    & \textbf{TV/Movie}             & \textbf{\textcolor{green}{\ding{51}}}                                                  \\
                \bottomrule
            \end{tabular}
        }
        \begin{tablenotes}[flushleft]
            \scriptsize
            \item  \parbox{\textwidth}{$^*$ For general long video, LongVideoBench includes life, movie, knowledge, and news.
            LVBench features 6 types: sports, documentary, self media, life, TV, and cartoon. 
            Video-MME spans 6 domains including knowledge, film \& television, sports competition, life record, and Multilingual. 
            CG-Bench categorizes videos into 14 root domains including life record, arts, and news.
            VRBench focuses on narrative videos such as movies, sports, travelogues. 
            Video-MMMU covers professional educational videos in 6 disciplines like science and art.  
            }
        \end{tablenotes}
    \end{threeparttable} 
    \arrayrulecolor{black} 
\end{table*}

\section{Details of StoryMindv2}
In this section, we discuss more detailed implementation of StoryMindv2, including data source, alignment details, and the prompt for the generator, supervisor and reviewers.

\subsection{Data Source}
As illustrated in Table \ref{tab:datasource}, the StoryVideoQA dataset is constructed from a diverse set of story videos, covering both TV series and movies.
For TV series, it includes popular sitcoms such as \textit{Friends} series and \textit{The Big Bang Theory} series (first eight seasons), as well as the fantasy drama \textit{Game of Thrones} series.
For movies, we collected scripts from top-rated movies on both IMDB\footnote{https://www.imdb.com/} and Douban\footnote{https://www.douban.com/}, spanning a wide range of genres:
\begin{itemize}
\item Classic dramas and crime movies, such as The Godfather, The Shawshank Redemption, Pulp Fiction, and American Beauty.

\item Fantasy and adventure movies, including The Lord of the Rings trilogy, Harry Potter series, The Hobbit, Pirates of the Caribbean, and The Avengers.

\item Science fiction and action movies, such as Inception, The Matrix, The Dark Knight trilogy, Star Wars: Return of the Jedi, and Jurassic Park.

\item Animated and family movies, including Toy Story 3, The Lion King, Up, and How to Train Your Dragon.

\item Psychological thrillers and mysteries, such as Black Swan, Memento, Gone Girl, Vertigo, and Rear Window.
\end{itemize}

In addition to the scripts, we also collect portrait images of characters to support face recognition and character grounding in videos. Specifically, we incorporate the portraits provided by the PAINS dataset \cite{TVCSINS} for \textit{Friends} and \textit{The Big Bang Theory}. For \textit{Game of Thrones}, since no public face database is available, we manually crop a
library of 471 portrait photos for the 63 main
characters directly from the videos. Furthermore, for the remaining movies, we crawl actor portraits from IMDB, ensuring that each major character has a corresponding visual reference.

\begin{figure*}[!t]
    \centering
    \includegraphics[width=\linewidth]{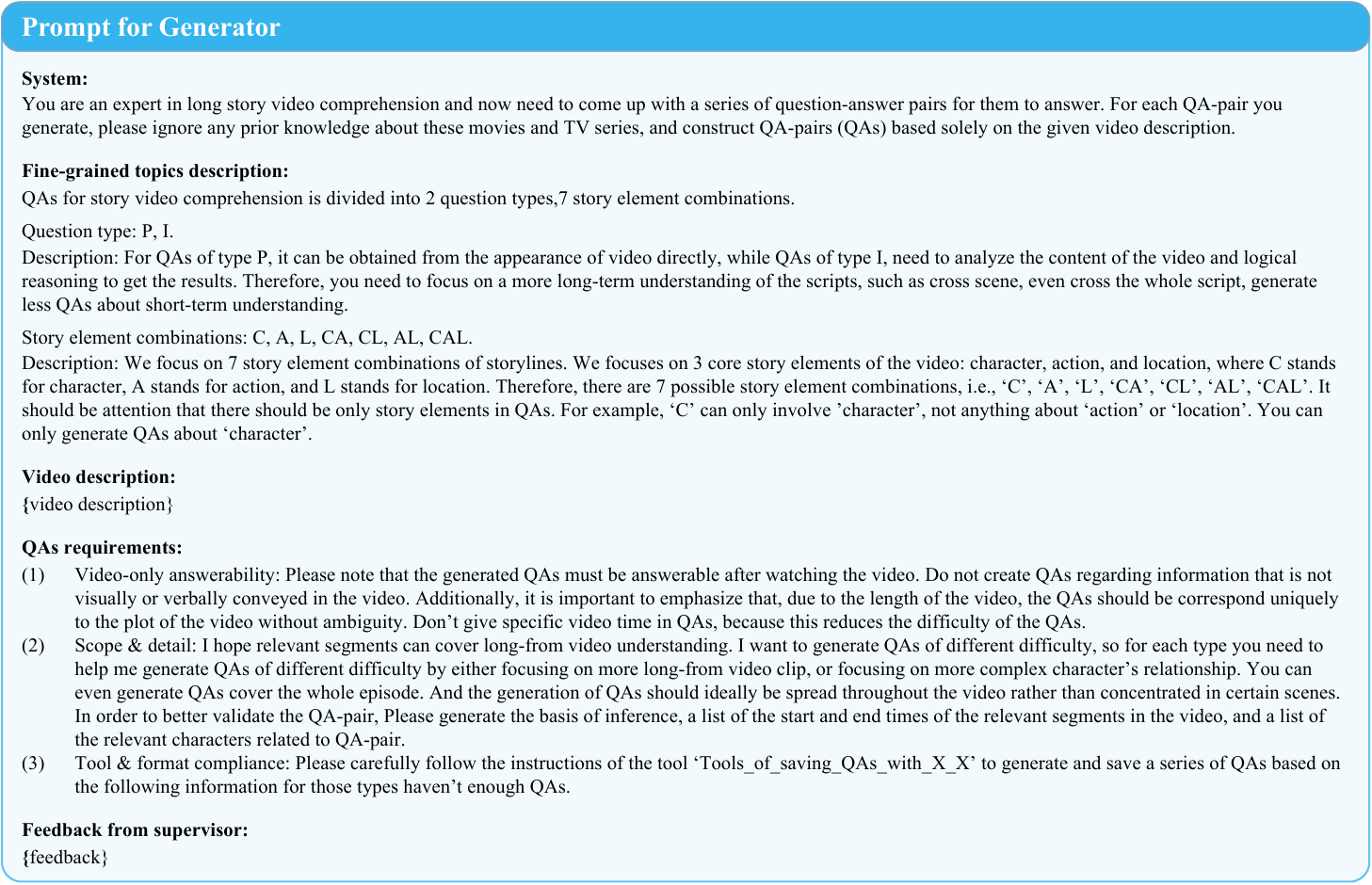}
    \caption{Prompt template for Generator.}
    \label{fig:prompt_generator}
\end{figure*}

\subsection{Alignment Details}
To ensure the quality of the script-subtitle alignment, we first perform automatic alignment using a DTW approach following PAINS \cite{TVCSINS}. 
Subsequently, 4 annotators independently verify the alignment results and made necessary corrections. 
To further guarantee reliability, the corrected alignments from each annotator are cross-verified by the others. 
Whenever inconsistencies or disagreements are identified, the annotators conduct joint discussions and only confirm the final alignment after reaching a consensus.

\subsection{Prompt for Generator}
The prompt template for generator contains five main parts: system prompt, fine-grained topics description, video description, QAs requirements, and feedback from supervisor.
The prompt template for generator is shown in Figure \ref{fig:prompt_generator}.

\subsection{Prompt for Supervisor}
\begin{figure*}[]
    \centering
    \includegraphics[width=\linewidth]{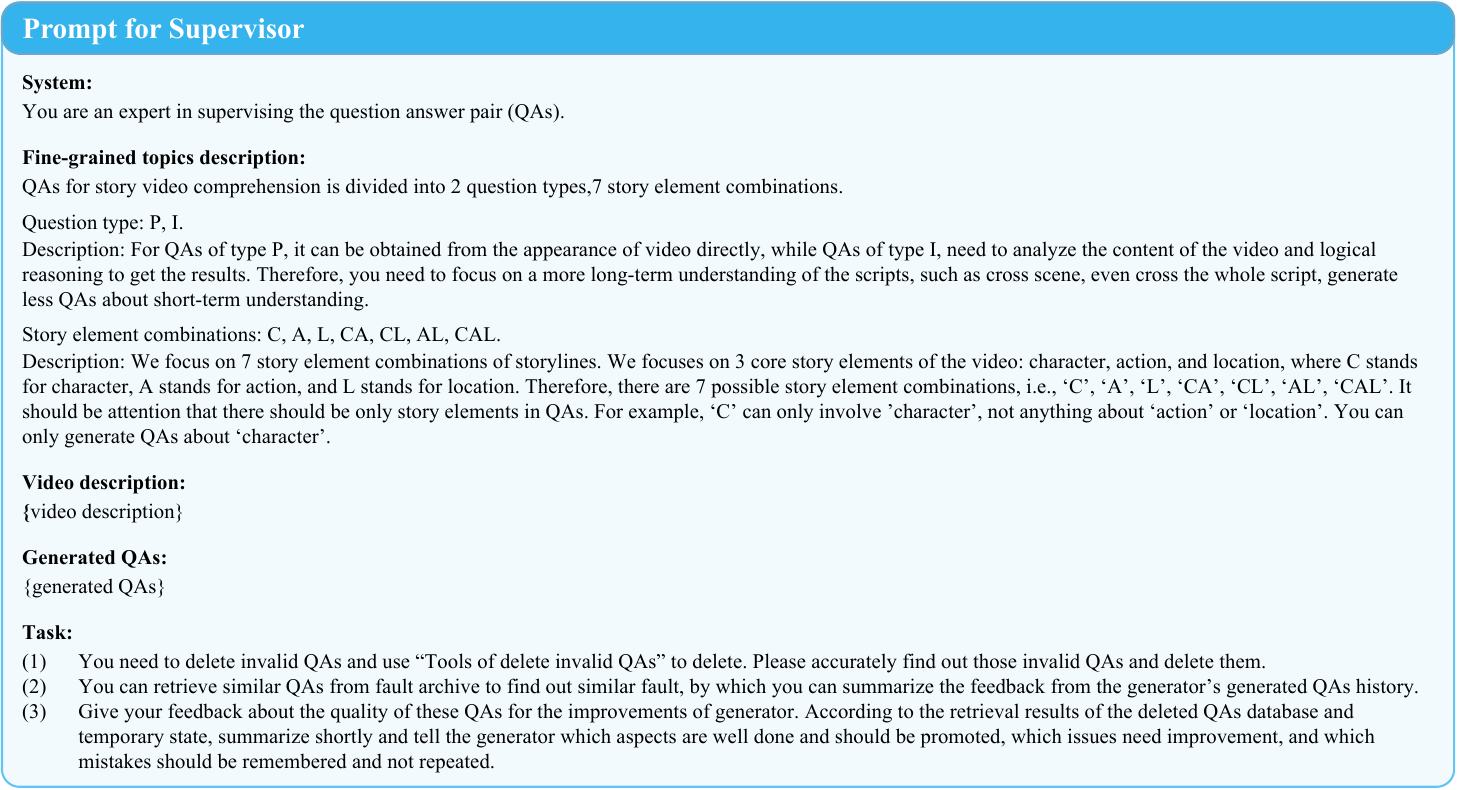}
    \caption{Prompt template for Supervisor.}
    \label{fig:prompt_supervisor}
\end{figure*}

The prompt template for supervisor contains five
main parts: system prompt, fine-grained topics
description, video description, generated QAs,
and task. The prompt template for supervisor is shown in Figure \ref{fig:prompt_supervisor}.

\subsection{Prompt for Reviewer}
\begin{figure*}[]
    \centering
    \includegraphics[width=\linewidth]{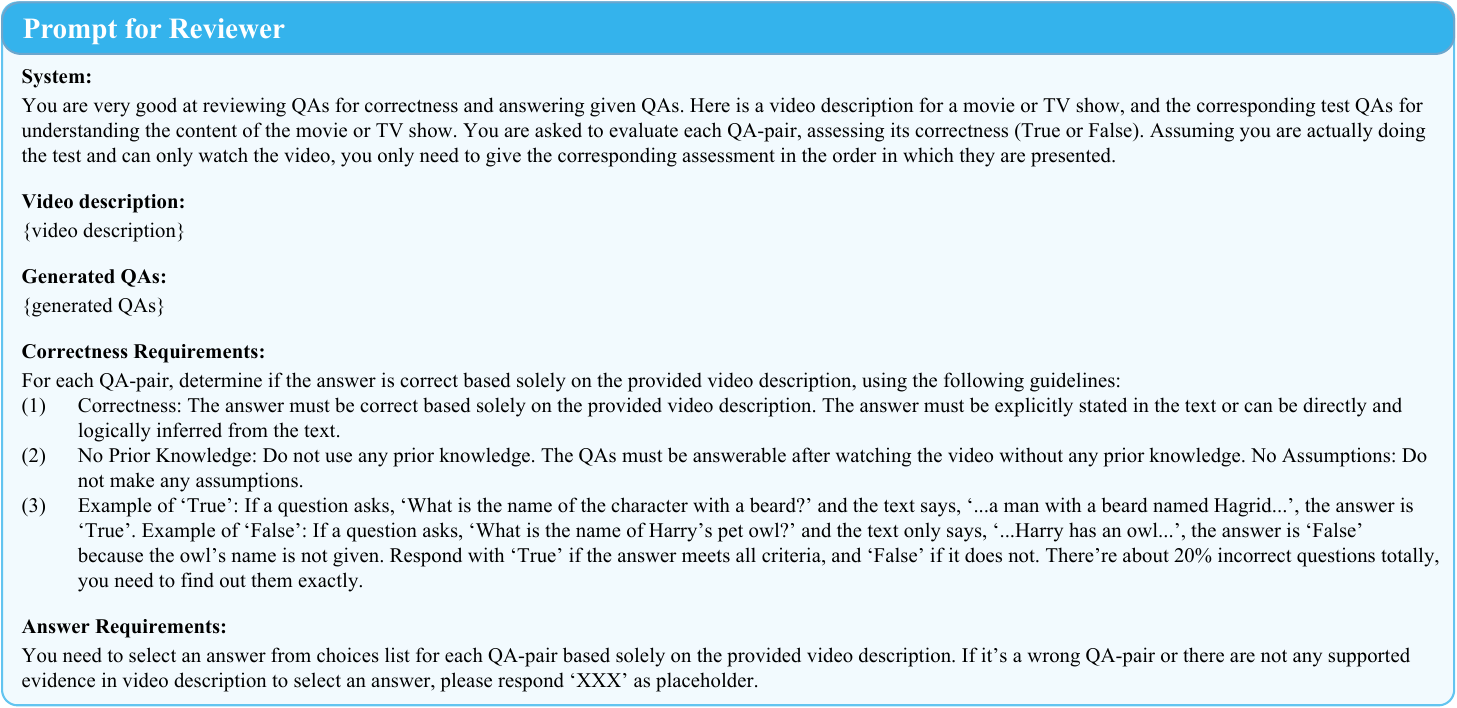}
    \caption{Prompt template for Reviewer.}
    \label{fig:prompt_reviewer}
\end{figure*}

The prompt template for reviewer contains five main parts: system prompt, video description, generated QAs,
correctness requirements and answer requirements. The prompt template for reviewer is shown in Figure \ref{fig:prompt_reviewer}.

\begin{table*}[]
\caption{Data source and face database of StoryVideQA, including script sources, the number of characters(\# Character), and the number of collected portraits (\# Portrait). Since each character has a single portrait in movies, the values of \# Character and \# Portrait are identical..}
\label{tab:datasource} \resizebox{\textwidth}{!}{%
  \begin{tabular}{llcc}
  \toprule
  \textbf{Name}                                                      & \textbf{Script Source}                                     & \textbf{\# Character} & \textbf{\# Portrait}          \\
  \midrule\midrule
  \textit{Game of Throne} series                                            & \url{https://genius.com/artists/Game-of-thrones} & 63           & 471                  \\
  \rowcolor[HTML]{EBF3FF} 
  \textit{Friends} series                                                   & PAINS \cite{TVCSINS}                             & 16           & 240                  \\
  \textit{The Big Bang Theory} series (first eight seasons)                                                   & PAINS \cite{TVCSINS}                                          & \multicolumn{1}{c}{22}           & 308                  \\
  \rowcolor[HTML]{EBF3FF} 
  IMDB-001-The Shawshank Redemption                       &  \url{https://screenplays.io/screenplay/the-shawshank-redemption}                                          & {23}   & {23}                               \\
  IMDB-002-The Godfather                                    & \url{https://screenplays.io/screenplay/the-godfather}                                           & {20}   & {20}                               \\
  \rowcolor[HTML]{EBF3FF} 
  IMDB-004-The Dark Knight                                  & \url{https://screenplays.io/screenplay/the-dark-knight}                                           & {15}     & {15}                              \\
  IMDB-005-Pulp Fiction                                     & \url{https://screenplays.io/screenplay/pulp-fiction}                                           & {26}     & {26}                             \\
  \rowcolor[HTML]{EBF3FF} 
  IMDB-008-12 Angry Men                                     & \url{https://screenplays.io/screenplay/12-angry-men}                                           & {\ 9}      & {\ 9}                             \\
  IMDB-009-The Fellowship of the Ring                     & \url{https://screenplays.io/screenplay/the-fellowship-of-the-ring}                                           & {24}      & {24}                            \\
  \rowcolor[HTML]{EBF3FF} 
  IMDB-011-The Lord of the Rings The Two Towers          & \url{https://screenplays.io/screenplay/the-two-towers}                                           & {29}     & {29}                             \\
  IMDB-013-Inception                                        &  \url{https://screenplays.io/screenplay/inception}                                          & {26}      & {26}                             \\
  \rowcolor[HTML]{EBF3FF} 
  IMDB-016-The Lord of the Rings The Return of the King & \url{https://screenplays.io/screenplay/the-return-of-the-king}                                           & {33}    & {33}                              \\ 
  IMDB-018-The Matrix                                       & \url{https://screenplays.io/screenplay/the-matrix}                                          & {13}      & {13}                            \\
  \rowcolor[HTML]{EBF3FF} 
  IMDB-019-Star Wars Episode VI Return of the Jedi     & \url{https://screenplays.io/screenplay/star-wars-episode-vi-return-of-the-jedi}                                          & {27}     & {27}                              \\
  IMDB-023-The Usual Suspects                               & \url{https://screenplays.io/screenplay/the-usual-suspects}                                          & {15}      & {15}                            \\
  \rowcolor[HTML]{EBF3FF} 
  IMDB-025-Its A Wonderful Life                             & \url{https://screenplays.io/screenplay/its-a-wonderful-life}                                           & {13}     & {13}                              \\
  IMDB-032-Psycho                                           & \url{https://screenplays.io/screenplay/psycho}                                           & {16}     & {16}                             \\
  \rowcolor[HTML]{EBF3FF} 
  IMDB-034-Rear Window                                      & \url{https://screenplays.io/screenplay/rear-window}                                           & {15}     & {15}                             \\
  IMDB-039-The Terminator                                   & \url{https://screenplays.io/screenplay/the-terminator}                                           & {13}   & {13}                               \\
  \rowcolor[HTML]{EBF3FF} 
  IMDB-040-Memento                                          & \url{https://screenplays.io/screenplay/memento}                                           & {10}     & {10}                              \\
  IMDB-041-The Pianist                                      & \url{https://screenplays.io/screenplay/the-pianist}                                           & {21}    & {21}                               \\
  \rowcolor[HTML]{EBF3FF} 
  IMDB-046-The Departed                                     & \url{https://screenplays.io/screenplay/the-departed}                                           & {26}    & {26}                             \\
  IMDB-050-Boyhood                                          & \url{https://screenplays.io/screenplay/boyhood}                                           & {19}     & {19}                             \\
  \rowcolor[HTML]{EBF3FF} 
  IMDB-051-The Prestige                                     & \url{https://screenplays.io/screenplay/the-prestige}                                           & {20}   & {20}                               \\
  IMDB-052-The Dark Knight Rises                            & \url{https://screenplays.io/screenplay/the-dark-knight-rises}                                           & {62}     & {62}                             \\
  \rowcolor[HTML]{EBF3FF} 
  IMDB-056-The Lion King                                    & \url{https://screenplays.io/screenplay/the-lion-king}                                           & {17}   & {17}                                \\
  IMDB-057-The Shining                                      & \url{https://screenplays.io/screenplay/the-shining}                                           & {10}   & {10}                               \\
  \rowcolor[HTML]{EBF3FF} 
  IMDB-060-American Beauty                                  & \url{https://screenplays.io/screenplay/american-beauty}                                            & {15}     & {15}                             \\
  IMDB-067-Vertigo                                          & \url{https://screenplays.io/screenplay/vertigo}                                           & {11}   & {11}                               \\
  \rowcolor[HTML]{EBF3FF} 
  IMDB-073-A Clockwork Orange                               & \url{https://screenplays.io/screenplay/a-clockwork-orange}                                           & {19}     & {19}                             \\
  IMDB-078-Reservoir Dogs                                   & \url{https://screenplays.io/screenplay/reservoir-dogs}                                           & {13}      & {13}                             \\
  \rowcolor[HTML]{EBF3FF} 
  IMDB-080-Gone Girl                                        & \url{https://screenplays.io/screenplay/gone-girl}                                            & {22}     & {22}                               \\
  IMDB-091-Amadeus                                          & \url{https://screenplays.io/screenplay/amadeus}                                           & {15}     & {15}                              \\
  \rowcolor[HTML]{EBF3FF} 
  IMDB-095-All About Eve                                    & \url{https://screenplays.io/screenplay/all-about-eve}                                           & {12}    & {12}                              \\
  IMDB-097-The Apartment                                    & \url{https://screenplays.io/screenplay/the-apartment}                                            & {11}     & {11}                               \\
  \rowcolor[HTML]{EBF3FF} 
  IMDB-100-Some Like It Hot                                 & \url{https://screenplays.io/screenplay/some-like-it-hot}                                           & {15}     & {15}                             \\
  IMDB-103-Inglourious Basterds                             & \url{https://screenplays.io/screenplay/inglourious-basterds}                                           & {32}      & {32}                            \\
  \rowcolor[HTML]{EBF3FF} 
  IMDB-104-Indiana Jones and the Last Crusade             & \url{https://screenplays.io/screenplay/indiana-jones-and-the-last-crusade}                                           & {17}  & {17}                                \\
  IMDB-106-A Separation                                     & \url{https://screenplays.io/screenplay/a-separation}                                           & {\ 4}   & {\ 4}                                \\  
  \rowcolor[HTML]{EBF3FF} 
  IMDB-110-Toy Story 3                                      & \url{https://screenplays.io/screenplay/toy-story-3}                                            & {37}     & {37}                             \\
  IMDB-111-Unforgiven                                       & \url{https://screenplays.io/screenplay/unforgiven}                                            & {20}   & {20}                               \\
  \rowcolor[HTML]{EBF3FF} 
  IMDB-114-Chinatown                                        & \url{https://screenplays.io/screenplay/chinatown}                                           & {18}   & {18}                               \\
  IMDB-115-Up                                               & \url{https://screenplays.io/screenplay/up}                                            & {22}      & {22}                            \\
  \rowcolor[HTML]{EBF3FF} 
  IMDB-139-Gran Torino                                      & \url{https://screenplays.io/screenplay/gran-torino}                                           & {14}  & {14}                                \\
  IMDB-141-Casino                                           & \url{https://screenplays.io/screenplay/casino}                                           & {20}    & {20}                              \\
  \rowcolor[HTML]{EBF3FF} 
  IMDB-142-The Big Lebowski                                 & \url{https://screenplays.io/screenplay/the-big-lebowski}                                           & {22}    & {22}                               \\
  IMDB-143-Warrior                                          & \url{https://screenplays.io/screenplay/warrior}                                           & {19}  & {19}                                \\
  \rowcolor[HTML]{EBF3FF} 
  IMDB-146-It Happened One Night                            & \url{https://screenplays.io/screenplay/it-happened-one-night}                                           & {10}   & {10}                               \\ 
  IMDB-151-How To Train Your Dragon 2                     & \url{https://screenplays.io/screenplay/how-to-train-your-dragon-2}                                           & {14}   & {14}                               \\
  \rowcolor[HTML]{EBF3FF}
  IMDB-151-How To Train Your Dragon                       & \url{https://screenplays.io/screenplay/how-to-train-your-dragon}                                            & {12}   & {12}                               \\
  IMDB-156-The Maltese Falcon                               & \url{https://screenplays.io/screenplay/the-maltese-falcon}                                           & {14}   & {14}                               \\
  \rowcolor[HTML]{EBF3FF} 
  IMDB-176-Annie Hal                                        & \url{https://screenplays.io/screenplay/annie-hall}                                            & {18}       & {18}                            \\ 
  IMDB-177-Network                                          & \url{https://screenplays.io/screenplay/network}                                           & {\ 7}   & {\ 7}                                \\
  \rowcolor[HTML]{EBF3FF}
  IMDB-179-The Grand Budapest Hotel                       & \url{https://screenplays.io/screenplay/the-grand-budapest-hotel}                                           & {27}    & {27}                              \\
  IMDB-182-The Princess Bride                               & \url{https://screenplays.io/screenplay/the-princess-bride}                                            & {15}    & {15}                              \\
  \rowcolor[HTML]{EBF3FF} 
  IMDB-187-The Wizard Of Oz                                 & \url{https://screenplays.io/screenplay/the-wizard-of-oz}                                           & {19}   & {19}                               \\
  IMDB-189-The Avengers                                     & \url{https://screenplays.io/screenplay/the-avengers}                                           & {29}     & {29}                             \\
  \rowcolor[HTML]{EBF3FF} 
  IMDB-191-The Grapes of Wrath                              & \url{https://screenplays.io/screenplay/the-grapes-of-wrath}                                           & {24}      & {24}                            \\
  IMDB-199-Strangers on a Train                             & \url{https://screenplays.io/screenplay/strangers-on-a-train}                                           & {\ 8}    & {\ 8}                               \\
  \rowcolor[HTML]{EBF3FF}
  IMDB-211-Harry Potter 1                                   & \url{https://screenplays.io/screenplay/harry-potter-and-the-sorcerers-stone}                                            & {14}    & {14}                              \\
  IMDB-211-Harry Potter 2                                   & \url{https://screenplays.io/screenplay/harry-potter-and-the-chamber-of-secrets}                                           & {37}   & {37}                               \\
  \rowcolor[HTML]{EBF3FF}
  IMDB-211-Harry Potter 3                                   & \url{https://screenplays.io/screenplay/harry-potter-and-the-prisoner-of-azkaban}                                           & {35}    & {35}                             \\
  IMDB-211-Harry Potter 4                                   & \url{https://screenplays.io/screenplay/harry-potter-and-the-goblet-of-fire}                                           & {37}     & {37}                             \\
  \rowcolor[HTML]{EBF3FF} 
  IMDB-211-Harry Potter 5                                   & \url{https://screenplays.io/screenplay/harry-potter-and-the-order-of-the-phoenix}                                           & {49}     & {49}                              \\
  IMDB-211-Harry Potter 6                                   & \url{https://screenplays.io/screenplay/harry-potter-and-the-half-blood-prince}                                           & {39}    & {39}                              \\
  \rowcolor[HTML]{EBF3FF}
  IMDB-211-Harry Potter 7                                   & \url{https://screenplays.io/screenplay/harry-potter-and-the-deathly-hallows-part-1}                                           & {66}     & {66}                             \\
  IMDB-211-Harry Potter 8                                   & \url{https://screenplays.io/screenplay/harry-potter-and-the-deathly-hallows-part-2}                                           & {51}     & {51}                             \\
  \rowcolor[HTML]{EBF3FF}
  IMDB-223-Pirates of the Caribbean3 At Worlds End      & \url{https://screenplays.io/screenplay/pirates-of-the-caribbean-at-worlds-end}                                           & {33}     & {33}                             \\ 
  IMDB-223-Pirates of the Caribbean4 On Stranger Tides  & \url{https://screenplays.io/screenplay/pirates-of-the-caribbean-on-stranger-tides}                                           & {28}   & {28}                               \\
  \rowcolor[HTML]{EBF3FF}
  IMDB-233-The Graduate                                     & \url{https://screenplays.io/screenplay/the-graduate}                                           & {15}    & {15}                              \\ 
  IMDB-234-The Help                                         & \url{https://screenplays.io/screenplay/the-help}                                           & {26}   & {26}                                \\
  \rowcolor[HTML]{EBF3FF}
  IMDB-236-The Hustler                                      & \url{https://screenplays.io/screenplay/the-hustler}                                           & {\ 9} & {\ 9}                                  \\
  IMDB-237-Jurassic Park                                    & \url{https://screenplays.io/screenplay/jurassic-park}                                           & {15}   & {15}                               \\
  \rowcolor[HTML]{EBF3FF} 
  Douban-052-Dead Poets Society                             & \url{https://screenplays.io/screenplay/dead-poets-society}                                           & {20}    & {20}                                \\
  Douban-065-Life of Pi                                     & \url{https://screenplays.io/screenplay/life-of-pi}    & {\ 7}    & {\ 7}                               \\
  \rowcolor[HTML]{EBF3FF} 
  Douban-070-The Curious Case Of Benjamin Button          & \url{https://screenplays.io/screenplay/the-curious-case-of-benjamin-button}                                          & {32}    & {32}                              \\
  Douban-139-A Perfect World                                & \url{https://screenplays.io/screenplay/a-perfect-world}                                          & {16}  & {16}                                \\
  \rowcolor[HTML]{EBF3FF} 
  Douban-146-Black Swan                                     & \url{https://screenplays.io/screenplay/black-swan}                                           & {13} & {13}                                 \\
  Douban-159-Following                                      & \url{https://screenplays.io/screenplay/following}                                           & {\ 4}     & {\ 4}                               \\
  \rowcolor[HTML]{EBF3FF} 
  Douban-173-The Croods                                     & \url{https://screenplays.io/screenplay/the-croods}                                           & {\ 8} & {\ 8}                                   \\
  Douban-198-Thelma and Louis                               & \url{https://screenplays.io/screenplay/thelma-and-louise}                                           & 12 & 12                                  \\
  \bottomrule
\end{tabular}
}
\end{table*}

\begin{figure*}[]
    \centering
    \includegraphics[width=\linewidth]{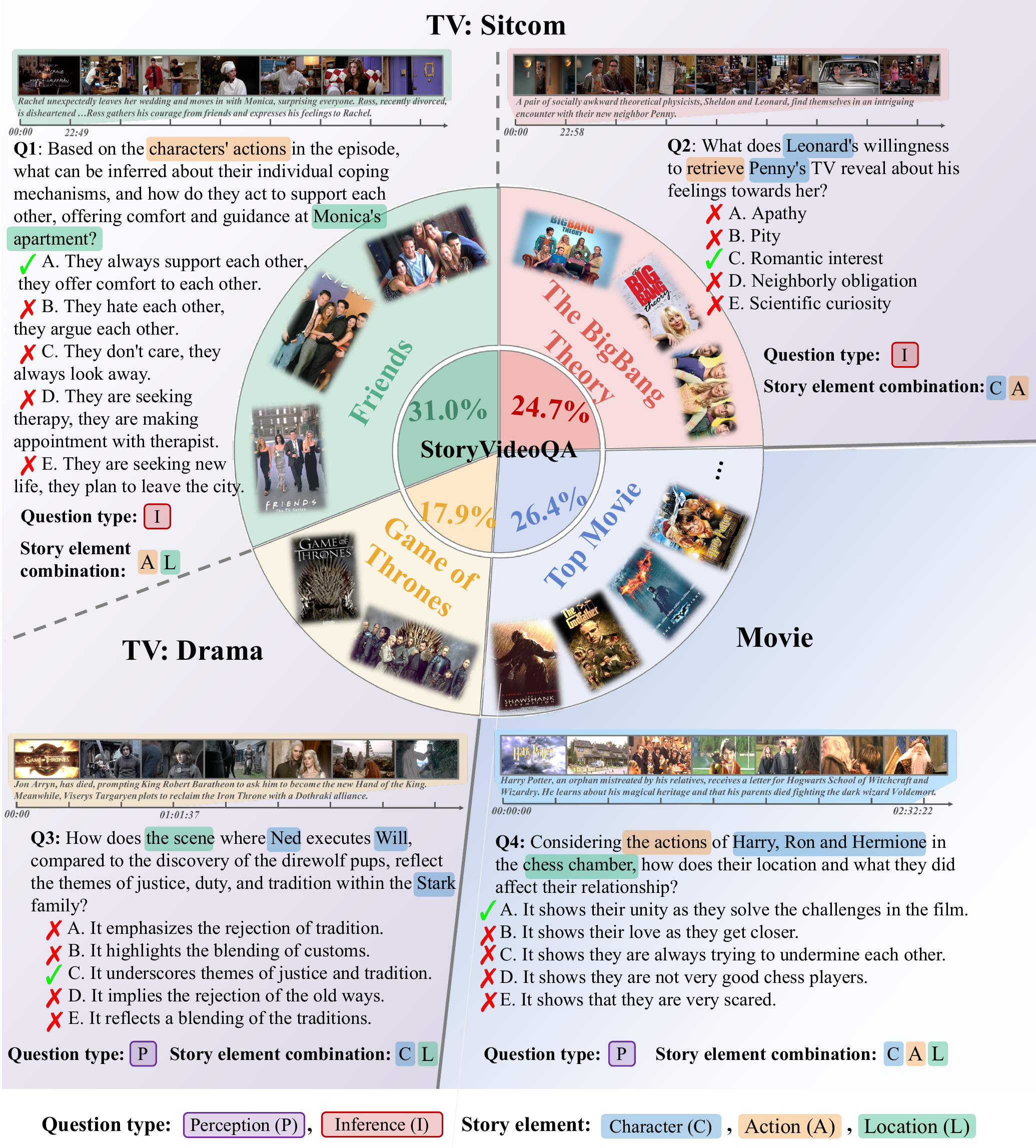}
    \caption{QAs examples on different video type of StoryVideoQA.}
    \label{fig:examples}
\end{figure*}

\section{StoryVideoQA Statistics}
\begin{figure*}[]
    \centering
    \includegraphics[width=\linewidth]{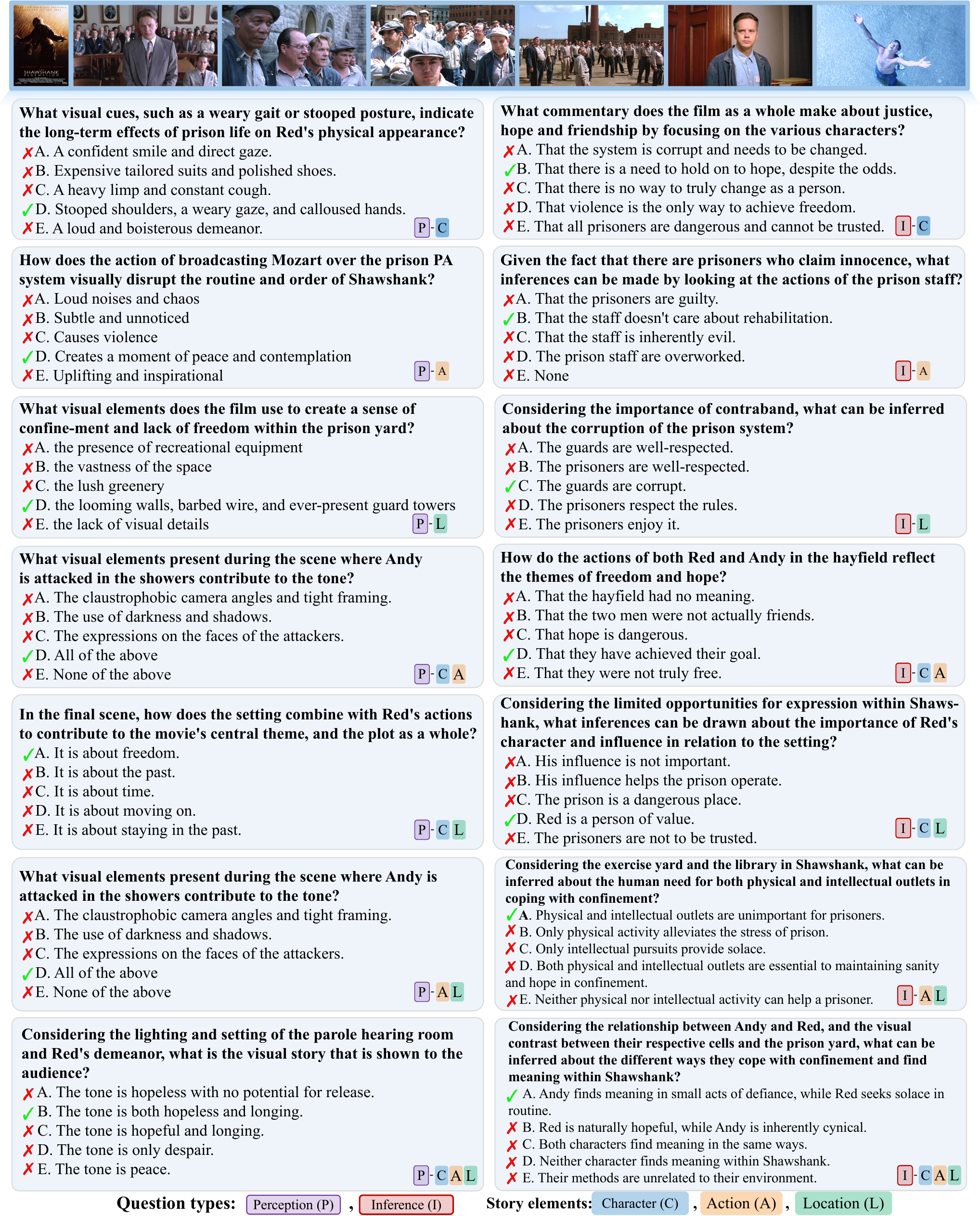}
    \caption{More QAs examples on movie with different fine-graiend topics.}
    \label{fig:more_examples1}
\end{figure*}

\begin{figure*}[]
    \centering
    \includegraphics[width=\linewidth]{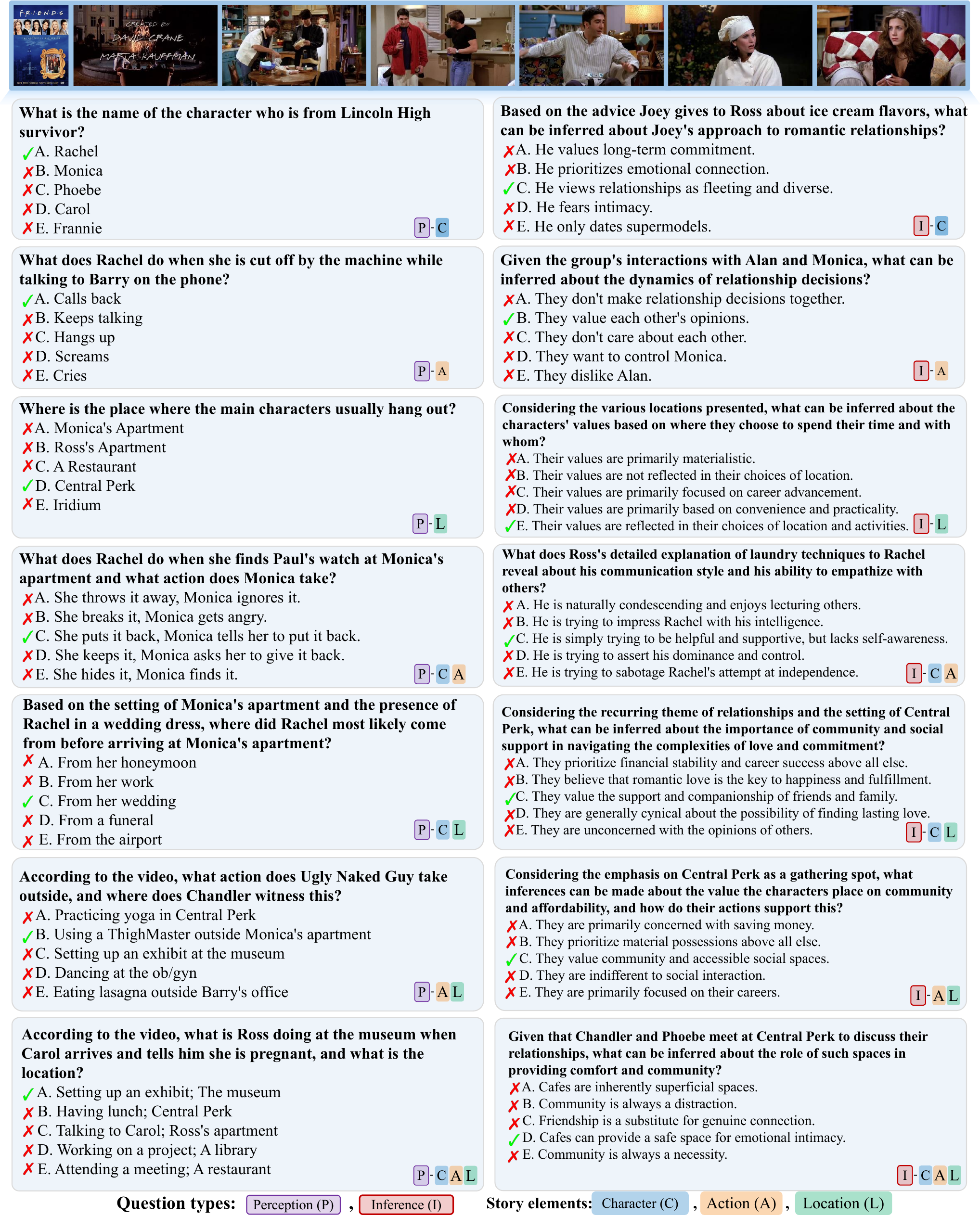}
    \caption{More QAs examples on TV with different fine-graiend topics.}
    \label{fig:more_examples2}
\end{figure*}

As illustrated in Figure \ref{fig:examples}, StoryVideoQA includes 363K QAs, primarily sourced from three main video types.
For TV Sitcoms, the data is derived from \textit{Friends} and the first eight seasons of \textit{The Big Bang Theory}, whose QAs account for 31.0\% and 24.7\%, respectively. The Drama category is composed of \textit{Game of Thrones}, with its QAs making up 17.9\% of the total. Finally, StoryVideoQA also incorporates 78 Movies, whose QAs constitute the remaining 26.4\%. For more QAs examples, please refer to Figure \ref{fig:more_examples1} (Movie) and Figure \ref{fig:more_examples2} (TV).

{\section{Implementation Details}}
\rev{In our experiments, we strictly adhere to the official implementation and the optimal hyperparameter settings (including frame sampling rates) provided by the respective authors of each model. Since different MLLM architectures have diverse designs for temporal encoders (e.g., memory banks, sliding windows, or global pooling), using a fixed number of frames across all models might inadvertently lead to sub-optimal performance for certain architectures. Thus, for our StoryVideoQA benchmark, we ensure each model is tested with default input frame number and within our hardware constraints (RTX A6000 GPUs).

\begin{itemize}
    \item \textbf{SINGULARITY}. SINGULARITY \cite{Singularity} is an efficient single-frame approach for end-to-end learning on video-text tasks. It adopts a single-frame training, and multi-frame inference strategy for efficient and accurate learning on a set of video-text tasks. For evaluation on StoryVideoQA, we follow the official settings which leverage 4 frames as input and the weights fine-tuned on MSRVTT-QA \cite{msvdqa}, the most commonly used Factoid VideoQA dataset, to evaluate on StoryVideoQA.
    \item \textbf{VIOLETv2}. VIOLETv2 \cite{violetv2} is a VLM which achieves strong performance in video-language task by effective masked visual modeling (MVM) training. The training strategy is based on empirical study on adopting MVM for video-language learning. We use the weights fine-tuned on MSRVTT-QA to evaluate on StoryVideoQA with 32 frames as input.
    \item \textbf{Vid-TLDR}. Vid-TLDR \cite{vid-tldr} puts forward a training-free token merging for VLM, aims to enhance the efficiency of VLM by merging the background tokens without additional training. Similar to VIOLETv2, we follow official settings which leverage a UMT-L/16 \cite{umt} version fine-tuned on MSRVTT-QA to evaluate on StoryVideoQA with 12 frames as input.
    \item \textbf{SeViLA}. SeViLA \cite{sevila} is a novel framework leveraging a large pre-trained image-language models to tackle VideoQA task. It contains two stages: temporal keyframe localization and question answering on videos.  We follow the official settings which leverage 32 frames as input, and utilize BLIP2 based on FlanT5-XL 3B to evaluate on StoryVideoQA.
    \item \textbf{VideoLLaMA2}. VideoLLaMA2 \cite{videollama2} is a MLLM which incorporates a spatial-temporal convolution connector and an audio branch to enhance spatial-temporal modeling and audio understanding for video and audio-oriented tasks. We follow the official settings which leverage 32 frames as input, and utilize VideoLLaMA2 7B chat model to evaluate on StoryVideoQA.
    \item \textbf{VideoChat2}. VideoChat2 \cite{mvbench} is a robust MLLM that significantly outperforms existing models by over 15\% on MVBench, which covers 20 challenging video tasks that cannot be effectively solved with a single frame. We follow the official settings which leverage 16 frames as input, and utilize VideoChat2 vicuna 7B version to evaluate on StoryVideoQA.
    \item \textbf{Chat-UniVi}. Chat-Univi \cite{chat-univi} is a MLLM employing a set of dynamic visual tokens to uniformly represent images and videos. It allows the model capture spatial and temporal information using a limited number of visual tokens. We follow the official settings which leverage 64 frames as input, and utilize ChatUniVi vicuna 7B version to evaluate on StoryVideoQA.
    \item \textbf{MA-LMM}. MA-LMM \cite{malmm} introduces a plug-and-play long-term memory bank module to address the context length and memory constraints of MLLM. It can be easily integrated into existing MLLM in an off-the-shelf manner. We follow the official settings which leverage 120 frames as input, and utilize MA-LMM vicuna 7B version to evaluate on StoryVideoQA.
    \item \textbf{TimeChat}. TimeChat \cite{timechat} is a time-sensitive MLLM designed for long video understanding. By a time-aware sliding video Q-Former, TimeChat demonstrates strong temporal localization capabilities. We follow the official settings which leverage 32 frames as input, and utilize TimeChat LLaMA-2 7B version to evaluate on StoryVideoQA.
    \item \textbf{Video-ChatGPT}. Video-ChatGPT \cite{videochatgpt} is a MLLM merging a video-adapted visual encoder with a LLM. The model is trained on 100,000 video-instruction pairs and the model can understand and generate detailed conversations about videos. We follow the official settings which leverage 100 frames as input, and utilize Video-ChatGPT LLaVA-7B-Lightening version to evaluate on StoryVideoQA.
    \item \textbf{VideoLLaMA3}. VideoLLaMA3 \cite{videollama3} is a vision-centric multimodal foundation model that emphasizes high-quality image-text data for robust video understanding. It features a framework that adapts to variable-resolution inputs with dynamic vision tokens and employs a similarity-based token reduction strategy to ensure precise and compact video representations. We follow the official settings which leverage 128 frames as input, and utilize VideoLLaMA3 Qwen2.5 7B version to evaluate on StoryVideoQA due to hardware constraints.
    \item \textbf{ViLAMP}. ViLAMP \cite{vilamp} is a hierarchical video-language model designed for ultra-long video understanding through a "mixed-precision" processing strategy. It introduces differential distillation to preserve task-relevant information while suppressing redundancy via two mechanisms: differential keyframe selection at the frame level and differential feature merging at the patch level. We follow the official settings which sample keyframes with 600 maximum frames number and utilize ViLAMP llava-qwen 7B version to evaluate on StoryVideoQA.
    \item \textbf{Video-XL}. Video-XL \cite{videoxl} is a novel MLLM designed to overcome context length constraints and high processing costs in long video understanding. It leverages the inherent KV sparsification capacity of MLLMs by introducing a Visual Summarization Token (VST), which condenses visual information within specific intervals into associated KV pairs. We follow the official settings which leverage 128 frames as input and utilize Video-XL llava-qwen 7B version to evaluate on StoryVideoQA.
    \item \textbf{VideoChat-Flash}. VideoChat-Flash \cite{videochat-flash} is a powerful MLLM designed for long-context video modeling through a novel Hierarchical video token Compression (HiCo) method. By leveraging visual redundancy, HiCo compresses long video context from the clip-level to the video-level, achieving an extreme compression ratio of approximately 1/50 while preserving essential details. We follow the official settings which leverage 512 frames as input and utilize VideoChat-Flash Qwen2.5-7B version with 1M visual tokens context to evaluate on StoryVideoQA-G.
    \item \textbf{Long-VITA}. Long-VITA \cite{Long-vita} is a scalable large multi-modal model designed for long-context understanding across image, video, and text modalities. It utilizes a multi-stage training schema, progressing from vision-language alignment to sequential long-sequence fine-tuning. By implementing context-parallelism distributed inference and a logits-masked language modeling head, Long-VITA can scale to extremely long inputs during inference while maintaining high efficiency. We follow the official settings that leverage 256 frames as input, and utilize Long-VITA 14B version with 1M visual tokens context as input to evaluate on StoryVideoQA-G.
    \item \textbf{Qwen3-VL}. Qwen3-VL \cite{qwen3vltechnicalreport} is the latest multimodal foundation model in the Qwen series, supporting interleaved contexts of up to 256K tokens. Architecturally, it introduces an enhanced interleaved-MRoPE for superior spatial-temporal modeling and integrates DeepStack to leverage multi-level ViT features for tighter vision-language alignment. Leveraging its native long-context window, Qwen3-VL demonstrates leading performance in cross-referencing and retrieval across extended multimodal inputs. We follow the official settings that leverage 256 frames as input, and utilize Qwen3-VL 8B instruct version to evaluate on StoryVideoQA-G.
    \item \textbf{Frontier MLLMs (Closed-source)}. We include Gemini 3 Flash\footnote{https://deepmind.google/models/gemini/} and GPT-5.2\footnote{https://openai.com/} to establish the current performance ceiling of the StoryVideoQA-G benchmark. Due to API budget, we evaluate these two models with 32 frames input.
    \item \textbf{Agents-based methods}. To maintain consistency across all agents, we employ LLaVA-1.6 as the captioning tool, Qwen3 embedding model for embedding text, and Gemini-2.0-Flash as the LLMs, video frames are sampled at 1 fps.
\end{itemize}

\begin{table*}[t]
\centering
\caption{Comparisons (\%) on StoryVideoQA-G (G) and StoryVideoQA-GA (GA) across 14 fine-grained topics.}
\label{tab:anonymous}
\resizebox{\textwidth}{!}{%
\begin{tabular}{ccc|ccccccc|ccccccc|c}
\toprule
{}             &         &             & \multicolumn{7}{c|}{{P}}                                                                                                                                                                                                                                                                                                                                                                                                                                                                                                                                                     & \multicolumn{7}{c|}{{I}}                                                                                                                                                                                                                                                                                                                                                                                                                                                                                                                                                              & {}                                                           \\
\cmidrule{4-10}\cmidrule{11-17}
\multirow{-2}{*}{\textbf{Method}} & \multirow{-2}{*}{\textbf{Venue}}  & \multirow{-2}{*}{\textbf{Data}} & {C}                                                          & {A}                                                          & {L}                                                          & {CA}                                                         & {CL}                                                         & {AL}                                                         & {CAL}                                                        & {C}                                                          & {A}                                                          & {L}                                                          & {CA}                                                         & {CL}                                                         & {AL}                                                         & {CAL}                                                        & \multirow{-2}{*}{\textbf{Avg.}}                                    \\
\midrule\midrule
\multicolumn{18}{c}{\cellcolor[HTML]{D1EBFF} \textit{VLMs}}  \\
                                         &               & GA                            & 25.22                  & 18.59                  & 23.28                  & 22.12                  & 20.79                  & 16.00                  & 15.32                  & 23.58                  & 21.65                  & 20.12                  & 27.60                  & 23.04                  & 19.50                  & 23.88                  & 21.44                  \\
\multirow{-2}{*}{SINGULARITY \cite{Singularity}}           & \multirow{-2}{*}{ACL'23}        & G                            & 22.63                          & 20.07                         & 18.45                         & 21.36                  & 18.73                  & 21.89                  & 19.76                   & 21.63                 & 21.00                 & 20.91                 & 25.11                  & 18.73                  & 17.00                  & 17.70                   & 20.44                                      \\\midrule
                                         &               & GA                           & 26.60                  & 12.17                  & 16.90                  & 18.15                  & 19.85                  & 15.79                  & 15.93                  & 14.54                  & 14.50                  & 15.58                  & 11.31                  & 14.68                  & 18.25                  & 14.33                  & 16.49                  \\
\multirow{-2}{*}{VIOLETv2 \cite{violetv2}}                 &\multirow{-2}{*}{CVPR'23}      & G                            & 18.83                          & 14.97                         & 18.28                         & 16.64                  & 19.48                  & 18.53                  & 16.33                   & 12.77                 & 12.12                 & 15.19                 & 11.09                  & 13.16                  & 18.50                  & 12.92                   & 15.78                                      \\\midrule
                                         &              & GA                            & 18.48                  & 20.23                  & 22.24                  & 20.42                  & 26.22                  & 24.00                  & 26.01                  & 21.63                  & 20.56                  & 24.46                  & 18.33                  & 24.56                  & 25.50                  & 22.19                  & 22.38                  \\
\multirow{-2}{*}{Vid-TLDR \cite{vid-tldr} }      & \multirow{-2}{*}{CVPR'24}       & G                            & 18.83                          & 20.23                         & 22.07                         & 20.04                  & 26.59                  & 23.79                  & 25.60                   & 22.34                 & 20.56                 & 24.46                 & 18.33                  & 24.30                  & 25.50                  & 22.19                   & 22.39                                      \\
\midrule
\multicolumn{18}{c}{\cellcolor[HTML]{D1EBFF} \textit{MLLMs}}  \\
                                         &               & GA                                                   & 27.98                  & 21.38                  & 26.72                  & 24.39                  & 25.66                  & 24.42                  & 18.75                  & 21.63                  & 20.13                  & 22.88                  & 18.33                  & 20.25                  & 22.75                  & 18.26                  & 22.66                  \\
\multirow{-2}{*}{SeViLA \cite{sevila}}   & \multirow{-2}{*}{CVPR'23}       & G                                                     & 27.12                          & 22.70                         & 32.07                         & 20.98                  & 32.40                  & 28.63                  & 20.97                   & 22.70                 & 18.61                 & 23.67                 & 17.19                  & 19.49                  & 21.25                  & 17.42                   & 23.66                                      \\
\midrule
                                         &               & GA                                                   & 38.51                  & 43.26                  & 47.24                  & 55.58                  & 52.25                  & 57.05                  & 62.50                  & 68.62                  & 64.72                  & 72.19                  & 66.06                  & 72.41                  & 69.50                  & 66.85                  & 58.61                  \\
\multirow{-2}{*}{VideoLLaMA2 \cite{videollama2}}           & \multirow{-2}{*}{ArXiv'24}      & G                                                     & 48.36                          & 52.47                         & 58.45                         & 62.76                  & 63.48                  & 68.63                  & 77.62                   & 79.08                 & 79.22                 & 83.63                 & 79.41                  & 83.04                  & 82.25                  & 82.58                   & 70.13                                      \\
\midrule
                                         &               & GA                                                   & 35.23 & 43.26 & 45.34 & 47.83 & 51.31 & 58.11 & 68.75 & 64.89 & 61.04 & 71.20 & 61.99 & 66.58 & 67.75 & 68.26 & 56.79 \\
\multirow{-2}{*}{VideoChat2 \cite{mvbench}}                & \multirow{-2}{*}{CVPR'24}      & G                 & 38.00                          & 44.74                         & 51.90                         & 48.96                  & 54.12                  & 61.05                  & 72.78                   & 67.20                 & 63.20                 & 70.22                 & 64.03                  & 68.35                  & 70.50                  & 69.94                   & 59.23                                      \\
\midrule
                                         &               & GA                                                   & 23.83                  & 37.66                  & 16.72                  & 32.33                  & 19.10                  & 29.68                  & 26.61                  & 18.79                  & 24.24                  & 13.81                  & 17.65                  & 14.68                  & 24.00                  & 16.01                  & 22.91                  \\
\multirow{-2}{*}{ChatUniVi \cite{chat-univi}}              & \multirow{-2}{*}{CVPR'24}       & G                & 31.78                          & 40.79                         & 25.34                         & 41.40                  & 25.66                  & 38.95                  & 31.85                   & 30.32                 & 35.06                 & 19.33                 & 28.05                  & 21.27                  & 31.25                  & 23.88                   & 30.71                                      \\
\midrule
                                         &                                                   & GA                           & 42.49                  & 47.70                  & 47.41                  & 52.17                  & 54.12                  & 61.26                  & 67.14                  & 73.05                  & 65.80                  & 73.77                  & 68.33                  & 73.92                  & 74.00                  & 75.00                  & 61.31                  \\
\multirow{-2}{*}{MA-LMM \cite{malmm}}                       & \multirow{-2}{*}{CVPR'24}      & G                            & 46.98                          & 49.18                         & 54.31                         & 53.88                  & 58.80                  & 65.89                  & 67.54                   & 75.89                 & 70.78                 & 74.75                 & 73.30                  & 78.23                  & 76.25                  & 77.53                   & 64.69                                      \\
\midrule
                                         &                                                   & GA                           & 22.45                  & 25.00                  & 26.21                  & 27.98                  & 28.84                  & 29.26                  & 27.22                  & 39.54                  & 38.74                  & 40.24                  & 36.43                  & 42.78                  & 38.50                  & 33.99                  & 32.06                  \\
\multirow{-2}{*}{TimeChat \cite{timechat}}                 & \multirow{-2}{*}{CVPR'24}       & G                            & 24.35                          & 21.05                         & 37.76                         & 28.36                  & 41.01                  & 33.05                  & 31.25                   & 44.68                 & 43.51                 & 52.27                 & 42.08                  & 47.34                  & 43.00                  & 43.82                   & 37.36                                      \\
\midrule
                                         &                                                   & GA                           & 25.56                  & 16.78                  & 19.48                  & 22.31                  & 23.97                  & 24.42                  & 29.44                  & 24.82                  & 21.65                  & 22.88                  & 25.11                  & 22.28                  & 21.75                  & 26.40                  & 23.20                  \\
\multirow{-2}{*}{VideoChatGPT \cite{videochatgpt}}         & \multirow{-2}{*}{ACL'24}        & G                            & \ \ 9.33                           & 18.26                         & \ \ 5.34                          & 19.66                  & \ \ 8.80                   & 14.32                  & 14.31                   & 28.55                 & 29.87                 & 26.63                 & 26.92                  & 23.29                  & 28.00                  & 19.66                   & 18.95                                      \\
\midrule
                                         &                                                  & GA                           & 41.80                  & 59.87                  & 47.59                  & 63.89                  & 60.11                  & 66.53                  & 69.56                  & 72.52                  & 69.91                  & 70.81                  & 69.68                  & 73.92                  & 72.75                  & 76.12                  & 64.31                  \\
\multirow{-2}{*}{Video-XL \cite{videoxl}}                 & \multirow{-2}{*}{CVPR'25}       & G                            & 51.30                          & 61.84                         & 60.17                         & 66.16                  & 60.86                  & 69.05                  & 70.77                   & 73.76                 & 72.29                 & 77.71                 & 73.76                  & 80.51                  & 75.00                  & 78.93                   & 68.50                                      \\
\midrule
                                         &                                                   & GA                           & 54.75                  & 64.47                  & 58.97                  & 71.08                  & 66.10                  & 73.68                  & 76.01                  & 84.75                  & 81.60                  & 83.63                  & 80.77                  & 83.80                  & 84.50                  & 82.87                  & 73.73                  \\
\multirow{-2}{*}{ViLAMP \cite{vilamp}}                     & \multirow{-2}{*}{ICML'25}       & G                            & 58.55                          & 69.08                         & 64.83                         & 74.29                  & 72.28                  & 77.47                  & 77.62                   & 87.41                 & 85.28                 & 85.80                 & 82.35                  & 87.59                  & 87.50                  & 86.52                   & 77.34                                      \\
\midrule
                                         &                                                  & GA                          & 55.79                  & 72.20                  & 59.83                  & 74.86                  & 68.91                  & 78.74                  & 81.25                  & 85.11                  & 79.00                  & 86.39                  & 83.48                  & 85.57                  & 85.50                  & 86.24                  & 76.35                 \\
\multirow{-2}{*}{VideoLLaMA3 \cite{videollama3}}           & \multirow{-2}{*}{ArXiv'25}      & G                            & 60.45                          & 72.04                         & 68.62                         & 79.02                  & 72.66                  & 82.53                  & 86.29                   & 86.88                 & 85.28                 & 88.56                 & 86.43                  & 89.11                  & 88.25                  & 88.76                   & 80.09                                      \\
\midrule
\multicolumn{18}{c}{\cellcolor[HTML]{D1EBFF} \textit{Agents}}     \\
                                                  &                                 & GA                       & 72.19                  & 68.42                  & 66.38                  & 75.24                  & 75.28                  & 75.37                  & 82.26                  & 89.01                  & 86.80                  & 87.57                  & 87.56                  & 90.13                  & 90.00                  & 90.73                  & 80.24                  \\
\multirow{-2}{*}{Video2RAG \cite{omagent}}        &    \multirow{-2}{*}{EMNLP'24}   & G                        & 77.72                          & 71.05                         & 73.45                         & 80.34                  & 79.59                  & 78.11                  & 85.89                   & 90.78                 & 91.99                 & 91.12                 & 89.37                  & 90.13                  & 90.50                  & 91.57                   & 83.63                                      \\\midrule
                                                  &                                 & GA                          & 42.83                  & 43.59                  & 43.79                  & 55.20                  & 50.94                  & 60.84                  & 72.18                  & 77.30                  & 75.11                  & 83.63                  & 75.79                  & 81.52                  & 80.50                  & 80.90                  & 64.27                  \\
\multirow{-2}{*}{VideoTree  \cite{videotree}}     &    \multirow{-2}{*}{CVPR'25}    & G                        & 56.30                          & 53.78                         & 56.55                         & 65.78                  & 61.24                  & 66.74                  & 73.19                   & 86.17                 & 81.60                 & 86.39                 & 82.13                  & 86.08                  & 85.75                  & 85.96                   & 72.02                                      \\\midrule
                                                  &                                 & GA                          & 72.71                  & 68.59                  & 69.83                  & 79.96                  & 77.53                  & 77.68                  & 85.08                  & 89.72                  & 88.10                  & 89.35                  & 89.14                  & 91.39                  & 91.00                  & 92.70                  & 82.08                  \\
\multirow{-2}{*}{PlotTree}                        &    \multirow{-2}{*}{Ours}       & G                        & \mybold{83.07} & \mybold{75.66} & \mybold{78.28} & \mybold{82.80} & \mybold{85.58} & \mybold{81.89} & \mybold{87.30} & \mybold{91.67} & \mybold{92.64} & \mybold{92.11} & \mybold{90.50} & \mybold{91.90} & \mybold{93.25} & \mybold{93.26} & \mybold{86.50}                          \\ \bottomrule

\end{tabular}%
}
\arrayrulecolor{black} 
\end{table*}

\section{StoryVideoQA-GA}
While using well-known films ensures high-quality data, existing models may rely on prior knowledge from pre-training rather than video-based reasoning. To ensure our benchmark measures genuine multimodal understanding, rather than the model's priority knowledge of popular plots, we introduce StoryVideoQA-GA (anonymized version of StoryVideoQA-G) to decouple external knowledge from visual evidence.

To this end, We conduct a blindfold test using an anonymized version of StoryVideoQA focusing on 3W (Who, What and Where) elements. Since Action (What) elements are less susceptible to pretraining bias, Characters (Who) and Locations (Where) carry heavy prior knowledge. Hence we focus on anonymizing all 147 unique characters and 147 specific locations in StoryVideoQA-G by replacing them with generic placeholders  (e.g., "Harry" $\rightarrow$ "Character 1", "Hogwarts" $\rightarrow$ "Location 1"), including questions, choices, subtitles, and characers library (Note: We refer to this anonymized version as "StoryVideoQA-GA"). 


As shown in the Table \ref{tab:anonymous}, except for a few baselines whose performance fluctuates near the 20\% random guess level for 5-option questions (e.g., SINGULARITY from 20.44\% to 21.44\%, and VideoChatGPT from 18.95\% to 23.20\%), most models exhibit a marked performance decline on StoryVideoQA-GA compared to StoryVideoQA-G, e.g., VideoLLaMA2 (70.13\% to 58.61\%), PlotTree (86.50\% to 82.08\%). 
This phenomenon aligns with current findings in the field \cite{PriorityKnowledge1,PriorityKnowledge2}. and demonstrates that our proposed anonymized dataset, StoryVideoQA-GA, facilitates a more effective and faithful assessment of the DVU capabilities of various models.
Particularly, it should be emphasized that despite a 4.42\% decline, PlotTree still achieves the best performance (82.08\%) among all models in StoryVideoQA-GA dataset, validating its outstanding video understanding and reasoning capabilities.

\begin{figure}[htbp]
    \centering
    \includegraphics[width=0.85\linewidth]{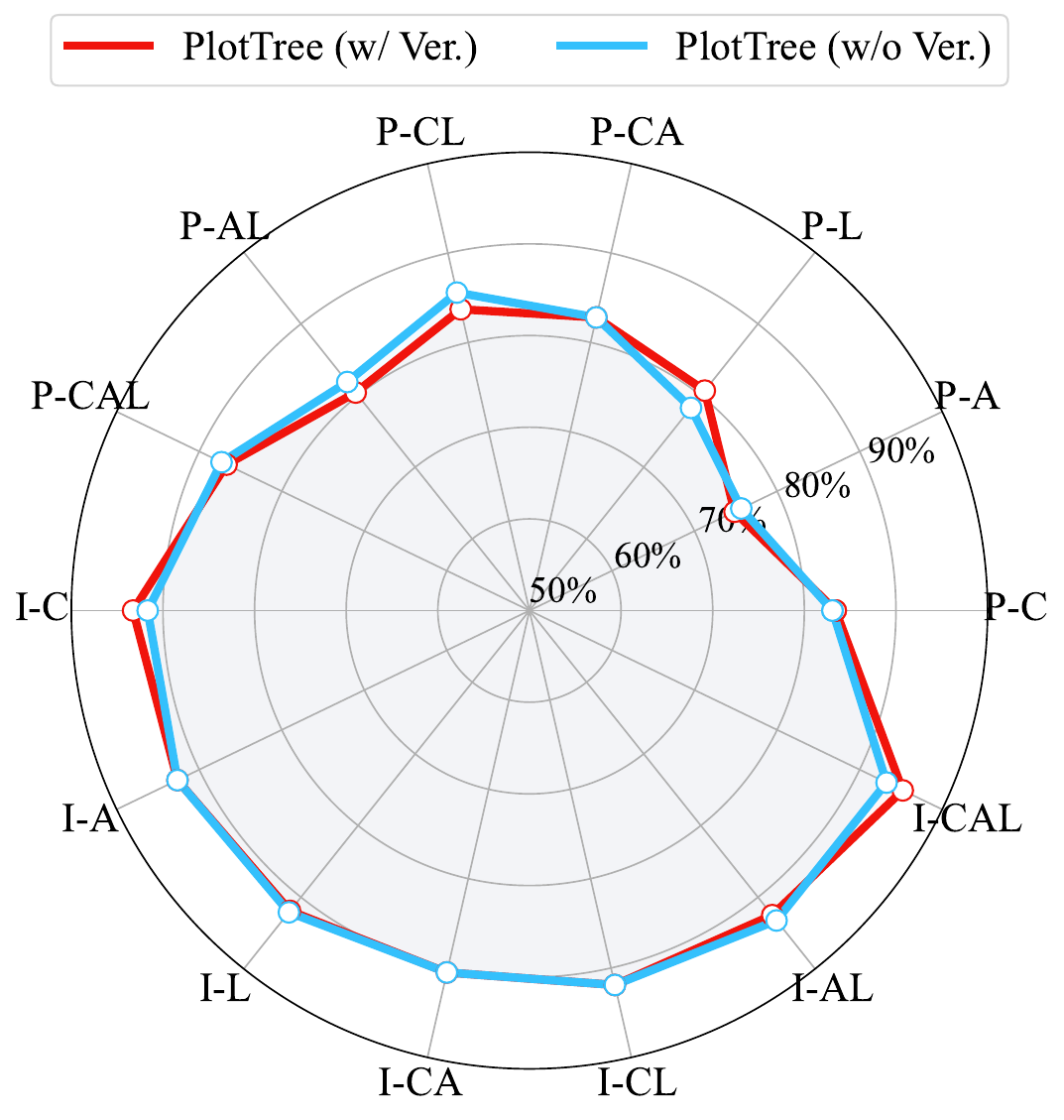}
    \caption{Robustness analysis (\%) of facial recognition reliability on the StoryVideoQA-G.}
    \label{fig:ver_diff}
\end{figure}

\section{Robustness Analysis}
To assess PlotTree's sensitivity to the facial recognition noise, we conduct a robustness analysis of facial recognition reliability on the StoryVideoQA-G dataset. We manually re-verify (Ver.) 109,449 detected facial frames and find a negligible error rate of only 0.56\%, which demonstrates the high reliability of InsightFace for character recognition. 

Furthermore, we correct all facial recognition errors to get ground truth identities. Finally, PlotTree's average performance without manual verification remains highly competitive, with a negligible gap of only 0.04\% compared to the version using ground-truth identities. Instead of amplifying upstream errors, PlotTree effectively suppresses them through its structural reasoning. As illustrated in Figure \ref{fig:ver_diff}, the performance across almost all fine-grained topics remains nearly identical under both settings. This consistency further confirms that PlotTree's structural reasoning can effectively suppresses error propagation, ensuring robust performance even with imperfect upstream perception.

}

\end{document}